\documentclass[acmsmall,screen]{acmart}

\AtBeginDocument{%
  \providecommand\BibTeX{{%
    \normalfont B\kern-0.5em{\scshape i\kern-0.25em b}\kern-0.8em\TeX}}}

\setcopyright{acmcopyright}
\copyrightyear{2018}
\acmYear{2018}
\acmDOI{XXXXXXX.XXXXXXX}

\acmJournal{JACM}
\acmVolume{37}
\acmNumber{4}
\acmArticle{1}
\acmMonth{8}

\usepackage{hyperref}       
\usepackage{amsfonts}       
\usepackage{nicefrac}       
\usepackage{microtype}      
\usepackage{lipsum}
\usepackage{graphicx}
\usepackage{amsmath}
\usepackage{bbding}
\usepackage{pifont}
\usepackage{wasysym}
\usepackage{color}
\usepackage{tikz}
\usepackage[edges]{forest}
\definecolor{hiddendraw}{RGB}{205, 44, 36}
\definecolor{hidden-blue}{RGB}{194,232,247}
\definecolor{hidden-orange}{RGB}{243,202,120}
\definecolor{hidden-yellow}{RGB}{242,244,193}
\tikzstyle{mybox}=[
    rectangle,
    draw=hiddendraw,
    rounded corners,
    text opacity=1,
    minimum height=1.5em,
    minimum width=5em,
    inner sep=2pt,
    align=center,
    fill opacity=.5,
    ]
\usepackage{amssymb}
\usepackage{pifont}
\usepackage{amsthm}
\usepackage{amsmath}
\usepackage{bbding}
\usepackage{multirow} 
\usepackage{subfigure}
\theoremstyle{definition}

\usepackage{mathtools}
\usepackage{colortbl}

\usepackage{color}
\usepackage{tikz}
\usepackage[edges]{forest}
\usepackage{enumitem}
\usepackage{wrapfig} 
\usepackage{makecell}

\begin{document}

\title{A Survey on Large Language Models for Code Generation}

\author{Juyong Jiang}
\email{jjiang472@connect.hkust-gz.edu.cn}
\authornote{Equally major contributors.}
\affiliation{
  \institution{The Hong Kong University of Science and Technology (Guangzhou)}
  \city{Guangzhou}
  \country{China}}

\author{Fan Wang}
\email{fwang380@connect.hkust-gz.edu.cn}
\authornotemark[1]
\affiliation{
  \institution{The Hong Kong University of Science and Technology (Guangzhou)}
  \city{Guangzhou}
  \country{China}}

\author{Jiasi Shen}
\email{sjs@cse.ust.hk}
\authornote{Corresponding authors.}
\affiliation{
  \institution{The Hong Kong University of Science and Technology}
  \city{Hong Kong}
  \country{China}}

\author{Sungju Kim}
\email{sungju.kim@navercorp.com}
\authornotemark[2]
\affiliation{
  \institution{NAVER Cloud} 
  \city{Seoul}
  \country{South Korea}}
  
\author{Sunghun Kim}
\email{hunkim@cse.ust.hk}
\authornotemark[2]
\affiliation{
  \institution{The Hong Kong University of Science and Technology (Guangzhou)}
  \city{Guangzhou}
  \country{China}}

\newcommand{\sjs}[1]{{\color{black} #1}}
\newcommand{\kath}[1]{{\color{black} #1}}
\newcommand{\revise}[1]{{\color{black} #1}}
\newcommand{\done}[1]{{\color{black} #1}}

\begin{abstract}
Large Language Models (LLMs) have garnered remarkable advancements across diverse code-related tasks, known as Code LLMs, particularly in code generation that generates source code with LLM from natural language descriptions. This burgeoning field has captured significant interest from both academic researchers and industry professionals due to its practical significance in software development, \textit{e.g., GitHub Copilot}. Despite the active exploration of LLMs for a variety of code tasks, either from the perspective of natural language processing (NLP) or software engineering (SE) or both, there is a noticeable absence of a comprehensive and up-to-date literature review dedicated to LLM for code generation. In this survey, we aim to bridge this gap by providing a systematic literature review that serves as a valuable reference for researchers investigating the cutting-edge progress in LLMs for code generation. We introduce a taxonomy to categorize and discuss the recent developments in LLMs for code generation, covering aspects such as data curation, latest advances, performance evaluation, ethical implications, environmental impact, and real-world applications. In addition, we present a historical overview of the evolution of LLMs for code generation and offer an empirical comparison using the HumanEval, MBPP, and BigCodeBench benchmarks across various levels of difficulty and types of programming tasks to highlight the progressive enhancements in LLM capabilities for code generation. We identify critical challenges and promising opportunities regarding the gap between academia and practical development. Furthermore, we have established a dedicated resource GitHub page (\href{https://github.com/juyongjiang/CodeLLMSurvey}{https://github.com/juyongjiang/CodeLLMSurvey}) to continuously document and disseminate the most recent advances in the field.
\end{abstract}

\ccsdesc[500]{General and reference~Surveys and overviews}
\ccsdesc[500]{Software and its engineering~Software development techniques}
\ccsdesc[500]{Computing methodologies~Artificial intelligence}

\keywords{Large Language Models, Code Large Language Models, Code Generation}

\maketitle

\section{Introduction}\label{sec:introduction}
The advent of Large Language Models (LLMs) such as ChatGPT\footnote{\href{https://chat.openai.com/}{https://chat.openai.com}} \cite{gpt-3.5-turbo} has profoundly transformed the landscape of automated code-related tasks \cite{chen2021evaluating}, including code completion \cite{wang2021code,lu2022reacc,guo2023longcoder,wu2024repoformer}, code translation \cite{lachaux2020unsupervised,szafraniec2022code,chen2018tree}, and code repair \cite{olausson2023self,fan2023automated,joshi2023repair,parasaram2024fact,xu2024aligning,zhang2024pydex}. 
A particularly intriguing application of LLMs is code generation, a task that involves producing source code from natural language descriptions. Despite varying definitions across studies \cite{ren2020codebleu,chen2023teaching,shojaee2023execution,wang2023codet5+}, \done{for the main scope of this survey, we focus on the code generation task and adopt a consistent definition of code generation as the natural-language-to-code (NL2Code) task \cite{austin2021program,athiwaratkun2022multi,zan2023large}.} 
\done{To enhance clarity, the differentiation between code generation and other code-related tasks, along with a more nuanced definition, is summarized in Table \ref{tab:code_tasks}.}
This area has garnered substantial interest from both academia and industry, as evidenced by the development of tools like GitHub Copilot\footnote{\href{https://github.com/features/copilot}{https://github.com/features/copilot}} \cite{chen2021evaluating}, CodeGeeX\footnote{\href{https://codegeex.cn/en-US}{https://codegeex.cn/en-US}} \cite{zheng2023codegeex}, and Amazon CodeWhisperer\footnote{\href{https://aws.amazon.com/codewhisperer}{https://aws.amazon.com/codewhisperer}}, which leverage groundbreaking code LLMs to facilitate software development.

Initial investigations into code generation primarily utilized heuristic rules or expert systems, such as probabilistic grammar-based frameworks \cite{joshi2003formalism,cohn2010inducing,allamanis2014mining,xiong2017precise,ji2020question} and specialized language models \cite{de2008z3,gulwani2010dimensions,jha2010oracle}. 
These early techniques were typically rigid and difficult to scale. 
However, the introduction of Transformer-based LLMs has shifted the paradigm, establishing them as the preferred method due to their superior proficiency and versatility.
One remarkable aspect of LLMs is their capability to follow instructions \cite{wei2022emergent,ouyang2022training,xu2023wizardlm,muennighoff2023octopack,chung2024scaling}, enabling even novice programmers to write code by simply articulating their requirements. This emergent ability has democratized coding, making it accessible to a broader audience \cite{zan2023large}. 
The performance of LLMs on code generation tasks has seen remarkable improvements, as illustrated by the HumanEval leaderboard\footnote{\href{https://paperswithcode.com/sota/code-generation-on-humaneval}{https://paperswithcode.com/sota/code-generation-on-humaneval}}, which showcases the evolution from PaLM 8B \cite{chowdhery2023palm} of 3.6\% to LDB \cite{zhong2024ldb} of 95.1\% on \texttt{Pass@1} metrics. 
As can be seen, the HumanEval benchmark \cite{chen2021evaluating} has been established as a de facto standard for evaluating the coding proficiency of LLMs \cite{chen2021evaluating}.
 
\begin{figure*}[p!]
\centering
\includegraphics[width=0.97\linewidth]{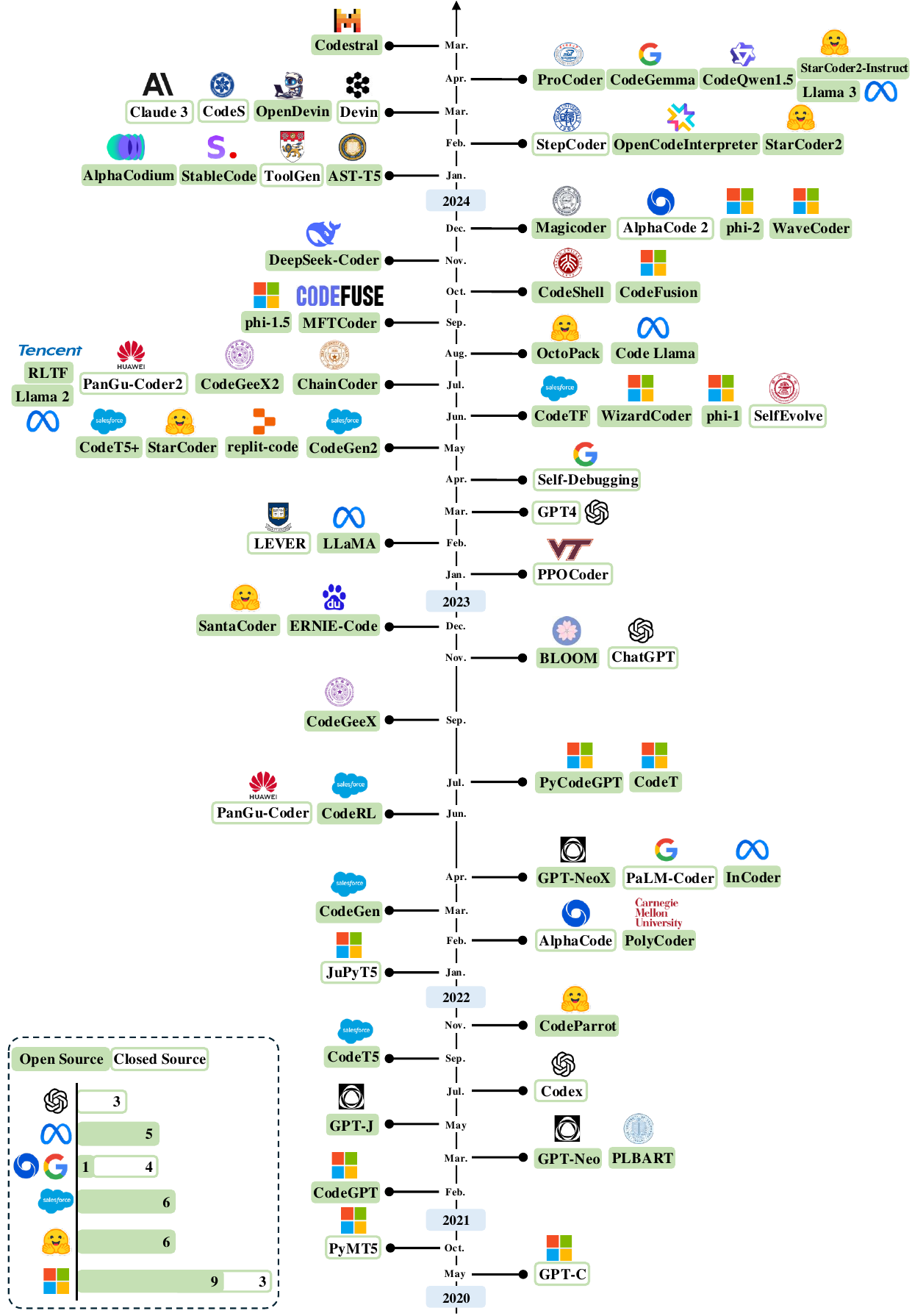}
\caption{A chronological overview of large language models (LLMs) for code generation in recent years. The timeline was established mainly according to the release date. The models with publicly available model checkpoints are highlighted in green color.}
\label{fig:timeline}
\end{figure*}

To offer a comprehensive chronological evolution, we present an overview of the development of LLMs for code generation, as illustrated in Figure \ref{fig:timeline}. 
The landscape of LLMs for code generation is characterized by a spectrum of models, with certain models like ChatGPT \cite{ouyang2022training}, GPT4 \cite{achiam2023gpt}, LLaMA \cite{touvron2023llama,touvron2023llama2}, and Claude 3 \cite{claude3} serving general-purpose applications, while others such as StarCoder \cite{li2023starcoder,lozhkov2024starcoder}, Code LLaMA \cite{roziere2023code}, DeepSeek-Coder \cite{guo2024deepseek}, and Code Gemma \cite{codegemma_2024} are tailored specifically for code-centric tasks.
The convergence of code generation with the latest LLM advancements is pivotal, especially when programming languages can be considered as distinct dialects of multilingual natural language \cite{athiwaratkun2022multi,zheng2023codegeex}. 
These models are not only tested against software engineering (SE) requirements but also propel the advancement of LLMs into practical production \cite{zhang2023unifying}.

\done{While recent surveys have shed light on code LLMs from the lenses of Natural Language Processing (NLP), Software Engineering (SE), or a combination of both disciplines \cite{zan2023large,zheng2023survey,zhang2023unifying,fan2023large,hou2024large,lyu2024automatic}, they have often encompassed a broad range of code-related tasks.} 
There remains a dearth of literature specifically reviewing advanced topics in code generation, such as meticulous data curation, instruction tuning, alignment with feedback, prompting techniques, the development of autonomous coding agents, retrieval augmented code generation, LLM-as-a-Judge for code generation, among others.
A notably pertinent study \cite{athiwaratkun2022multi,zan2023large} also concentrates on LLMs for text-to-code generation (NL2Code), yet it primarily examines models released from 2020 to 2022. 
Consequently, this noticeable temporal gap has resulted in an absence of up-to-date literature reviews that contemplate the latest advancements, including models like CodeQwen \cite{codeqwen}, WizardCoder \cite{luo2023wizardcoder}, CodeFusion \cite{singh2023codefusion}, and PPOCoder \cite{shojaee2023execution}, as well as the comprehensive exploration of the advanced topics previously mentioned.

\begin{table}[t] 
\caption{\done{
The applications of code LLMs in various code-related understanding and generation tasks. 
The I-O column indicates the type of input and output for each task, where C, NL, and K represent code, natural language, and label, respectively. 
Note that the detailed definitions of each task aligns with the descriptions in \cite{ahmad2021unified,austin2021program,niu2022deep,athiwaratkun2022multi,zan2023large}. 
The main scope of this survey focuses on code generation while it may involve code completion in Section \ref{sec:repository_level} and \ref{sec:retrieval_augmented}, aiming to illustrate the corresponding advancements.
}}
\label{tab:code_tasks}
\centering
\scalebox{0.77}{
    \begin{tabular}{clll} 
    \toprule
    \textbf{Type} & \textbf{I-O} & \textbf{Task} & \textbf{Definition}  \\
    \midrule
\multirow{12}*{Understanding} 
    & \multirow{8}*{C-K} & Code Classification & \makecell[l]{Classify code snippets based on functionality, purpose, or attributes \\ to aid in organization and analysis.} \\
    &  & Bug Detection & \makecell[l]{Detect and diagnose bugs or vulnerabilities in code to ensure \\ functionality and security.} \\
    &  & Clone Detection & \makecell[l]{Identifying duplicate or similar code snippets in software to enhance \\ maintainability, reduce redundancy, and check plagiarism.} \\
    &  & Exception Type Prediction & \makecell[l]{Predict different exception types in code to manage and handle \\ exceptions effectively.} \\
    \cmidrule{2-4}
    & C-C & Code-to-Code Retrieval & \makecell[l]{Retrieve relevant code snippets based on a given \\ code query for reuse or analysis.} \\ 
    \cmidrule{2-4}
    & NL-C & Code Search & \makecell[l]{Find relevant code snippets based on natural language \\queries to facilitate coding and development tasks.} \\
\midrule
\multirow{15}*{Generation} 
    & \cellcolor{gray!15} \multirow{11}*{C-C} & \cellcolor{gray!15}Code Completion & \cellcolor{gray!15} \makecell[l]{Predict and suggest the next portion of code, given contextual \\information from the prefix (and suffix), while typing to enhance \\ development speed and accuracy.} \\
    &  & Code Translation & \makecell[l]{Translate the code from one programming language to another \\while preserving functionality and logic.} \\
    &  & Code Repair & \makecell[l]{Identify and fix bugs in code by generating the correct version to \\improve functionality and reliability.} \\
    &  & Mutant Generation & \makecell[l]{Generate modified versions of code to test and evaluate the \\effectiveness of testing strategies.} \\
    &  & Test Generation & \makecell[l]{Generate test cases to validate code functionality, performance,\\ and robustness.} \\
    \cmidrule{2-4}
    & C-NL & Code Summarization & \makecell[l]{Generate concise textual descriptions or explanations of code to \\enhance understanding and documentation.} \\ 
    \cmidrule{2-4}
    & \cellcolor{yellow!40}NL-C & \cellcolor{yellow!40} Code Generation & \cellcolor{yellow!40} \makecell[l]{Generate source code from natural language descriptions to \\streamline development and reduce manual coding efforts.} \\
    \bottomrule
    \end{tabular}
}
\end{table}
Recognizing the need for a dedicated and up-to-date literature review, this survey endeavors to fill that void. We provide a systematic review that will serve as a foundational reference for researchers quickly exploring the latest progress in LLMs for code generation. 
A taxonomy is introduced to categorize and examine recent advancements, encompassing data curation \cite{wang2023self,luo2023wizardcoder,wei2023magicoder}, advanced topics \cite{parvez2021retrieval,lu2022reacc,le2022coderl,muennighoff2023octopack,liu2023rltf,chen2022codet,ni2023lever,chen2023teaching,huang2023agentcoder,shrivastava2023repofusion,zhang2023repocoder}, evaluation methods \cite{chen2021evaluating,hendrycks2021measuring,jimenez2023swe,zhuo2024ice}, and practical applications \cite{chen2021evaluating,zheng2023codegeex}. This category aligns with the comprehensive lifecycle of an LLM for code generation.
Furthermore, we pinpoint critical challenges and identify promising opportunities to bridge the research-practicality divide. Therefore, this survey allows NLP and SE researchers to seamlessly equip with a thorough understanding of LLM for code generation, highlighting cutting-edge directions and current hurdles and prospects.

The remainder of the survey is organized following the structure outlined in our taxonomy in Figure \ref{fig:taxonomy}. In Section \ref{sec:background}, we introduce the preliminaries of LLM with Transformer architecture and formulate the task of LLM for code generation. 
\done{Section \ref{sec:methodology}, we detail the systematic methodologies employed in conducting literature reviews.}
Then, in Section \ref{sec:taxonomy}, we propose a taxonomy, categorizing the complete process of LLMs in code generation. Section \ref{sec:overview} delves into the specifics of LLMs for code generation within this taxonomy framework. In Section \ref{sec:challenges}, we underscore the critical challenges and promising opportunities for bridging the research-practicality gap and conclude this work in Section \ref{sec:conclusion}. 

\section{Background}\label{sec:background}
\subsection{Large Language Models}\label{sec:llms}
The effectiveness of large language models (LLMs) is fundamentally attributed to their substantial quantity of model parameters, large-scale and diversified datasets, and the immense computational power utilized during training \cite{kaplan2020scaling,hoffmann2022training}. 
Generally, scaling up language models consistently results in enhanced performance and sample efficiency across a broad array of downstream tasks \cite{wei2022emergent,zhao2023survey}. 
However, with the expansion of the model size to a certain extent (e.g., GPT-3 \cite{brown2020language} with 175B-parameters and PaLM \cite{chowdhery2023palm} with 540B), LLMs have exhibited an unpredictable phenomenon known as emergent abilities\footnote{It should be noted that an LLM is not necessarily superior to a smaller language model, and emergent abilities may not manifest in all LLMs \cite{zhao2023survey}.}, including instruction following \cite{ouyang2022training}, in-context learning \cite{dong2022survey}, and step-by-step reasoning \cite{wei2022chain,huang2022towards}, which are absent in smaller models but apparent in larger ones \cite{wei2022emergent}.

\begin{figure}[tbp]
\begin{center}
\centerline{
\begin{minipage}[b]{1.0\linewidth}
    \subfigure[Encoder-Decoder Models]{\includegraphics[width=0.65\linewidth]{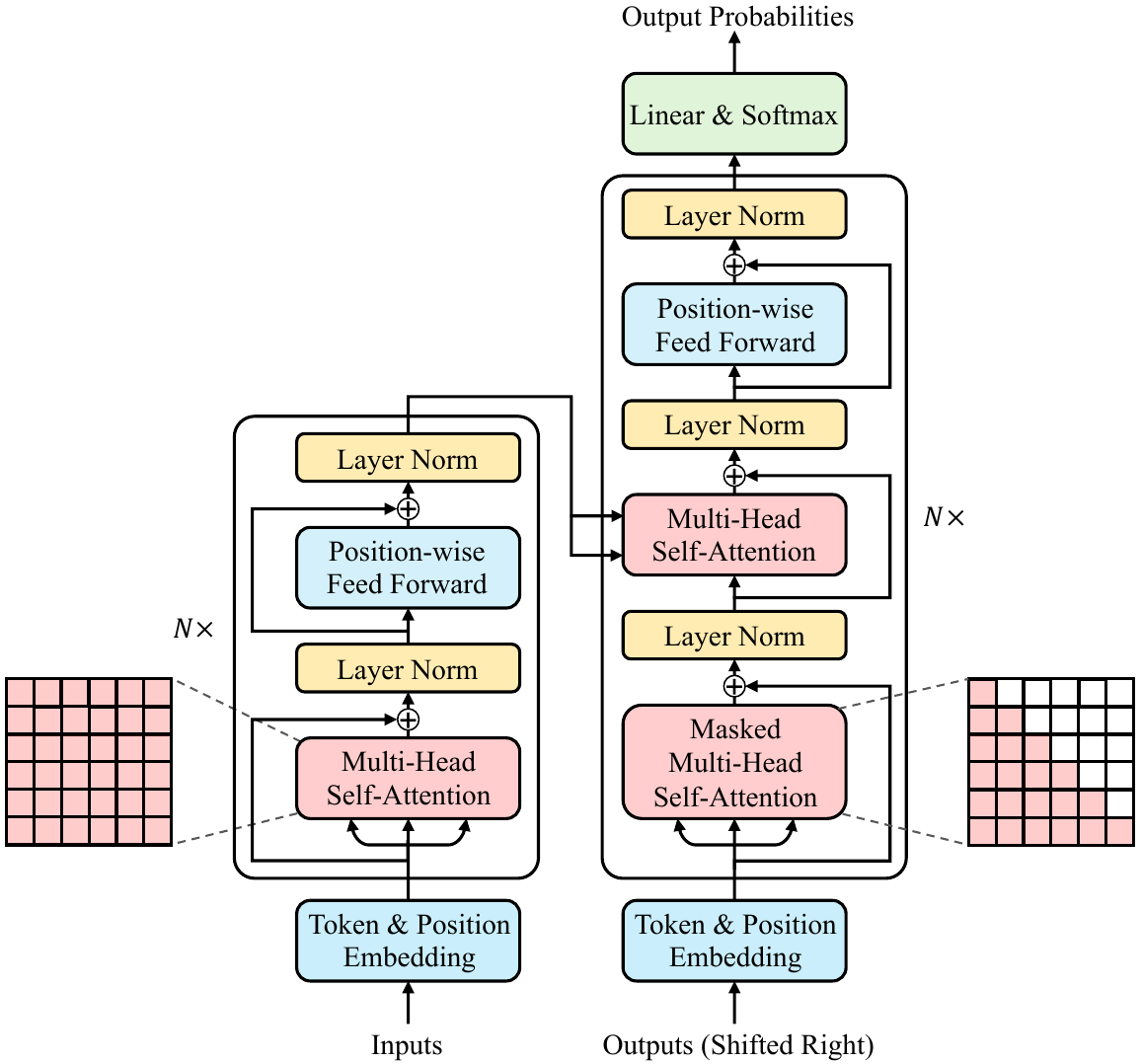}}
    \hspace{0.2in}
    \subfigure[Decoder-only Models]{\includegraphics[width=0.3\linewidth]{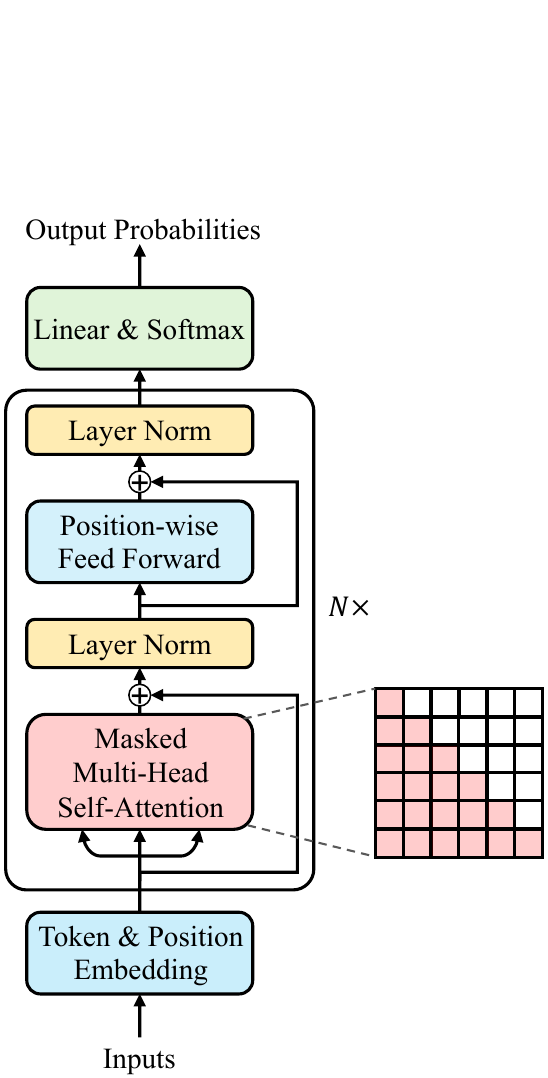}}
\end{minipage}}
\caption{The overview of large language models (LLMs) with encoder-decoder and decoder-only Transformer architecture for code generation, adapted from \cite{vaswani2017attention}.}
\label{fig:architecture}   
\end{center}
\end{figure}

Adhering to the same architectures of the Transformer \cite{vaswani2017attention} in LLMs, \done{code LLMs are specifically pre-trained (or continually pre-trained on general LLMs) using large-scale unlabeled code corpora with a smaller portion of text (and math) data, whereas general-purpose LLMs are pre-trained primarily on large-scale text data, incorporating a smaller amount of code and math data to enhance logical reasoning capabilities.
Additionally, some code LLMs, such as Qwen2.5-Coder \cite{hui2024qwen2}, incorporate synthetic data in their training processes, a practice that is attracting increasing attention from both industry and academia.}
Analogous to LLMs, Code LLMs can also be classified into three architectural categories: encoder-only models, decoder-only models, and encoder-decoder models. 
For encoder-only models, such as CodeBERT \cite{feng2020codebert}, they are typically suitable for code comprehension tasks including type prediction, code retrieval, and clone detection. 
For decoder-only models, such as StarCoder \cite{brown2020language}, they predominantly excel in generation tasks, such as code generation, code translation, and code summarization. 
Encoder-decoder models, such as CodeT5 \cite{wang2021codet5}, can accommodate both code understanding and generation tasks but do not necessarily outperform encoder-only or decoder-only models. 
The overall architectures of the different Code LLMs for code generation are depicted in Figure \ref{fig:architecture}.

In the following subsection, we will delineate the key modules of the Transformer layers in Code LLMs.

\subsubsection{Multi-Head Self-Attention Modules}
Each Transformer layer incorporates a multi-head self-attention (MHSA) mechanism to discern the inherent semantic relationships within a sequence of tokens across $h$ distinct latent representation spaces. 
Formally, the MHSA employed by the Transformer can be formulated as follows:
\begin{equation}\label{eq:multihead}
\begin{aligned}
    \mathbf{h}^{(l)}=\operatorname{MultiHeadSelfAttn}(\mathbf{Q},\mathbf{K},\mathbf{V}) =\operatorname{Concat}\left\{\mathrm{Head}_i\right\}_{i=1}^h\mathbf{W^O},
\end{aligned} 
\end{equation}
\begin{equation}
\begin{aligned}
    \operatorname{Head}_i =\operatorname{Attention}(\underbrace{\mathbf{H}^{(l-1)}\mathbf{W}_i^\mathbf{Q}}_{\mathbf{Q}},\underbrace{\mathbf{H}^{(l-1)}\mathbf{W}_i^\mathbf{K}}_{\mathbf{K}}, \underbrace{\mathbf{H}^{(l-1)}\mathbf{W}_i^\mathbf{V}}_\mathbf{V}), 
\end{aligned} 
\end{equation}
\begin{equation}\label{eq:attention}
\begin{aligned}
    \operatorname{Attention}(\mathbf{Q}, \mathbf{K}, \mathbf{V})=\operatorname{softmax}\left(\frac{\mathbf{Q}\mathbf{K}^T}{\sqrt{d_{model}/h}}\right)\mathbf{V},
\end{aligned}
\end{equation}
where $\mathbf{H}^{(l-1)} \in \mathbb{R}^{n\times d_{model}}$ denotes the input to the $l$-\textit{th} Transformer layer, while $\mathbf{h}^{(l)} \in \mathbb{R}^{n\times d_{model}}$ represents the output of MHSA sub-layer.  
The quantity of distinct attention heads is represented by $h$, and $d_{model}$ refers to the model dimension. 
\done{The set of projections $\left\{\mathbf{W}_i^\mathbf{Q}, \mathbf{W}_i^\mathbf{K}, \mathbf{W}_i^\mathbf{V}, \mathbf{W}_i^\mathbf{O}\right\} \in \mathbb{R}^{d_{model} \times d_{model}/ h}$ encompasses the affine transformation parameters for each attention head $\operatorname{Head}_i$, transforming the Query $\mathbf{Q}$, Key $\mathbf{K}$, Value $\mathbf{V}$, and the output of the attention sub-layer.}
The $\operatorname{softmax}$ function is applied in a row-wise manner. 
The dot-products of queries and keys are divided by a scaling factor $\sqrt{d_{model}/h}$ to counteract the potential risk of excessive large inner products and correspondingly diminished gradients in the $\operatorname{softmax}$ function, thus encouraging a more balanced attention landscape.

In addition to multi-head self-attention, there are two other types of attention based on the source of queries and key-value pairs:
\begin{itemize}
    \item \textbf{Masked Multi-Head Self-Attention}. 
    Within the decoder layers of the Transformer, the self-attention mechanism is constrained by introducing an attention mask, ensuring that queries at each position can only attend to all key-value pairs up to and inclusive of that position. 
    To facilitate parallel training, this is typically executed by assigning a value of 0 to the lower triangular part and setting the remaining elements to $-\infty$. Consequently, each item attends only to its predecessors and itself. Formally, this modification in Equation \ref{eq:attention} can be depicted as follows:
    \begin{equation}\label{eq:attention_modified}
    \begin{aligned}
        \operatorname{Attention}(\mathbf{Q}, \mathbf{K}, \mathbf{V})=\operatorname{softmax}\left(\frac{\mathbf{Q}\mathbf{K}^T}{\sqrt{d_{model}/h}} + \mathbf{M}_{mask} \right)\mathbf{V},
    \end{aligned}
    \end{equation}
    \begin{equation}
    \begin{aligned}
        \mathbf{M}_{mask} = \Big(m_{ij}\Big)_{n\times n} = \Big(\mathbb{I}(i\ge j)\Big)_{n\times n} = 
        \begin{cases}
        0 & \text{for $i \ge j$ } \\
        -\infty & \text{otherwise}
        \end{cases},
    \end{aligned}
    \end{equation}
    This form of self-attention is commonly denoted as autoregressive or causal attention \cite{lin2022survey}.
    \item \textbf{Cross-Layer Multi-Head Self-Attention}. 
    The queries are derived from the outputs of the preceding (decoder) layer, while the keys and values are projected from the outputs of the encoder.
\end{itemize}

\subsubsection{Position-wise Feed-Forward Networks}
Within each Transformer layer, a Position-wise Feed-Forward Network (PFFN) is leveraged following the MHSA sub-layer to refine the sequence embeddings at each position $i$ in a separate and identical manner, thereby encoding more intricate feature representations. 
The PFFN is composed of a pair of linear transformations, interspersed with a ReLU activation function. Formally,
\begin{equation}
\begin{aligned}
\operatorname{PFFN}(h^{(l)})=\left(\operatorname{Concat}\left\{\operatorname{FFN}(h^{(l)}_i)^T\right\}_{i=1}^{n}\right)^T,
\end{aligned}   
\end{equation}
\begin{equation}
\begin{aligned}
\operatorname{FFN}(h^{(l)}_i)=\operatorname{ReLU}(h^{(l)}_i\mathbf{W}^{(1)}+b^{(1)})\mathbf{W}^{(2)}+b^{(2)},
\end{aligned} 
\end{equation}
where $h^{(l)} \in \mathbb{R}^{n\times d_{model}}$ is the outputs of MHSA sub-layer in $l$-th Transformer layer, and $h^{(l)}_i \in \mathbb{R}^{d_{model}}$ denotes the latent representation at each sequence position. The projection matrices $\left\{\mathbf{W}^{(1)}, (\mathbf{W}^{(2)})^T \right\} \in \mathbb{R}^{d_{model} \times 4d_{model}}$ and bias vectors $\{\mathbf{b}^{(1)}, \mathbf{b}^{(2)}\} \in \mathbb{R}^{d_{model}}$ are parameters learned during training. These parameters remain consistent across all positions while are individually initialized from layer to layer. In this context, $T$ represents the transpose operation on a matrix.

\subsubsection{Residual Connection and Normalization}
To alleviate the issue of vanishing or exploding gradients resulting from network deepening, the Transformer model incorporates a residual connection \cite{he2016deep} around each of the aforementioned modules, followed by Layer Normalization \cite{ba2016layer}. 
For the placement of Layer Normalization operation, there are two widely used approaches:
1) \textbf{Post-Norm}: Layer normalization is implemented subsequent to the element-wise residual addition, in accordance with the vanilla Transformer \cite{vaswani2017attention}.
2) \textbf{Pre-Norm}: Layer normalization is applied to the input of each sub-layer, as seen in models like GPT-2 \cite{radford2019language}. 
Formally, it can be formulated as:
\begin{equation}
\begin{aligned}
\textbf{Post-Norm}: 
    \mathbf{H^{(l)}} &=\operatorname{LayerNorm}(\operatorname{PFFN}(\mathbf{h^{(l)}})+\mathbf{h^{(l)}}),\\
    \mathbf{h^{(l)}}&=\operatorname{LayerNorm}(\operatorname{MHSA}(\mathbf{H^{(l-1)}})+\mathbf{H^{(l-1)}}) 
\end{aligned}
\end{equation}
\begin{equation}
\begin{aligned}
\textbf{Pre-Norm}: 
    \mathbf{H^{(l)}} &=\operatorname{PFFN}(\operatorname{LayerNorm}(\mathbf{h^{(l)}}))+\mathbf{h^{(l)}},\\
    \mathbf{h^{(l)}}&=\operatorname{MHSA}(\operatorname{LayerNorm}(\mathbf{H^{(l-1)}}))+\mathbf{H^{(l-1)}} 
\end{aligned}
\end{equation}
\subsubsection{Positional Encoding}
Given that self-attention alone cannot discern the positional information of each input token, the vanilla Transformer introduces an absolute positional encoding method to supplement this positional information, known as sinusoidal position embeddings \cite{vaswani2017attention}. 
Specifically, for a token at position $pos$, the position embedding is defined as:
\begin{equation}\label{eq:sin}
\begin{aligned}
    \mathbf{p}_{pos,2i}=\sin(\frac{pos}{10000^{2i/d_{model}}}),
\end{aligned}
\end{equation}
\begin{equation}\label{eq:cos}
\begin{aligned}
    \mathbf{p}_{pos,2i+1}=\cos(\frac{pos}{10000^{2i/d_{model}}}), 
\end{aligned}
\end{equation}
where $2i, 2i+1$ represent the dimensions of the position embedding, while $d_{model}$ denotes the model dimension. Subsequently, each position embedding is added to the corresponding token embedding, and the sum is fed into the Transformer. 
Since the inception of this method, a variety of innovative positional encoding approaches have emerged, such as learnable embeddings \cite{devlin2018bert}, relative position embeddings \cite{shaw2018self}, RoPE \cite{su2024roformer}, and ALiBi \cite{press2021train}. 
For more detailed descriptions of each method, please consult \cite{lin2022survey,zhao2023length}.

\subsubsection{\done{Architecture}}
\done{
There are two types of Transformer architecture for code generation task, including encoder-decoder and decoder-only. 
For the encoder-decoder architecture, it consists of both an encoder and a decoder, in which the encoder processes the input data and generates a set of representations, which are then used by the decoder to produce the output.
However, for decoder-only architecture, it consists only of the decoder part of the transformer, where it uses a single stack of layers to both process input data and generate output.
Therefore, the encoder-decoder architecture is suited for tasks requiring mapping between different input and output domains, while the decoder-only architecture is designed for tasks focused on sequence generation and continuation.
The overview of LLMs with these two architectures are illustrated in Figure \ref{fig:architecture}.
}

\subsection{Code Generation}
Large language models (LLMs) for code generation refer to the use of LLM to generate source code from natural language descriptions, a process also known as a natural-language-to-code task.
Typically, these natural language descriptions encompass programming problem statements (or docstrings) and may optionally include some programming context (e.g., function signatures, assertions, etc.). 
Formally, these natural language (NL) descriptions can be represented as $\mathbf{x}$.
Given $\mathbf{x}$, the use of an LLM with model parameters $\theta$ to generate a code solution $\mathbf{y}$ can be denoted as $P_{\theta}(\mathbf{y}\mid\mathbf{x})$.
The advent of in-context learning abilities in LLM \cite{wei2022emergent} has led to the appending of exemplars to the natural language description $\mathbf{x}$ as demonstrations to enhance code generation performance or constrain the generation format \cite{li2023towards,patel2023evaluating}. 
A fixed set of $M$ exemplars is denoted as $\{(\mathbf{x_i}, \mathbf{y_i})\}_{i=1}^M$. 
Consequently, following \cite{ni2023lever}, a more general formulation of LLMs for code generation with few-shot (or zero-shot) exemplars can be revised as:
\begin{equation}
\begin{aligned}
   P_\theta(\mathbf{y}\mid\mathbf{x}) \Rightarrow P_\theta(\mathbf{y}\mid\operatorname{prompt}(\mathbf{x}, \{(\mathbf{x_i}, \mathbf{y_i})\}_{i=1}^k)), k\in\{0, 1, \dots, M\}
\end{aligned}
\end{equation}
where $\operatorname{prompt}(\mathbf{x}, \{(\mathbf{x_i}, \mathbf{y_i})\}_{i=1}^k))$ is a string representation of the overall input, and $\{(\mathbf{x_i}, \mathbf{y_i})\}_{i=1}^k$ denotes a set of $k$ exemplars randomly selected from $\{(\mathbf{x_i}, \mathbf{y_i})\}_{i=1}^M$. 
In particular, when $k=0$, this denotes zero-shot code generation, equivalent to vanilla ones without in-context learning.  
\done{In the decoding process}, a variety of decoding strategies can be performed for code generation, including deterministic-based strategies (e.g., greedy search and beam search) and sampling-based strategies (e.g., temperature sampling, top-k sampling, and top-p (nucleus) sampling). 
For more detailed descriptions of each decoding strategy, please consult \cite{holtzman2019curious}. \done{For example, the greedy search and sampling-based decoding strategies can be formulated as follows:}
\begin{equation}
\begin{aligned}
\textbf{Greedy Search}:
   \mathbf{y^*} = \mathop{\mathrm{argmax}}_\mathbf{y} P_\theta(\mathbf{y}\mid\operatorname{prompt}(\mathbf{x}, \{(\mathbf{x_i}, \mathbf{y_i})\}_{i=1}^k)), k\in\{0, 1, \dots, M\}
\end{aligned}
\end{equation}
\begin{equation}
\begin{aligned}
\textbf{Sampling}: 
   \mathbf{y} \sim P_\theta(\mathbf{y}\mid\operatorname{prompt}(\mathbf{x}, \{(\mathbf{x_i}, \mathbf{y}_i)\}_{i=1}^k)), k\in\{0, 1, \dots, M\}
\end{aligned}
\end{equation}

\done{To verify the functionality correctness of the generated code solution, $\mathbf{y}$ is subsequently executed via a compiler or interpreter, represented as $\mathbf{Exe}(\cdot)$, on a suit of unit tests $\mathbf{\mathcal{T}}$. 
The feedback from this execution can be denoted as $\mathbf{Feedback}(\mathbf{Exe}(\mathbf{y},\mathbf{\mathcal{T}}))$. 
If the generated code solution fails to pass all test cases, the error feedback can be iteratively utilized to refine the code by leveraging the previous attempt ($\mathbf{y}_{pre}$) and the associated feedback. Formally,
\begin{equation}
\begin{aligned}
    \mathbf{y} \sim P_\theta(\mathbf{y}\mid\operatorname{prompt}(\mathbf{x}, \{(\mathbf{x_i}, \mathbf{y_i})\}_{i=1}^k, \mathbf{y}_{pre}, \mathbf{Feedback}(\mathbf{Exe}(\mathbf{y},\mathbf{\mathcal{T}})))), k\in\{0, 1, \dots, M\}
\end{aligned}
\end{equation}
Further details and relevant studies on using feedback to improve code generation are comprehensively discussed in Section \ref{sec:reinforcement_learning} and \ref{sec:prompting}.
}

\done{\section{Methodology}\label{sec:methodology}
\begin{figure}[t]
\centering
\includegraphics[width=1\linewidth]{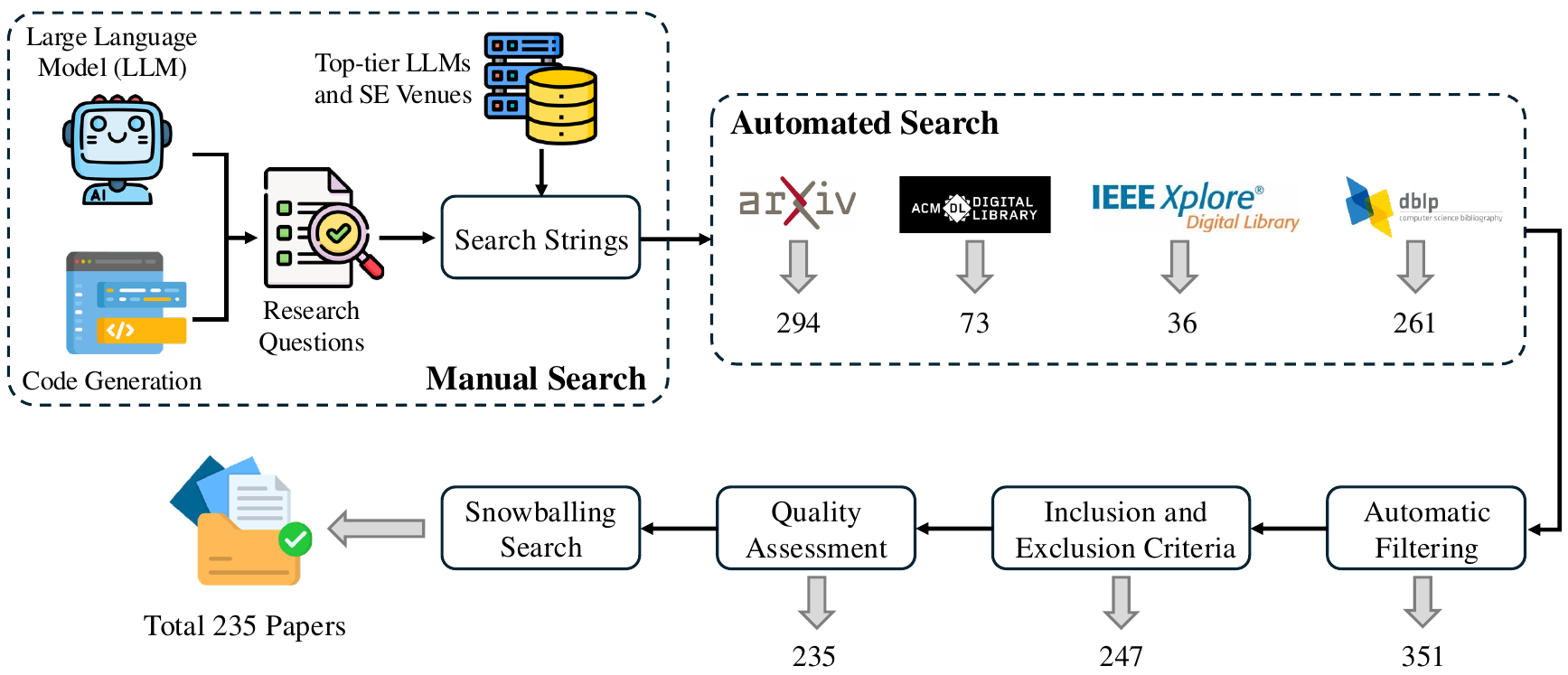}
\caption{\done{Overview of the paper search and collection process.}}
\label{fig:review_process}
\end{figure}

In this section, we detail the systematic methodologies employed in conducting literature reviews. We follow the systematic literature review methodology outlined by \cite{kitchenham2009systematic}, which has been widely adopted in numerous software engineering literature reviews \cite{hou2024large,li2017static,liu2022deep,ramirez2018systematic,wang2022machine}. 
The overall process is illustrated in Figure \ref{fig:review_process}, and the detailed steps in our methodology are documented below.

\subsection{Research Questions}
To deliver a comprehensive and up-to-date literature review on the latest advancements in large language models (LLMs) for code generation, this systematic literature review addresses the following research questions (RQs):

\textbf{RQ1: How can we categorize and evaluate the latest advances in LLMs for code generation?}
The recent proliferation of LLMs has resulted in many of these models being adapted for code generation task. 
While the adaptation of LLMs for code generation essentially follows the evolution of LLMs, this evolution encompasses a broad spectrum of research directions and advancements. 
For software engineering (SE) researchers, it can be challenging and time-consuming to fully grasp the comprehensive research landscape of LLMs and their adaptation to code generation. 
RQ1 aims to propose a taxonomy that serves as a comprehensive reference for researchers, enabling them to quickly familiarize themselves with the state-of-the-art in this dynamic field and identify specific research problems and directions of interest.

\textbf{RQ2: What are the key insights into LLMs for code generation?}
RQ2 seeks to assist researchers in establishing a comprehensive, up-to-date, and advanced understanding of LLMs for code generation. 
This includes discussing various aspects of this rapidly evolving domain, such as data curation, latest advancements, performance evaluation, ethical and environmental implications, and real-world applications. 
A historical overview of the evolution of LLMs for code generation is provided, along with an empirical comparison using the widely recognized HumanEval and MBPP benchmarks, as well as the more practical and challenging BigCodeBench benchmark, to highlight the progressive enhancements in LLM capabilities for code generation.
RQ2 offers an in-depth analysis of critical insights related to LLMs for code generation.

\textbf{RQ3: What are the critical challenges and promising research opportunities in LLMs for code generation?}
Despite the revolutionary impact of LLMs on the paradigm of code generation and their remarkable performance, numerous challenges remain unaddressed. 
These challenges primarily stem from the gap between academic research and practical development. 
For instance, while the HumanEval benchmark is established as a de facto standard for evaluating the coding proficiency of LLMs in academia, it has been shown that this evaluation does not adequately reflect practical development scenarios \cite{jimenez2023swe,du2024evaluating,liu2024your,ding2024crosscodeeval}. 
RQ3 aims to identify critical challenges and highlight promising opportunities to bridge the gap between research and practical application.

\subsection{Search Process}
\begin{table}[t] 
\centering
\caption{\done{Publication venues for conference proceedings and journals articles for manual search.}}
\label{tab:venues}
\scalebox{0.7}{
\begin{tabular}{lll}
\toprule
\textbf{Domain} & \textbf{Venue} & \textbf{Acronym} \\ 
\midrule
\multirow{7}{*}{\textbf{LLMs}} & International Conference on Learning Representations & ICLR \\
 & Conference on Neural Information Processing Systems & NeurIPS \\
 & International Conference on Machine Learning & ICML \\
 & Annual Meeting of the Association for Computational Linguistics & ACL \\
 & Conference on Empirical Methods in Natural Language Processing & EMNLP \\
 & International Joint Conference on Artificial Intelligence & NAACL \\
 & AAAI Conference on Artificial Intelligence & AAAI \\
\midrule
\multirow{6}{*}{\textbf{SE}} & International Conference on Software Engineering & ICSE \\
 & Joint European Software Engineering Conference and Symposium on the Foundations of Software Engineering & ESEC/FSE  \\
 & International Conference on Automated Software Engineering & ASE  \\
 & Transactions on Software Engineering and Methodology & TOSEM \\
 & Transactions on Software Engineering & TSE \\
 & International Symposium on Software Testing and Analysis & ISSTA \\
\bottomrule
\end{tabular}
}
\end{table}
\begin{table}[t] 
\centering
\caption{\done{Keywords related to LLMs and code generation task for automated search.}}
\label{tab:keywords}
\scalebox{0.7}{
\begin{tabular}{l|l}
\toprule
\textbf{Keywords Related to LLMs} & \textbf{Keywords Related to Code Generation Task} \\ 
\midrule
\makecell[l]{
Code Large Language Model$^*$, Code LLMs, Code Language Model, \\
Code LMs, Large Language Model$^*$, LLM, Language Model$^*$, LM, \\
Pre-trained Language Model$^*$, PLM, Pre-trained model, \\ 
Natural Language Processing, NLP, GPT-3, ChatGPT, GPT-4, LLaMA, \\CodeLlama, PaLM$^*$,  
CodeT5, Codex, CodeGen, InstructGPT
} & 
\makecell[l]{Code Generation, Program Synthesis, Code Intelligence, \\
$^*$Coder$^*$, natural-language-to-code, NL2Code, Programming} \\
\bottomrule
\end{tabular}
}
\end{table}

\subsubsection{Search Strings}
To address the aforementioned three research questions (RQs), we initiate a manual review of conference proceedings and journal articles from top-tier venues in the fields of LLMs and SE, as detailed in Table \ref{tab:venues}. 
This process allowed us to identify relevant studies and derive search strings, which are subsequently utilized for an automated search across various scientific databases. The complete set of search keywords is presented in Table \ref{tab:keywords}.


\subsubsection{Search Databases}
Following the development of search strings, we executed an automated search using four popular scientific databases: the ACM Digital Library, IEEE Xplore Digital Library, arXiv, and DBLP. 
Our search focus on identifying papers whose titles contain keywords pertinent to LLMs and code generation. 
This approach enhances the likelihood of retrieving relevant papers since both sets of keywords must be present in the title. 
Although this title-based search strategy effectively retrieves a large volume of papers, it is important to note that in some instances \cite{shojaee2023execution}, the scope of code generation can be broader, encompassing areas such as code completion, code translation, and program synthesis. 
As outlined in Section \ref{sec:introduction}, this survey adopts a prevalent definition of code generation as the natural-language-to-code (NL2Code) task \cite{austin2021program,athiwaratkun2022multi,zan2023large}.

Consequently, we conduct further automatic filtering based on the content of the papers. 
Papers focusing on ``code completion'' and ``code translation'' are excluded unless they pertain to the specific topics discussed in Section \ref{sec:repository_level} and Section \ref{sec:retrieval_augmented}, where code completion is a primary focus. 
After completing the automated search, the results from each database are merged and deduplicated using scripts. 
This process yields 294 papers from arXiv, 73 papers from the ACM Digital Library, 36 papers from IEEE Xplore, and 261 papers from DBLP.

\subsection{Inclusion and Exclusion Criteria}\label{sec:review_criteria}
The search process conducted across various databases and venues is intentionally broad to gather a comprehensive pool of candidate papers. 
This approach maximizes the collection of potentially relevant studies. 
However, such inclusivity may lead to the inclusion of papers that do not align with the scope of this survey, as well as duplicate entries from multiple sources. 
To address this, we have established a clear set of inclusion and exclusion criteria, based on the guidelines from \cite{hou2024large,wang2024software}. 
These criteria are applied to each paper to ensure alignment with our research scope and questions, and to eliminate irrelevant studies.

\textbf{Inclusion Criteria.} 
A paper will be included if it meets any of the following criteria:
\begin{itemize}
    \item It is available in full text.
    \item It presents a dataset or benchmark specifically designed for code generation with LLMs.
    \item It explores specific LLM techniques, such as pre-training or instruction tuning, for code generation.
    \item It provides an empirical study or evaluation related to the use of LLMs for code generation.
    \item It discusses the ethical considerations and environmental impact of deploying LLMs for code generation.
    \item It proposes tools or applications powered by LLMs for code generation.
\end{itemize}

\textbf{Exclusion Criteria.} 
Conversely, papers will be excluded if they meet any of the following conditions:
\begin{itemize}
    \item They are not written in English.
    \item They are found in books, theses, monographs, keynotes, panels, or venues (excluding arXiv) that do not undergo a full peer-review process.
    \item They are duplicate papers or different versions of similar studies by the same authors.
    \item They focus on text generation rather than source code generation, such as generating code comments, questions, test cases, or summarization.
    \item They do not address the task of code generation, for instance, focusing on code translation instead.
    \item They leverage software engineering methods to enhance code generation without emphasizing LLMs.
    \item They do not utilize LLMs, opting for other models like Long Short-Term Memory (LSTM) networks.
    \item They use encoder-only language models, such as BERT, which are not directly applicable to code generation task.
    \item LLMs are mentioned only in future work or discussions without being central to the proposed approach.
\end{itemize}

Papers identified through both manual and automated searches undergo a detailed manual review to ensure they meet the inclusion criteria and do not fall under the exclusion criteria. 
Specifically, the first two authors independently review each paper to determine its eligibility. 
In cases of disagreement, the third author makes the final inclusion decision.

\subsection{Quality Assessment}\label{sec:review_quality}
To ensure the inclusion of high-quality studies, we have developed a comprehensive set of ten Quality Assessment Criteria (QAC) following \cite{hou2024large}. 
These QAC are designed to evaluate the relevance, clarity, validity, and significance of the papers considered for our review.

In accordance with \cite{hou2024large}, the first three QAC assess the study’s alignment with our objectives. These criteria are rated as ``irrelevant/unmet'', ``partially relevant/met'', or ``relevant/fully met'', corresponding to scores of -1, 0, and 1, respectively. 
If a study receive a score of -1 across these initial three criteria, it is deemed ineligible for further consideration and subsequently excluded from our review process.

The subsequent seven QAC focus on a more detailed content evaluation, employing a scoring range of -1 to 2, representing ``poor'', ``fair'', ``good'', and ``excellent''. 
We compute a cumulative score based on the responses to QAC4 through QAC10 for each paper. 
For published works, the maximum achievable score is 14 (2 points per question). We retain those with a score of 11.2 (80\% of the total score) or higher. 
For unpublished papers available on arXiv, QAC4 defaults to a score of 0, making the maximum possible score for the remaining criteria 12. Accordingly, we retain papers scoring 9.6 (80\% of the adjusted total score) or above .
\begin{itemize} 
    \item QAC1: Is the research not classified as a secondary study, such as a systematic literature review or survey? (-1, 0, 1) 
    \item QAC2: Does the study incorporate the use of LLMs? (-1, 0, 1) 
    \item QAC3: Is the study relevant to the code generation task? (-1, 0, 1) 
    \item QAC4: Is the research published in a prestigious venue? (-1, 0, 1, 2) 
    \item QAC5: Does the study present a clear research motivation? (-1, 0, 1, 2) 
    \item QAC6: Are the key contributions and limitations of the study discussed? (-1, 0, 1, 2) 
    \item QAC7: Does the study contribute to the academic or industrial community? (-1, 0, 1, 2) 
    \item QAC8: Are the LLM techniques employed in the study clearly described? (-1, 0, 1, 2) 
    \item QAC9: Are the experimental setups, including experimental environments and dataset information, thoroughly detailed? (-1, 0, 1, 2) 
    \item QAC10: Does the study clearly confirm its experimental findings? (-1, 0, 1, 2) 
\end{itemize}

\subsection{Snowballing Search}\label{sec:review_snowballing}
Following the quality assessment, we establish an initial set of papers for our study. 
To minimize the risk of excluding pertinent literature, we implement a snowballing search strategy. Snowballing search involves utilizing a paper's reference list or its citations to discover additional relevant studies, known as backward and forward snowballing, respectively. 
In this survey, we exclusively employed backward snowballing following \cite{wang2024software}.
Despite this effort, no additional studies are identified through this method. This could be attributed to the task-specific nature of the code generation (natural-language-to-code), where reference studies are typically published earlier. 
Consequently, our methodology, which encompassed an extensive manual and automated search, likely covered the relevant literature comprehensively, explaining the lack of additional studies through snowballing search.

\begin{figure}[t]
\centering
\includegraphics[width=\linewidth]{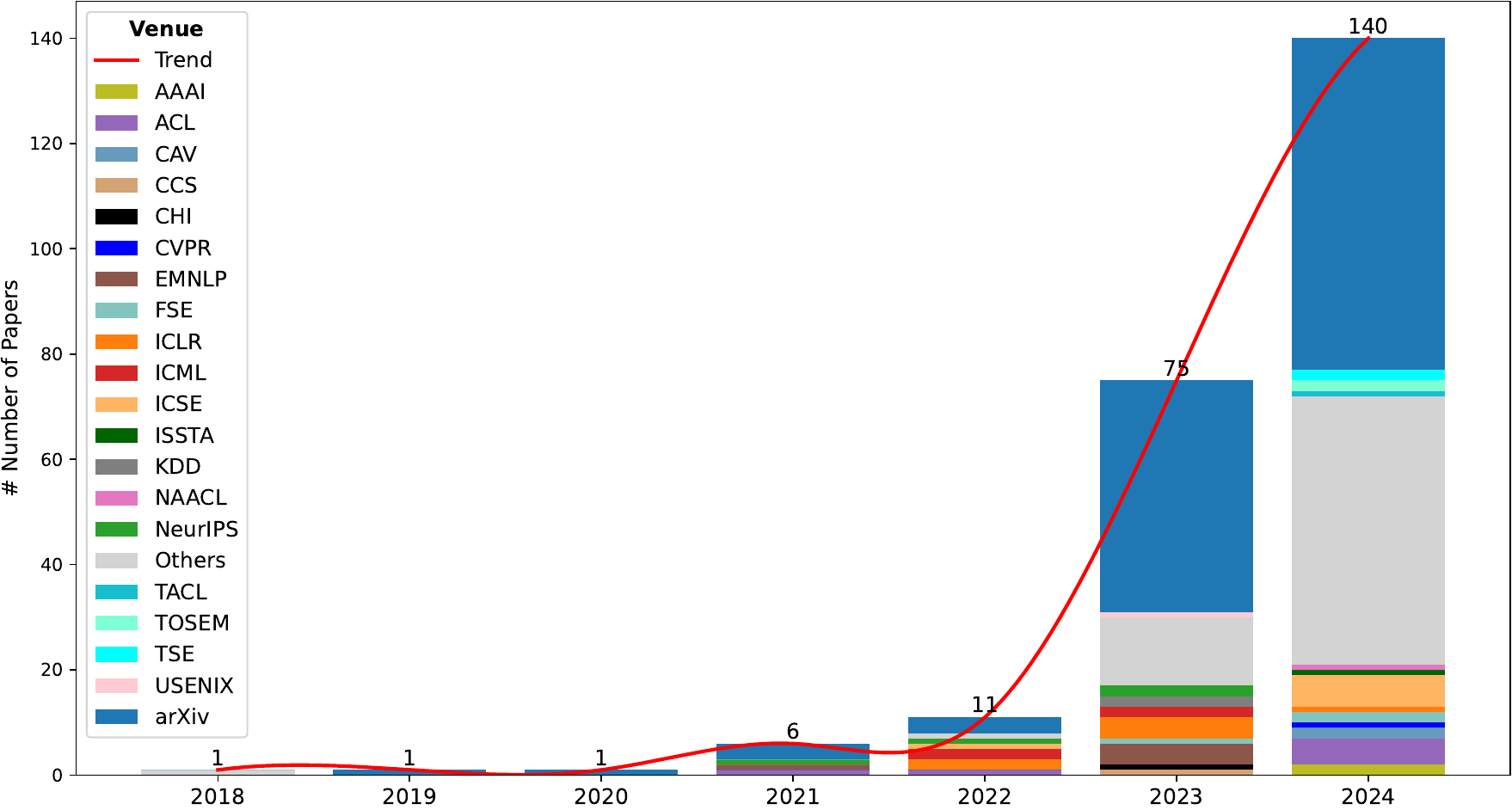}
\includegraphics[width=\linewidth]{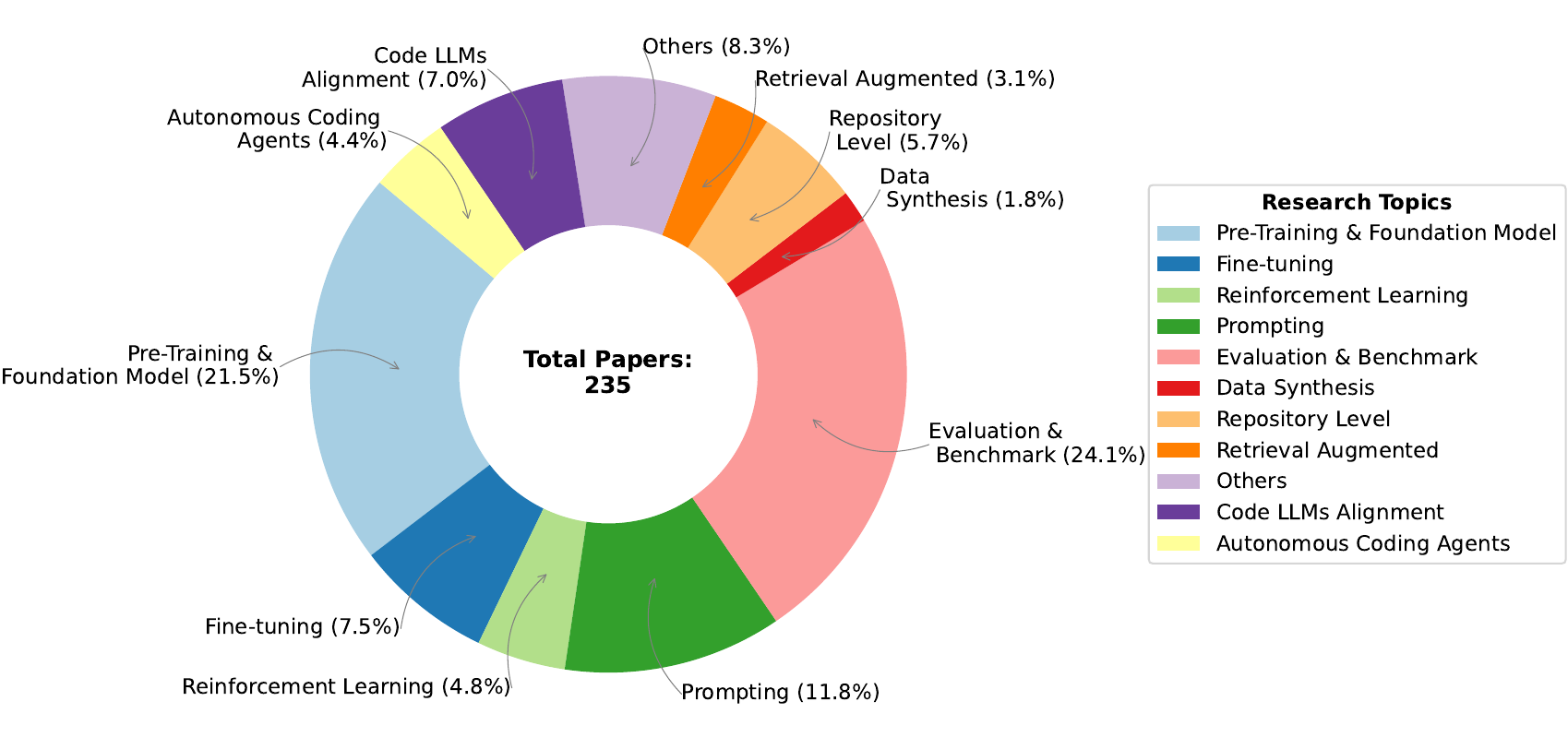}
\caption{\done{Data qualitative analysis. \textbf{Top}: Annual distribution of selected papers across various publication venues. \textbf{Bottom}: Distribution analysis of research topics covered in the included papers.}}
\label{fig:venues_num_topic_dist}
\end{figure}

\subsection{Data Collection and Analysis}
The data collection process for our study, illustrated in Figure \ref{fig:review_process}, began with a manual search through conference proceedings and journal articles from leading venues in LLMs and SE. This initial step yielded 42 papers, from which we extracted relevant search strings. Following this, we performed an automated search across four academic databases using keyword-based queries, resulting in the retrieval of 664 papers. After performing automatic filtering (351 papers), applying inclusion and exclusion criteria (247 papers), conducting quality assessments (235 papers), and utilizing snowballing search (235 papers), we finalize a collection of 235 papers focusing on LLMs for code generation.

To provide insights from the selected papers, we begin by presenting an overview of their distribution across publication venues each year, as illustrated at the top of Figure \ref{fig:venues_num_topic_dist}. 
Our analysis indicates that 14\% of the papers are published in LLM-specific venues and 7\% in SE venues. Remarkably, 49\% of the papers remain unpublished in peer-reviewed venues and are available on arXiv. 
This trend is understandable given the emerging nature of this field, with many works being recent and pending formal submission. 
Despite the absence of peer review on arXiv, our quality assessment process ensures that only high-quality papers are included, thereby maintaining the integrity of this survey.
Furthermore, the annual trend in the number of collected papers indicates nearly exponential growth in the field. From a single paper in the period 2018 to 2020, the numbers increased to 6 in 2021, 11 in 2022, 75 in 2023, and 140 in 2024. This trend reflects growing interest and attention in this research area, with expectations for continued expansion in the future.
Additionally, to capture the breadth of advancements in LLMs for code generation, we conducted a distribution analysis of the research topics covered in the included papers, as shown at the bottom of Figure \ref{fig:venues_num_topic_dist}. We observe that the development of LLMs for code generation closely aligns with broader trends in general-purpose LLM research. Notably, the most prevalent research topics are Pre-training and Foundation Models (21.5\%), Prompting (11.8\%), and Evaluation and Benchmarks (24.1\%). These areas hold significant promise for enhancing, refining, and evaluating LLM-driven code generation.

}
\section{Taxonomy}\label{sec:taxonomy}
The recent surge in the development of \done{LLMs} has led to a significant number of these models being repurposed for code generation task through continual pre-training or fine-tuning. 
This trend is particularly observable in the realm of open-source models. 
For instance, Meta AI initially made the LLaMA \cite{touvron2023llama} model publicly available, which was followed by the release of Code Llama \cite{roziere2023code}, designed specifically for code generation. 
Similarly, DeepSeek LLM \cite{bi2024deepseek} developed and released by DeepSeek has been extended to create DeepSeek Coder \cite{guo2024deepseek}, a variant tailored for code generation. 
The Qwen team has developed and released Code Qwen \cite{codeqwen}, building on their original Qwen \cite{bai2023qwen} model. 
Microsoft, on the other hand, has unveiled WizardLM \cite{xu2023wizardlm} and is exploring its coding-oriented counterpart, WizardCoder \cite{luo2023wizardcoder}. 
Google has joined the fray by releasing Gemma \cite{team2024gemma}, subsequently followed by Code Gemma \cite{codegemma_2024}.
Beyond simply adapting general-purpose LLMs for code-related tasks, there has been a proliferation of models specifically engineered for code generation. Notable examples include StarCoder \cite{li2023starcoder}, OctoCoder \cite{muennighoff2023octopack}, and CodeGen \cite{nijkamp2022codegen}. These models underscore the trend of LLMs being developed with a focus on code generation.

Recognizing the importance of these developments, 
\done{we conduct a thorough analysis of selected papers on LLMs for code generation, sourced from widely used scientific databases as mentioned in Section \ref{sec:methodology}. 
Based on this analysis, we propose a taxonomy that categorizes and evaluates the latest advancements in LLMs for code generation.}
This taxonomy, depicted in Figure \ref{fig:taxonomy}, serves as a comprehensive reference for researchers seeking to quickly familiarize themselves with the state-of-the-art in this dynamic field.
\done{It is important to highlight that the category of recent advances emphasizes the core techniques used in the current state-of-the-art code LLMs.}

In the subsequent sections, we will provide an in-depth analysis of each category related to code generation. This will encompass a definition of the problem, the challenges to be addressed, and a comparison of the most prominent models and their performance evaluation.

\begin{figure}[t]
\centering
\includegraphics[width=0.95\linewidth]{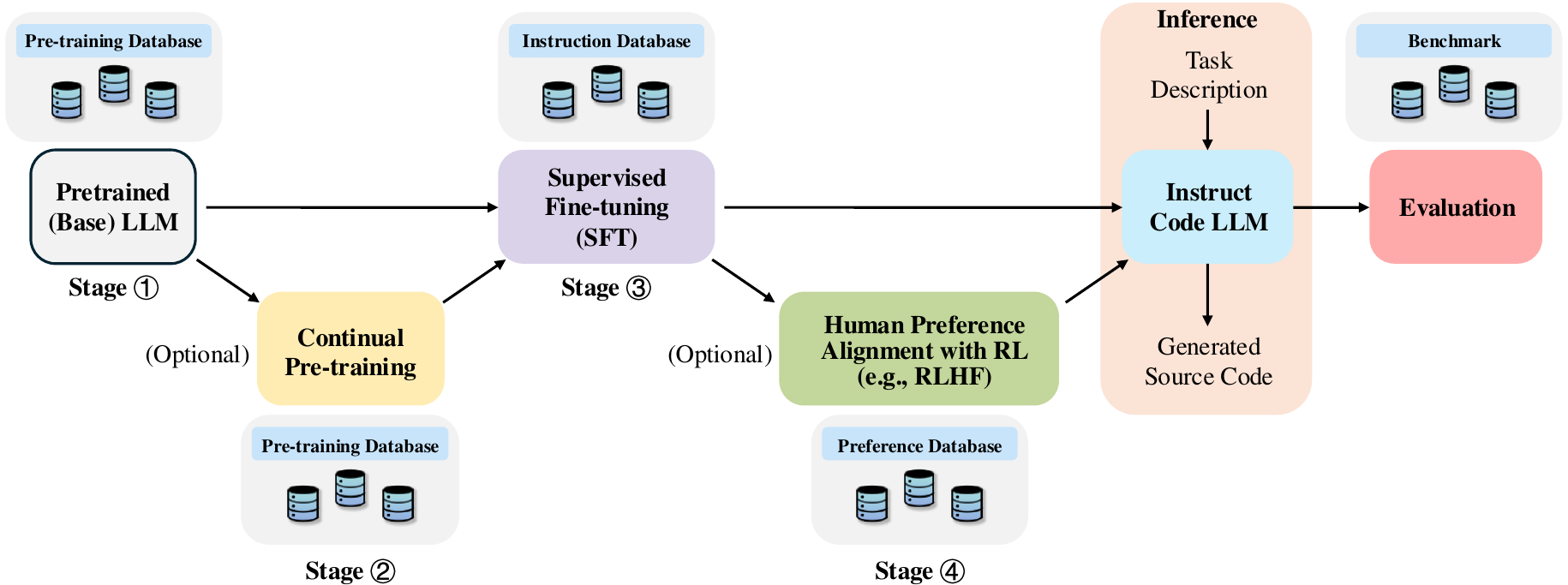}
\caption{\done{A diagram illustrating the general training, inference, and evaluation workflow for Code LLMs and their associated databases. 
The training workflow is mainly divided into four distinct stages: 
Stage \textcircled{1} and \textcircled{2} are the pre-training phase, whereas Stages \textcircled{3} and \textcircled{4} represent the post-training phases. It is important to note that Stage \textcircled{2} and \textcircled{4} are optional. 
For instance, StarCoder \cite{li2023starcoder} incorporates only Stage \textcircled{1}. WizardCoder \cite{luo2023wizardcoder}, fine-tuned upon StarCoder, includes only Stage \textcircled{3}, while Code Llama \cite{roziere2023code}, continually pre-trained on Llama 2, encompasses Stages \textcircled{2} and \textcircled{3}. DeepSeek-Coder-V2 \cite{zhu2024deepseek}, continually pre-trained on DeepSeek-V2, covers Stages \textcircled{2}, \textcircled{3}, and \textcircled{4}. 
Note that pre-trained model can be directly used for inference through prompt engineering.
}}
\label{fig:codellm_workflow}
\end{figure}

\definecolor{line-color}{RGB}{0, 119, 182}
\definecolor{fill-color}{RGB}{114, 200, 222}
\tikzstyle{category}=[
    rectangle,
    draw=line-color,
    rounded corners,
    text opacity=1,
    minimum height=1.5em,
    minimum width=5em,
    inner sep=2pt,
    align=center,
    fill opacity=.5,
]

\tikzstyle{leaf}=[category,minimum height=1em,
fill=fill-color!40, text width=20em,  text=black,align=left,font=\tiny,
inner xsep=2pt,
inner ysep=1pt,
]

\begin{figure*}[tp]
\centering
\scalebox{0.97}{
\begin{forest}
  forked edges,
  for tree={
  grow=east,
  reversed=true,
  anchor=base west,
  parent anchor=east,
  child anchor=west,
  base=left,
  font=\small,
  rectangle,
  draw=line-color,
  rounded corners,align=left,
  minimum width=2.5em,
s sep=3pt,
inner xsep=2pt,
inner ysep=1pt,
ver/.style={rotate=90, child anchor=north, parent anchor=south, anchor=center},
  },
  where level=1{text width=3.2em,font=\scriptsize,}{},
  where level=2{text width=3.8em,font=\tiny}{},
  where level=3{text width=4.0em,font=\tiny}{},
  where level=4{text width=4.2em,font=\tiny}{},
  [LLMs for Code Generation, ver
    [Data \\Curation \\(Sec. \ref{sec:data_curation})
        [Pre-training
            [CodeSearchNet\cite{husain2019codesearchnet}{,}
             Google BigQuery\cite{hoffa2016github}{,}
             The Pile\cite{gao2020pile}{,}
             CodeParrot\cite{tunstall2022natural}{,}
             GitHub Code\cite{tunstall2022natural}\\
             ROOTS\cite{laurenccon2022bigscience}{,}
             The Stack\cite{kocetkov2022stack}{,}
             The Stack v2\cite{lozhkov2024starcoder}
            ,leaf,text width=24.5em]
        ]
        [Instruction \\Tuning
            [CommitPackFT \cite{muennighoff2023octopack}{,}
             Code Alpaca\cite{codealpaca}{,}
             OA-Leet\cite{oa-leet10k}{,}
             OSS-Instruct\cite{wei2023magicoder}{,}
             Evol-instruction\cite{evol_instruction}\\
             Self-OSS-Instruct-SC2-Exec-Filter\cite{starcoder2instruct}
            ,leaf,text width=24.5em]
        ]
        [Benchmarks
            [General
                [HumanEval\cite{chen2021evaluating}{,}
                 HumanEval+\cite{liu2024your}{,}
                 HumanEvalPack\cite{muennighoff2023octopack}{,}
                 MBPP\cite{austin2021program}\\
                 MBPP+\cite{liu2024your}{,}
                 CoNaLa\cite{yin2018learning}{,}
                 Spider\cite{yu2018spider}{,}
                 CONCODE\cite{iyer2018mapping}{,}
                 ODEX\cite{wang2022execution}\\
                 CoderEval\cite{yu2024codereval}{,}
                 ReCode\cite{wang2022recode}{,}
                 StudentEval\cite{babe2023studenteval}
                ,leaf,text width=19em]
            ]
            [Competitions
                [APPS\cite{hendrycks2021measuring}{,}
                 CodeContests\cite{li2022competition}
                ,leaf,text width=19em]
            ]
            [Data Science
                [DSP\cite{chandel2022training}{,}
                DS-1000\cite{lai2023ds}{,}
                ExeDS\cite{huang2022execution}
                ,leaf,text width=19em]
            ]
            [Multilingual
                [MBXP\cite{athiwaratkun2022multi}{,}
                 Multilingual HumanEval\cite{athiwaratkun2022multi}{,}
                 HumanEval-X\cite{zheng2023codegeex}{,}
                 MultiPL-E\cite{cassano2022scalable}\\
                 xCodeEval\cite{khan2023xcodeeval}
                ,leaf,text width=19em]
            ]
            [Reasoning
                [MathQA-X\cite{athiwaratkun2022multi}{,}
                 MathQA-Python\cite{austin2021program}{,}
                 GSM8K\cite{cobbe2021training}{,}
                 GSM-HARD\cite{gao2023pal}
                ,leaf,text width=19em]
            ]
            [Repository
                [RepoEval\cite{zhang2023repocoder}{,}
                 Stack-Repo\cite{shrivastava2023repofusion}{,} 
                 Repobench\cite{liu2023repobench}{,}
                 EvoCodeBench\cite{li2024evocodebench}\\
                 SWE-bench\cite{jimenez2023swe}{,}
                 CrossCodeEval\cite{ding2024crosscodeeval}{,}
                 SketchEval\cite{zan2024codes}
                ,leaf,text width=19em]
            ]
        ]
    ]
    [Recent \\Advances
        [Data \\Synthesis \\(Sec. \ref{sec:data_synthesis})
            [Self-Instruct \cite{wang2023self}{,}
            Evol-Instruct \cite{xu2023wizardlm}{,}
            Phi-1\cite{gunasekar2023textbooks}{,}
            Code Alpaca\cite{codealpaca}{,}
            WizardCoder\cite{luo2023wizardcoder}\\
            Magicoder\cite{wei2023magicoder}{,}
            StarCoder2-instruct \cite{starcoder2instruct}
            ,leaf,text width=24.5em]
        ]
        [Pre-training \\(Sec. \ref{sec:pre-training})
            [Model \\Architectures
                [Encoder-Decoder
                    [PyMT5\cite{clement2020pymt5}{,}
                    PLBART\cite{ahmad2021unified}{,}
                    CodeT5\cite{wang2021codet5}{,}
                    JuPyT5\cite{chandel2022training}\\
                    AlphaCode\cite{li2022competition}{,}
                    CodeRL\cite{le2022coderl}{,}
                    ERNIE-Code\cite{chai2022ernie}\\
                    PPOCoder\cite{shojaee2023execution}{,}
                    CodeT5+\cite{wang2023codet5+}{,}
                    CodeFusion\cite{singh2023codefusion}\\
                    AST-T5\cite{gong2024ast}
                    ,leaf,text width=13em]
                ]
                [Decoder-Only
                    [GPT-C\cite{svyatkovskiy2020intellicode}{,}
                    GPT-Neo\cite{gpt-neo}{,}
                    GPT-J\cite{gpt-j}{,}
                    Codex\cite{chen2021evaluating}\\
                    CodeGPT\cite{lu2021codexglue}{,}
                    CodeParrot\cite{tunstall2022natural}{,}
                    PolyCoder\cite{xu2022systematic}\\
                    CodeGen\cite{nijkamp2022codegen}{,}
                    GPT-NeoX\cite{black2022gpt}{,}
                    PaLM-Coder\cite{chowdhery2023palm}\\
                    InCoder\cite{fried2022incoder}{,}
                    PanGu-Coder\cite{christopoulou2022pangu}{,}
                    PyCodeGPT\cite{zan2022cert}\\
                    CodeGeeX\cite{zheng2023codegeex}{,}
                    BLOOM\cite{le2023bloom}{,}
                    ChatGPT\cite{gpt-3.5-turbo}\\
                    SantaCoder\cite{allal2023santacoder}{,}
                    LLaMA\cite{touvron2023llama}{,}
                    GPT-4\cite{achiam2023gpt}\\
                    CodeGen2\cite{nijkamp2023codegen2}{,}
                    replit-code\cite{replit-code}{,}
                    StarCoder\cite{li2023starcoder}\\
                    WizardCoder\cite{luo2023wizardcoder}{,}
                    phi-1\cite{gunasekar2023textbooks}{,}
                    ChainCoder\cite{zheng2023outline}\\
                    CodeGeeX2\cite{zheng2023codegeex}{,}
                    PanGu-Coder2\cite{shen2023pangu}{,}
                    Llama 2\cite{touvron2023llama2}\\
                    OctoPack\cite{muennighoff2023octopack}{,}
                    Code Llama\cite{roziere2023code}{,}
                    MFTCoder\cite{liu2023mftcoder}\\
                    phi-1.5\cite{li2023textbooks}{,}
                    CodeShell\cite{xie2024codeshell}{,}
                    Magicoder\cite{wei2023magicoder}\\
                    AlphaCode 2\cite{alphacode2}{,} 
                    StableCode\cite{pinnaparaju2024stable}{,}
                    WaveCoder\cite{yu2023wavecoder}\\
                    phi-2\cite{phi-2}{,}
                    DeepSeek-Coder\cite{guo2024deepseek}{,}
                    StepCoder\cite{dou2024stepcoder}\\
                    OpenCodeInterpreter\cite{zheng2024opencodeinterpreter}{,}
                    StarCoder 2\cite{lozhkov2024starcoder}\\
                    Claude 3\cite{claude3}{,}
                    ProCoder\cite{bi2024iterative}{,}
                    CodeGemma\cite{codegemma_2024}\\
                    CodeQwen\cite{codeqwen}{,}
                    Llama3\cite{llama3}\\
                    StarCoder2-Instruct\cite{starcoder2instruct}{,}
                    Codestral\cite{codestral}
                    ,leaf,text width=13em]
                ]
            ]
            [Pre-training \\Tasks
                [CLM\cite{li2023starcoder,luo2023wizardcoder,wei2023magicoder,guo2024deepseek}{,} 
                DAE\cite{ahmad2021unified,wang2021codet5,wang2023codet5+}{,}
                Auxiliary\cite{wang2021codet5,chai2022ernie,wang2023codet5+} 
                ,leaf,text width=16.5em]
            ]
        ]
        [Fine-tuning
            [Instruction \\Tuning \\(Sec. \ref{sec:instruction_tuning})
                [Full Parameter\\ Fine-tuning
                    [Code Alpaca\cite{codealpaca}{,} 
                    CodeT5+\cite{wang2021codet5}{,} WizardCoder\cite{luo2023wizardcoder}\\ StarCoder\cite{li2023starcoder}{,} 
                    Pangu-Coder2\cite{shen2023pangu}{,} OctoPack\cite{muennighoff2023octopack}\\ CodeGeeX2\cite{zheng2023codegeex}{,} Magicoder\cite{wei2023magicoder}{,} CodeGemma\cite{codegemma_2024}\\ 
                    StarCoder2-instruct\cite{starcoder2instruct}
                    ,leaf,text width=13em]
                ]
                [Parameter \\Efficient \\Fine-tuning
                    [CodeUp\cite{codeup}{,} 
                    ASTRAIOS\cite{zhuo2024astraios}
                    ,leaf,text width=8em]
                ]
            ]
            [Reinforcement \\Learning \\with Feedback \\(Sec. \ref{sec:reinforcement_learning})
              [CodeRL\cite{le2022coderl}{,} 
              CompCoder\cite{wang2022compilable}{,} PPOCoder\cite{shojaee2023execution}{,} 
              RLTF\cite{liu2023rltf}\\
              PanGu-Coder2\cite{shen2023pangu}{,}
              StepCoder\cite{dou2024stepcoder}
                ,leaf,text width=15em]
            ]
        ]
        [Prompting \\Engineering \\(Sec. \ref{sec:prompting})
          [Reflexion\cite{shinn2024reflexion}{,} 
          LATS\cite{zhou2023language}{,} 
          Self-Debugging\cite{chen2023teaching}{,} 
          SelfEvolve\cite{jiang2023selfevolve}\\ 
          Theo X. et al.\cite{olausson2023self}{,} 
          CodeT\cite{chen2022codet}{,} 
          LEVER\cite{ni2023lever}{,}
          AlphaCodium\cite{ridnik2024code}
            ,leaf,text width=17em]
        ]
        [Repository\\ Level \& Long \\Context \\(Sec. \ref{sec:repository_level}) 
            [RepoCoder\cite{zhang2023repocoder}{,} 
            CoCoMIC\cite{ding2022cocomic}{,} 
            RepoHyper\cite{phan2024repohyper}{,} 
            RLPG\cite{shrivastava2023repository}\\
            Repoformer\cite{wu2024repoformer}{,} 
            RepoFusion\cite{shrivastava2023repofusion}{,} 
            ToolGen\cite{wang2024teaching}{,}
            CodePlan\cite{bairi2023codeplan}\\
            CodeS\cite{zan2024codes} 
            ,leaf,text width=17em]
        ]
        [Retrieval \\Augmented \\(Sec. \ref{sec:retrieval_augmented})
          [HGNN\cite{liu2020retrieval}{,} 
          REDCODER\cite{parvez2021retrieval}{,} 
          ReACC\cite{lu2022reacc}{,} 
          DocPrompting\cite{zhou2022docprompting}\\ 
          RepoCoder\cite{zhang2023repocoder}{,} 
          Su et al.\cite{su2024arks}
            ,leaf,text width=17em]
        ]
        [Autonomous \\Coding Agents \\(Sec. \ref{sec:autonomous_agents})
          [AgentCoder \cite{huang2023agentcoder}{,}
          MetaGPT\cite{hong2023metagpt}{,}
          CodeAct \cite{wang2024executable}{,}
          AutoCodeRover \cite{zhang2024autocoderover}{,}
          Devin\cite{Devin}\\
          OpenDevin\cite{OpenDevin}{,}
          SWE-agent\cite{swe-agent}{,}
          L2MAC\cite{holt2023l2mac}{,}
          OpenDevin CodeAct 1.0\cite{OpenDevin_CodeAct}
            ,leaf,text width=22em]
        ]
    ]
    [Evaluation \\(Sec. \ref{sec:evaluation})
        [Metrics
            [Exact Match{,} 
            BLEU\cite{papineni2002bleu}{,} 
            ROUGE\cite{lin2004rouge}{,}
            METEOR\cite{banerjee2005meteor}{,}
            CodeBLEU\cite{ren2020codebleu}{,}
            pass@k\cite{chen2021evaluating}\\
            n@k\cite{li2022competition}{,}
            test case average\cite{hendrycks2021measuring}{,}
            execution accuracy\cite{rajkumar2022evaluating}{,}
            pass@t\cite{olausson2023self}{,}
            perplexity\cite{jelinek1977perplexity}
            ,leaf,text width=22em]
        ]
        [Human \\Evaluation
            [CodePlan\cite{bairi2023codeplan}{,} 
            RepoFusion\cite{shrivastava2023repofusion}{,}
            CodeBLEU\cite{ren2020codebleu}
            ,leaf,text width=13em]
        ]
        [LLM-as-a-Judge
            [AlpacaEval\cite{alpaca_eval}{,}
            MT-bench\cite{zheng2024judging}{,}
            ICE-Score\cite{zhuo2024ice}
            ,leaf,text width=13em]
        ]
    ]
    [\textcolor{black}{Code LLMs} \\\textcolor{black}{Alignment} \\(Sec. \ref{sec:grest_llm4code})
      [\textcolor{black}{Green\cite{shi2024greening,wei2023towards}{,}}
      \textcolor{black}{Responsibility\cite{liu2023uncovering,xu2024first}{,}}
      \textcolor{black}{Efficiency\cite{yang2024robustness}{,}}
      \textcolor{black}{Safety\cite{schuster2021you,hajipour2024codelmsec,yang2024unveiling,al2024traces,yuan2023gpt,fried2022incoder,allal2023santacoder}{,}}
      \textcolor{black}{Trustworthiness\cite{ji2023benchmarking,palacio2023evaluating}}
        ,leaf,text width=29.8em]
    ]
    [Application \\(Sec. \ref{sec:application})
        [GitHub Copilot\cite{chen2021evaluating}{,} 
        CodeGeeX\cite{zheng2023codegeex}{,} 
        CodeWhisperer\cite{CodeWhisperer}{,} 
        Codeium\cite{Codeium}{,} 
        CodeArts Snap\cite{shen2023pangu}{,} 
        TabNine\cite{TabNine}{,} 
        Replit\cite{Replit}
         ,leaf,text width=29.8em]
    ]
  ]
\end{forest}
}
\caption{Taxonomy of \done{LLMs} for code generation.}
\label{fig:taxonomy}
\end{figure*}
\section{Large Langauge Models for Code Generation}\label{sec:overview}
\done{LLMs} with Transformer architecture have revolutionized a multitude of fields, and their application in code generation has been particularly impactful. These models follow a comprehensive process that starts with the curation and synthesis of code data, followed by a structured training approach that includes pre-training and fine-tuning (instruction tuning), reinforcement learning with various feedback, and the use of sophisticated prompt engineering techniques. Recent advancements have seen the integration of repository-level and retrieval-augmented code generation, as well as the development of autonomous coding agents. Furthermore, the evaluation of coding abilities of LLMs has become a critical component of this research area. 
\done{
Figure \ref{fig:codellm_workflow} illustrates the general training, inference, and evaluation workflow for Code LLMs and their associated databases.
}

In the forthcoming sections, we will explore these dimensions of LLMs in the context of code generation in detail. Section \ref{sec:data_curation} will address the data curation and processing strategies employed throughout the various stages of LLM development. 
Section \ref{sec:data_synthesis} will discuss data synthesis methods designed to mitigate the scarcity of high-quality data. 
Section \ref{sec:pre-training} will outline the prevalent model architectures used in LLMs for code generation.
Moving to Section \ref{sec:instruction_tuning}, we will examine the techniques for full parameter fine-tuning and parameter-efficient fine-tuning, which are essential for tailoring LLMs to code generation task. 
Section \ref{sec:reinforcement_learning} will shed light on enhancing code quality through reinforcement learning, utilizing the power of feedback.
Section \ref{sec:prompting} will delve into the strategic use of prompts to maximize the coding capabilities of LLMs. The innovative approaches of repository-level and retrieval-augmented code generation will be elaborated in Sections \ref{sec:repository_level} and \ref{sec:retrieval_augmented}, respectively. 
Additionally, Section \ref{sec:autonomous_agents} will discuss the exciting field of autonomous coding agents.
\done{Section \ref{sec:evaluation} discusses various evaluation strategies and offer an empirical comparison using the widely recognized HumanEval, MBPP, and the more practical and challenging BigCodeBench benchmarks to highlight the progressive enhancements in LLM capabilities for code generation.}
\done{Furthermore, the ethical implications and the environmental impact of using LLMs for code generation are discussed in Section \ref{sec:responsible_codeai}, aiming to establish a trustworthiness, responsibility, safety, efficiency, and green of LLM for code generation.}
Lastly, Section \ref{sec:application} will provide insights into some of the practical applications that leverage LLMs for code generation, demonstrating the real-world impact of these sophisticated models. 
Through this comprehensive exploration, we aim to highlight the significance and potential of LLMs within the domain of automated code generation.

\subsection{Data Curation \& Processing}\label{sec:data_curation}
The exceptional performance of \done{LLMs} can be attributed to their training on large-scale and diverse datasets \cite{zan2023large}. 
Meanwhile, the extensive parameters of these models necessitate substantial data to unlock their full potential, in alignment with established scaling law \cite{kaplan2020scaling,hoffmann2022training}. 
For a general-purpose LLM, amassing a large-scale corpus of natural language from a variety of sources is imperative. Such sources include webpages, conversation data, books and news, scientific data, and code \cite{brown2020language,chowdhery2023palm,bai2023qwen,touvron2023llama,touvron2023llama2,yoo2024hyperclova}, while these data are often crawled from the web and must undergo meticulous and aggressive pre-processing \cite{raffel2020exploring,zhang2023unifying}. 
Fortunately, multiple platforms and websites offer large-scale, open-source, and permissively licensed code corpora, such as GitHub\footnote{\href{https://github.com}{https://github.com}} and Stack Overflow\footnote{\href{https://stackoverflow.com}{https://stackoverflow.com}}. 
Notably, the number of stars or forks of GitHub repositories has emerged as a valuable metric for filtering high-quality code datasets. In a similar vein, the quantity of votes on Stack Overflow can serve to discern the most relevant and superior answers.

\begin{figure*}[t]
\centering
\includegraphics[width=\linewidth]{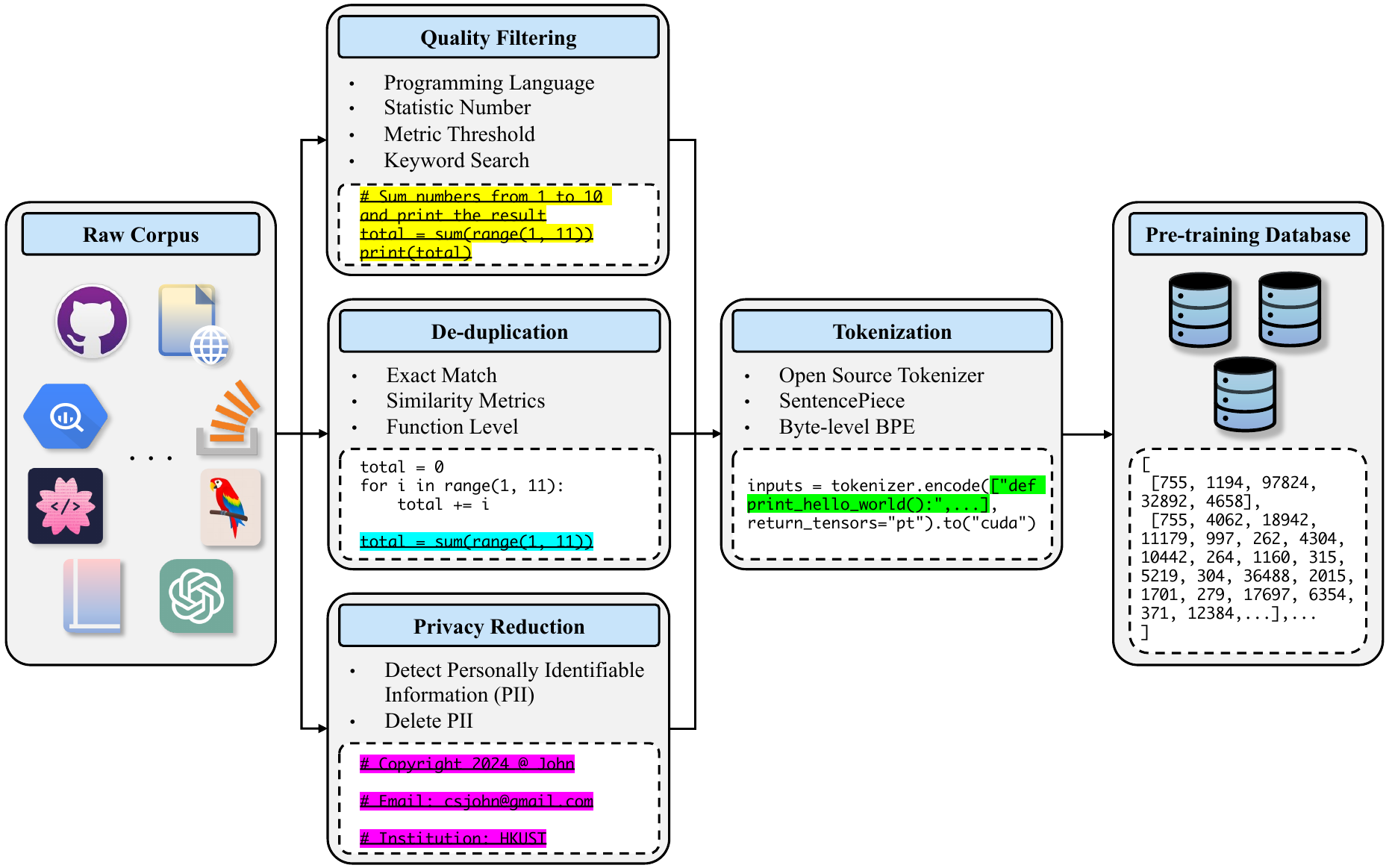}
\caption{A diagram depicting the standard data preprocessing workflow utilized in the pre-training phase of \done{LLMs} for code generation.}
\label{fig:datapipeline}
\end{figure*}

Nonetheless, raw datasets are frequently laden with redundant, noisy data and personal information, eliciting concerns regarding privacy leakage, which may include the names and email addresses of repository contributors \cite{carlini2021extracting,laurenccon2022bigscience,al2024traces}. 
Consequently, it is essential to undertake rigorous data-cleaning procedures. 
Typically, this process encompasses exact match deduplication, code data filtering based on average line length and a defined threshold for the fraction of alphanumeric characters, the removal of auto-generated files through keyword searches, and the expunction of personal user data \cite{tunstall2022natural,kocetkov2022stack}. Specifically, the standard data preprocessing workflow is depicted in Figure \ref{fig:datapipeline}. 

The development of a proficient LLM for code generation necessitates the utilization of various types of code data at different developmental stages. 
Therefore, we categorize code data into three distinct classes: pre-training datasets, instruction-tuning datasets, and benchmarks for performance evaluation. 
The subsequent subsections will provide a detailed illustration of code data within each classification.

\subsubsection{Pre-training}
The remarkable success of bidirectional pre-trained language models (PLMs) such as BERT \cite{devlin2018bert} and unidirectional PLMs like GPT \cite{radford2018improving} has firmly established the practice of pre-training on large-scale unlabeled datasets to endow models with a broad spectrum of general knowledge. 
Extending this principle to the realm of code generation enables \done{LLMs} to assimilate fundamental coding principles, including the understanding of code structure dependencies, the semantics of code identifiers, and the intrinsic logic of code sequences \cite{chen2021evaluating,wang2021codet5,guo2022unixcoder,wang2023codet5+}.
In light of this advancement, there has been a proliferation of large-scale unlabeled code datasets proposed to serve as the foundational training ground for LLMs to develop coding proficiency. 
A brief introduction of these datasets is as follows, with the statistics available in Table \ref{tab:pretraining_dataset}.

\begin{itemize}
    \item CodeSearchNet \cite{husain2019codesearchnet}: CodeSearchNet corpus is a comprehensive dataset, consisting of 2 million (comment, code) pairs from open-source repositories on GitHub. It includes code and documentation in several programming languages including Go, Java, PHP, Python, JavaScript, and Ruby. The dataset was primarily compiled to promote research into the problem of code retrieval using natural language.
    \item Google BigQuery \cite{hoffa2016github}: the Google BigQuery Public Datasets program offers a full snapshot of the content of more than 2.8 million open source GitHub repositories in BigQuery.
    \item The Pile \cite{gao2020pile}: the Pile is an 825 GiB diverse and open source language modeling dataset aggregating 22 smaller, high-quality datasets including GitHub, Books3, and Wikipedia (en). It aims to encompass text from as many modalities as possible, thereby facilitating the development of models with broader generalization capabilities. For code generation, the GitHub composite is specifically utilized.
    \item CodeParrot \cite{tunstall2022natural}: the CodeParrot dataset contains Python files used to train the code generation model in Chapter 10: Training Transformers from Scratch in the ``NLP with Transformers book'' \cite{tunstall2022natural}. Created with the GitHub dataset available via Google's BigQuery, the CodeParrot dataset includes approximately 22 million Python files and is 180 GB (50 GB compressed) big. 
    \item GitHub Code \cite{tunstall2022natural}: the GitHub Code dataset comprises 115M code files derived from GitHub, spanning 32 programming languages and 60 extensions totaling 1TB of data. The dataset was created from the public GitHub dataset on Google BiqQuery.
    \item ROOTS \cite{laurenccon2022bigscience}: the BigScience ROOTS Corpus is a 1.6TB dataset spanning 59 languages that was used to train the 176B BigScience Large Open-science Open-access Multilingual (BLOOM) language model. For the code generation task, the code subset of the ROOTS Corpus will be specifically utilized.  
    \item The Stack \cite{kocetkov2022stack}: the Stack contains over 6TB of permissively licensed source code files that cover 358 programming languages. The dataset was compiled as part of the BigCode Project, an open scientific collaboration working on the responsible development of Large Language Models for Code (Code LLMs).
    \item The Stack v2 \cite{lozhkov2024starcoder}: The Stack v2, a dataset created as part of the BigCode Project, contains over 3B files across more than 600 programming and markup languages. The dataset is derived from the Software Heritage archive\footnote{\href{https://archive.softwareheritage.org/}{https://archive.softwareheritage.org}}, the largest public archive of software source code and accompanying development history.
\end{itemize}

\begin{table}[t]
\caption{
The statistics of some commonly-used pre-training datasets for \done{LLMs} aimed at code generation. The column labeled `\textbf{\#PL}' indicates the number of programming languages included in each dataset. It should be noted that in the CodeSearchNet \cite{husain2019codesearchnet} dataset, each file represents a function, and for the Pile \cite{gao2020pile} and ROOTS \cite{laurenccon2022bigscience} datasets, only the code components are considered.
}
\label{tab:pretraining_dataset}
\centering
\scalebox{0.73}{
\rotatebox{0}{
    \begin{tabular}{llllcl}
    \toprule
        \textbf{Dataset} & \textbf{Size (GB)} & \textbf{Files (M)} & \textbf{\#PL} & \textbf{Date} & \textbf{Link}\\
    \midrule
        CodeSearchNet \cite{husain2019codesearchnet} & 20 & 6.5 & 6  & 2022-01 & \url{https://huggingface.co/datasets/code_search_net}\\
        Google BigQuery\cite{hoffa2016github}  & - & - & - & 2016-06 & \href{https://cloud.google.com/blog/topics/public-datasets/github-on-bigquery-analyze-all-the-open-source-code}{\url{github-on-bigquery-analyze-all-the-open-source-code}} \\
        The Pile \cite{gao2020pile} & 95 & 19 & - &  2022-01 & \url{https://huggingface.co/datasets/EleutherAI/pile}\\
        CodeParrot \cite{tunstall2022natural} & 180 & 22 & 1 & 2021-08 & \url{https://huggingface.co/datasets/transformersbook/codeparrot}\\
        GitHub Code\cite{tunstall2022natural} & 1,024 & 115 & 32 & 2022-02 & \url{https://huggingface.co/datasets/codeparrot/github-code}\\
        ROOTS \cite{laurenccon2022bigscience} & 163 & 15 & 13 & 2023-03 & \url{https://huggingface.co/bigscience-data} \\
        The Stack \cite{kocetkov2022stack} & 3,136 & 317 & 30 & 2022-10 & \url{https://huggingface.co/datasets/bigcode/the-stack}\\
        The Stack v2 \cite{lozhkov2024starcoder} & 32K & 3K & 619 & 2024-04 & \url{https://huggingface.co/datasets/bigcode/the-stack-v2}\\
    \bottomrule
    \end{tabular}
}
}
\end{table}

\subsubsection{Instruction Tuning}\label{sec:instruction_data}
\done{Instruction tuning refers to the process of supervised fine-tuning \done{LLMs} using a collection of datasets structured as various instructions, with the purpose of following a wide range of task instructions \cite{wei2021finetuned,sanh2022multitask,ouyang2022training,chung2024scaling}.
}
This method has demonstrated a considerable improvement in model performance and an enhanced ability to generalize to unseen tasks that the model has not previously encountered, as evidenced by recent studies \cite{ouyang2022training,chung2024scaling}.
Leveraging the benefits of instruction tuning, instruction tuning has been expanded into coding domains, especially for code generation, which involves the automatic generation of the intended code from a natural language description. The promise of instruction tuning in this area has led numerous researchers to develop large-scale instruction-tuning datasets tailored for code generation. 
Below, we provide an overview of several notable datasets tailored for instruction tuning, with their respective statistics detailed in Table \ref{tab:instruction_dataset}.

\begin{itemize}
    \item CodeAlpaca-20k \cite{codealpaca}: CodeAlpaca-20k is a collection of 20K instruction-following data generated using the data synthesis techniques termed Self-Instruct outlined in \cite{wang2023self}, with modifications for code generation, editing, and optimization tasks instead of general tasks. 
    \item CommitPackFT \cite{muennighoff2023octopack}: CommitPackFT is a 2GB refined version of CommitPack. It is filtered to only include high-quality commit messages that resemble natural language instructions.
    \item Evol-Instruct-Code-80k \cite{evol_instruction}: Evol-Instruct-Code-80k is an open-source implementation of Evol-Instruct-Code described in the WizardCoder paper \cite{luo2023wizardcoder}, which enhances the fine-tuning effect of pre-trained code large models by adding complex code instructions.
    \item Magicoder-OSS-Instruct-75k \cite{wei2023magicoder}: is a 75k synthetic data generated through OSS-Instruct with \texttt{gpt-3.5-turbo-1106} and used to train both Magicoder and Magicoder-S series models.
    \item Self-OSS-Instruct-SC2-Exec-Filter-50k \cite{starcoder2instruct}: Self-OSS-Instruct-SC2-Exec-Filter-50k is generated by StarCoder2-15B using the OSS-Instruct \cite{wei2023magicoder} data synthesis approach. It was subsequently used to fine-tune StarCoder-15B without any human annotations or distilled data from huge and proprietary LLMs.
\end{itemize}

\begin{table}[t]
\caption{
The statistics of several representative datasets used in instruction-tuning \done{LLMs} for code generation. The column labeled `\textbf{\#PL}' indicates the number of programming languages encompassed by each dataset. 
}
\label{tab:instruction_dataset}
\centering
\scalebox{0.6}{
\rotatebox{0}{
    \begin{tabular}{lllll}
    \toprule
        \textbf{Dataset} & \textbf{Size} & \textbf{\#PL} & \textbf{Date} & \textbf{Link}\\
    \midrule
        CodeAlpaca-20K \cite{codealpaca} & 20k &  -  & 2023-03  & \url{https://huggingface.co/datasets/sahil2801/CodeAlpaca-20k}\\
        CommitPackFT \cite{muennighoff2023octopack} & 2GB & 277  & 2023-08  & \url{https://huggingface.co/datasets/bigcode/commitpackft}\\
        Evol-Instruct-Code-80k \cite{evol_instruction} & 80k &  - & 2023-07 & \url{https://huggingface.co/datasets/nickrosh/Evol-Instruct-Code-80k-v1}\\
        evol-codealpaca-v1 \cite{evol-codealpaca-v1} & 110K & - & 2023-07 & \href{https://huggingface.co/datasets/theblackcat102/evol-codealpaca-v1}{https://huggingface.co/datasets/theblackcat102/evol-codealpaca-v1}\\
        Magicoder-OSS-Instruct-75k \cite{wei2023magicoder} & 75k & \begin{tabular}[c]{@{}l@{}}Python, Shell, \\TypeScript, C++, \\Rust, PHP, Java, \\Swift, C\#\end{tabular}  & 2023-12 & \url{https://huggingface.co/datasets/ise-uiuc/Magicoder-OSS-Instruct-75K}\\
        \makecell[l]{Self-OSS-Instruct-SC2-Exec-Filter-50k \cite{starcoder2instruct}} & 50k &  Python  & 2024-04  & \url{https://huggingface.co/datasets/bigcode/self-oss-instruct-sc2-exec-filter-50k}\\
    \bottomrule
    \end{tabular}
}
}
\end{table}

\subsubsection{Benchmarks}\label{sec:benchmark}
To rigorously assess the efficacy of \done{LLMs} for code generation, the research community has introduced a variety of high-quality benchmarks in recent years. 
Building on the foundational work by \cite{chen2021evaluating}, numerous variations of the HumanEval dataset and additional benchmarks have emerged, aiming to evaluate a broader spectrum of code generation capabilities in LLMs. 
We roughly divide these benchmarks into six distinct categories based on their application contexts, including general-purpose, competitive programming, data science, multilingual, logical reasoning, and repository-level. 
\done{
It is important to highlight that logical reasoning encompasses math-related benchmarks, as it aims to create ``code-based solutions'' for solving complex mathematical problems \cite{chen2022program,gao2023pal,zhou2023solving}. This strategy can therefore mitigate the limitations of LLMs in performing intricate mathematical computations.
}
The statistics for these benchmarks are presented in Table \ref{tab:benchmark}.

\textbf{General}
\begin{itemize}
    \item HumanEval \cite{chen2021evaluating}: HumanEval comprises 164 manually scripted Python programming problems, each featuring a function signature, docstring, body, and multiple unit tests.
    \item HumanEval+ \cite{liu2024your}: HumanEval+ extends the original HumanEval \cite{chen2021evaluating} benchmark by increasing the scale of the test cases by 80 times. As the test cases increase, HumanEval+ can catch significant amounts of previously undetected incorrect code synthesized by LLMs.
    \item HumanEvalPack \cite{muennighoff2023octopack}: expands HumanEval \cite{chen2021evaluating} by extending it to encompass three coding tasks across six programming languages, namely code synthesis, code repair, and code explanation. 
    \item MBPP \cite{austin2021program}: MBPP is a collection of approximately 974 Python programming problems, crowd-sourced and designed for entry-level programmers. Each problem comes with an English task description, a code solution, and three automated test cases. 
    \item MBPP+ \cite{liu2024your}: MBPP+ enhances MBPP \cite{austin2021program} by eliminating ill-formed problems and rectifying problems with incorrect implementations. The test scale of MBPP+ is also expanded by 35 times for test augmentation.
    \item CoNaLa \cite{yin2018learning}: CoNaLa contains almost 597K data samples for evaluating Python code generation. The curated part of CoNaLa is crawled from Stack Overflow, automatically filtered, and then curated by annotators. The mined part of CoNaLais automatically mined, with almost 600k examples.
    \item Spider \cite{yu2018spider}: Spider is large-scale complex text-to-SQL dataset covering 138 different domains. It has over 10K questions and 5.6K complex SQL queries on 200 databases. This dataset aims to test a model's ability to generalize to SQL queries, database schemas, and new domains.
    \item CONCODE \cite{iyer2018mapping}: CONCODE is a dataset with over 100K samples consisting of Java classes from public GitHub repositories. It provides near zero-shot conditions that can test the model's ability to generalize to unseen natural language tokens with unseen environments.
    \item ODEX \cite{wang2022execution}: ODEX is an open-domain dataset focused on the execution-based generation of Python code from natural language. It features 945 pairs of natural language queries and their corresponding Python code, all extracted from StackOverflow forums.
    \item CoderEval \cite{yu2024codereval}: CoderEval is a pragmatic code generation benchmark that includes 230 Python and 230 Java code generation problems. It can be used to evaluate the model performance in generating pragmatic code beyond just generating standalone functions.
    \item ReCode \cite{wang2022recode}: Recode serves as a comprehensive robustness evaluation benchmark. ReCode applies perturbations to docstrings, function and variable names, code syntax, and code format, thereby providing multifaceted assessments of a model's robustness performance. 
    \item StudentEval \cite{babe2023studenteval}: StudentEval is a dataset of 1,749 prompts for 48 problems, authored by 80 students who have only completed a one-semester Python programming class. Unlike many other benchmarks, it has multiple prompts per problem and multiple attempts by the same participant, each problem is also accompanied by a set of instructor-written test cases.
    \done{\item BigCodeBench \cite{zhuo2024bigcodebench}: BigCodeBench has 1,140 complex Python programming tasks, covering 723 function calls from 139 popular libraries across 7 domains. This benchmark is specifically designed to assess LLMs' ability to call multiple functions from cross-domain libraries and follow complex instructions to solve programming tasks, helping to bridge the evaluation gap between isolated coding exercises and the real-world programming scenario.}
    \done{\item ClassEval \cite{du2024evaluating}: ClassEval is a manually-crafted benchmark consisting of 100 classes and 412 methods for evaluating LLMs in the class-level code generation scenario. Particularly, the task samples of ClassEval present higher complexities, involving long code generation and sophisticated docstring information, thereby benefiting the evaluation of the LLMs' capabilities in generating complicated code.}
    \done{\item NaturalCodeBench \cite{zhang2024naturalcodebench}: NaturalCodeBench is a comprehensive code benchmark featuring 402 high-quality problems in Python and Java. These problems are selected from natural user queries from online coding services and span 6 distinct domains, shaping an evaluation environment aligned with real-world applications.}
\end{itemize}

\textbf{Competitions}
\begin{itemize}
    \item APPS \cite{hendrycks2021measuring}: The APPS benchmark is composed of 10K Python problems, spanning three levels of difficulty: introductory, interview, and competition. Each entry in the dataset includes a programming problem described in English, corresponding ground truth Python solutions, test cases defined by their inputs and outputs or function names if provided.
    \item CodeContests \cite{li2022competition}: CodeContests is a competitive programming dataset consisting of samples from various sources including Aizu, AtCoder, CodeChef, Codeforces, and HackerEarth. The dataset encompasses programming problems accompanied by test cases in the form of paired inputs and outputs, along with both correct and incorrect human solutions in multiple programming languages.
    \done{\item LiveCodeBench \cite{naman2024livecodebench}: LiveCodeBench is a comprehensive and contamination-free benchmark for evaluating a wide array of code-related capabilities of LLMs, including code generation, self-repair, code execution, and test output prediction. It continuously gathers new coding problems from contests across three reputable competition platforms: LeetCode, AtCoder, and CodeForces. The latest release of the dataset includes 713 problems that were released between May 2023 and September 2024.
    }
\end{itemize}

\textbf{Data Science}
\begin{itemize}
    \item DSP \cite{chandel2022training}: DSP allows for model evaluation based on real data science pedagogical notebooks. It includes well-structured problems, along with unit tests to verify the correctness of solutions and a Docker environment for reproducible execution.
    \item DS-1000 \cite{lai2023ds}: DS-1000 has 1K science questions from seven Python libraries, namely NumPy, Pandas, TensorFlow, PyTorch, SciPy, Scikit-learn, and Matplotlib. The DS-1000 benchmark features: (1) realistic problems with diverse contexts (2) implementation of multi-criteria evaluation metrics, and (3) defense against memorization.
    \item ExeDS \cite{huang2022execution}: ExeDS is a data science code generation dataset specifically designed for execution evaluation. It contains 534 problems with execution outputs from Jupyter Notebooks, as well as 123K examples for training and validation.
\end{itemize}

\textbf{Multilingual}
\begin{itemize}
    \item MBXP \cite{athiwaratkun2022multi}: MBXP is a multilingual adaptation of the original MBPP \cite{austin2021program} dataset. It is created using a framework that translates prompts and test cases from the original Python datasets into the corresponding data in the targeted programming language.
    \item Multilingual HumanEval \cite{athiwaratkun2022multi}: Multilingual HumanEval is a dataset derived from HumanEval \cite{chen2021evaluating}. It is designed to assess the performance of models in a multilingual context.
    It helps uncover the generalization ability of the given model on languages that are out-of-domain.
    \item HumanEval-X \cite{zheng2023codegeex}: HumanEval-X is developed for evaluating the multilingual ability of code generation models with 820 hand-writing data samples in C++, Java, JavaScript, and Go.
    \item MultiPL-E \cite{cassano2022scalable}: 
    MultiPL-E is a dataset for evaluating LLMs for code generation across 18 programming languages. It adopts the HumanEval \cite{chen2021evaluating} and the MBPP \cite{austin2021program} Python benchmarks and uses little compilers to translate them to other languages.
    \item xCodeEval \cite{khan2023xcodeeval}: xCodeEval is an executable multilingual multitask benchmark consisting of 25M examples covering 17 programming languages. Its tasks include code understanding, generation, translation, and retrieval.
\end{itemize}

\textbf{Reasoning}
\begin{itemize}
    \item MathQA-X \cite{athiwaratkun2022multi} MathQA-X is the multilingual version of MathQA \cite{amini2019mathqa}. It is generated by utilizing a conversion framework that converts samples from Python datasets into the target language.
    \item MathQA-Python \cite{austin2021program} MathQA-Python is a Python version of the MathQA benchmark\cite{amini2019mathqa}. The benchmark, containing more than 23K problems, is designed to assess the capability of models to synthesize code from complex textual descriptions.
    \item GSM8K \cite{cobbe2021training}: GSM8K is a dataset of 8.5K linguistically diverse grade school math problems. The dataset is crafted to facilitate the task of question answering on basic mathematical problems that requires multi-step reasoning.
    \item GSM-HARD \cite{gao2023pal}: GSM-HARD is a more challenging version of the GSM8K \cite{cobbe2021training} dataset. It replaces the numbers in the GSM8K questions with larger, less common numbers, thereby increasing the complexity and difficulty level of the problems.
    \done{\item CRUXEval \cite{gu2024cruxeval}: CRUXEval contains 800 Python functions, each paired with an input-output example. This benchmark supports two tasks: input prediction and output prediction, designed to evaluate the code reasoning, understanding, and execution capabilities of code LLMs.}
\end{itemize}

\begin{table}[t]
\caption{
The detailed statistics of commonly-used benchmarks used in evaluating \done{LLMs} for code generation. 
The column labeled `\textbf{\#PL}' indicates the number of programming languages included in each dataset. For the sake of brevity, we list the programming languages (PLs) for benchmarks that support fewer than or include five PLs. For benchmarks with six or more PLs, we provide only a numerical count of the PLs supported.
}
\label{tab:benchmark}
\centering
\scalebox{0.63}{
\rotatebox{0}{
    \begin{tabular}{llllcl}
    \toprule
        \textbf{Scenario} & \textbf{Benchmark} & \textbf{Size} & \textbf{\#PL} & \textbf{Date} & \textbf{Link}\\
    \midrule
        \multirow{15}*{General}& HumanEval \cite{chen2021evaluating} &164&Python& 2021-07&\url{https://huggingface.co/datasets/openai_humaneval}\\
         & HumanEval+ \cite{liu2024your} &164&Python&2023-05&\url{https://huggingface.co/datasets/evalplus/humanevalplus}\\
         & HumanEvalPack \cite{muennighoff2023octopack} &164&6&2023-08&\url{https://huggingface.co/datasets/bigcode/humanevalpack}\\
         & MBPP \cite{austin2021program} &974&Python&2021-08&\url{https://huggingface.co/datasets/mbpp}\\
         & MBPP+ \cite{liu2024your} &378&Python&2023-05&\url{https://huggingface.co/datasets/evalplus/mbppplus}\\
         & CoNaLa \cite{yin2018learning} &596.88K&Python&2018-05&\url{https://huggingface.co/datasets/neulab/conala}\\
         & Spider \cite{yu2018spider} &8,034&SQL&2018-09&\url{https://huggingface.co/datasets/xlangai/spider}\\
         & CONCODE \cite{iyer2018mapping} &104K&Java&2018-08&\href{https://huggingface.co/datasets/AhmedSSoliman/CodeXGLUE-CONCODE}{\url{https://huggingface.co/datasets/AhmedSSoliman/CONCOD}}\\
         & ODEX \cite{wang2022execution} &945&Python&2022-12&\url{https://huggingface.co/datasets/neulab/odex}\\
         & CoderEval \cite{yu2024codereval} &460&Python, Java&2023-02&\url{https://github.com/CoderEval/CoderEval}\\
         & ReCode \cite{wang2022recode} &1,138&Python&2022-12&\url{https://github.com/amazon-science/recode}\\
         & StudentEval \cite{babe2023studenteval} &1,749&Python&2023-06&\url{https://huggingface.co/datasets/wellesley-easel/StudentEval} \\
         & \done{BigCodeBench \cite{zhuo2024bigcodebench}} & \done{1,140} & \done{Python} & \done{2024-06}& \done{\url{https://huggingface.co/datasets/bigcode/bigcodebench}} \\
         & \done{ClassEval \cite{du2024evaluating}} &\done{100} & \done{Python} & \done{2023-08}&\done{\url{https://huggingface.co/datasets/FudanSELab/ClassEval}} \\
         & \done{NaturalCodeBench \cite{zhang2024naturalcodebench}} &\done{402} & \done{Python, Java} & \done{2024-05}&\done{\url{https://github.com/THUDM/NaturalCodeBench}} \\
    \midrule
        \multirow{3}*{Competitions} & APPS \cite{hendrycks2021measuring} &10,000&Python&2021-05&\url{https://huggingface.co/datasets/codeparrot/apps} \\
        & CodeContests \cite{li2022competition} &13,610&\makecell[l]{C++, Python,\\ Java}&2022-02&\url{https://huggingface.co/datasets/deepmind/code_contests} \\
        & \done{LiveCodeBench \cite{naman2024livecodebench}} &\done{\makecell[l]{713\\ Updating}} & \done{Python} & \done{2024-03}&\done{\url{https://github.com/LiveCodeBench/LiveCodeBench}} \\
    \midrule
        \multirow{3}*{Data Science} & DSP \cite{chandel2022training} &1,119&Python&2022-01&\url{https://github.com/microsoft/DataScienceProblems}\\
        & DS-1000 \cite{lai2023ds} &1,000&Python&2022-11&\url{https://huggingface.co/datasets/xlangai/DS-1000} \\
        & ExeDS \cite{huang2022execution} &534&Python&2022-11&\url{https://github.com/Jun-jie-Huang/ExeDS} \\
    \midrule
        \multirow{5}*{Multilingual} & MBXP \cite{athiwaratkun2022multi}  &12.4K&13&2022-10&\url{https://huggingface.co/datasets/mxeval/mbxp} \\
        & Multilingual HumanEval \cite{athiwaratkun2022multi}  &1.9K&12&2022-10&\url{https://huggingface.co/datasets/mxeval/multi-humaneval} \\
        & HumanEval-X \cite{zheng2023codegeex}  &820&\makecell[l]{Python, C++, \\Java, JavaScript,\\ Go}&2023-03&\url{https://huggingface.co/datasets/THUDM/humaneval-x} \\
        & MultiPL-E \cite{cassano2022scalable}  &161&18&2022-08&\url{https://huggingface.co/datasets/nuprl/MultiPL-E} \\
        & xCodeEval \cite{khan2023xcodeeval}  &5.5M&11&2023-03&\url{https://github.com/ntunlp/xCodeEval} \\
    \midrule
        \multirow{5}*{Reasoning} & MathQA-X \cite{athiwaratkun2022multi}  &5.6K& \makecell[l]{Python, Java, \\JavaScript} &2022-10&\url{https://huggingface.co/datasets/mxeval/mathqa-x} \\
        & MathQA-Python \cite{austin2021program} &23,914&Python&2021-08& \url{https://github.com/google-research/google-research} \\
        & GSM8K \cite{cobbe2021training} &8.5K&Python&2021-10&\url{https://huggingface.co/datasets/gsm8k} \\
        & GSM-HARD \cite{gao2023pal} &1.32K&Python&2022-11&\url{https://huggingface.co/datasets/reasoning-machines/gsm-hard} \\
        & \done{CRUXEval \cite{gu2024cruxeval}} &\done{800} & \done{Python} & \done{2024-01}&\done{\url{https://huggingface.co/datasets/cruxeval-org/cruxeval}} \\
    \midrule
        \multirow{7}*{Repository} & RepoEval \cite{zhang2023repocoder} &3,573&Python, Java&2023-03&\url{https://paperswithcode.com/dataset/repoeval} \\
        & Stack-Repo \cite{shrivastava2023repofusion} & 200 &Java&2023-06&\url{https://huggingface.co/datasets/RepoFusion/Stack-Repo} \\
        & Repobench \cite{liu2023repobench} & 27k  &Python, Java&2023-01&\url{https://github.com/Leolty/repobench} \\
        & EvoCodeBench \cite{li2024evocodebench} &275&Python&2024-03&\url{https://huggingface.co/datasets/LJ0815/EvoCodeBench}\\
        & SWE-bench \cite{jimenez2023swe} &2,294&Python&2023-10&\url{https://huggingface.co/datasets/princeton-nlp/SWE-bench} \\
        & CrossCodeEval \cite{ding2024crosscodeeval} &10K&\makecell[l]{Python, Java,\\ TypeScript, C\#}&2023-10&\url{https://github.com/amazon-science/cceval} \\
        & SketchEval \cite{zan2024codes} &20,355&Python&2024-03&\url{https://github.com/nl2code/codes} \\
    \bottomrule
    \end{tabular}
}
}
\end{table}
\textbf{Repository}
\begin{itemize}
    \item RepoEval \cite{zhang2023repocoder}: RepoEval enables the evaluation of repository-level code completion. It can offer different levels of granularity and improved evaluation accuracy through the use of unit tests.
    \item Stack-Repo \cite{shrivastava2023repofusion}: Stack-Repo is a dataset of 200 Java repositories from GitHub with near-deduplicated files. These files are augmented with three types of repository contexts: prompt proposal contexts, BM25 Contexts (based on BM25 similarity scores), and RandomNN Contexts (obtained using the nearest neighbors in the representation space of an embedding model).
    \item Repobench \cite{liu2023repobench}: Repobench is a benchmark specifically used for evaluating repository-level code auto-completion systems. Supporting both Python and Java, it consists of three interconnected evaluation tasks: RepoBench-R (Retrieval), RepoBench-C (Code Completion), and RepoBench-P (Pipeline).
    \item EvoCodeBench \cite{li2024evocodebench}: EvoCodeBench is an evolutionary code generation benchmark, constructed through a rigorous pipeline and aligned with real-world repositories. This benchmark also provides comprehensive annotations and robust evaluation metrics.
    \item SWE-bench \cite{jimenez2023swe}: SWE-bench is a dataset that tests a model’s ability to automatically solve GitHub issues. The dataset has 2,294 Issue-Pull Request pairs from 12 popular Python repositories.
    \item CrossCodeEval \cite{ding2024crosscodeeval}: CrossCodeEval is a diverse and multilingual scope completion dataset covering four languages: Python, Java, TypeScript, and C\#. This benchmark tests the model's ability to understand in-depth cross-file information and accurately complete the code. 
    \item SketchEval \cite{zan2024codes}: SketchEval is a repository-oriented benchmark that encompasses data from 19 repositories, each varying in complexity. In addition to the dataset, SketchEval introduces a metric, known as SketchBLEU, to measure the similarity between two repositories based on their structures and semantics. 
\end{itemize}

\subsection{Data Synthesis}\label{sec:data_synthesis}
\begin{figure*}[t]
\centering
\includegraphics[width=\linewidth]{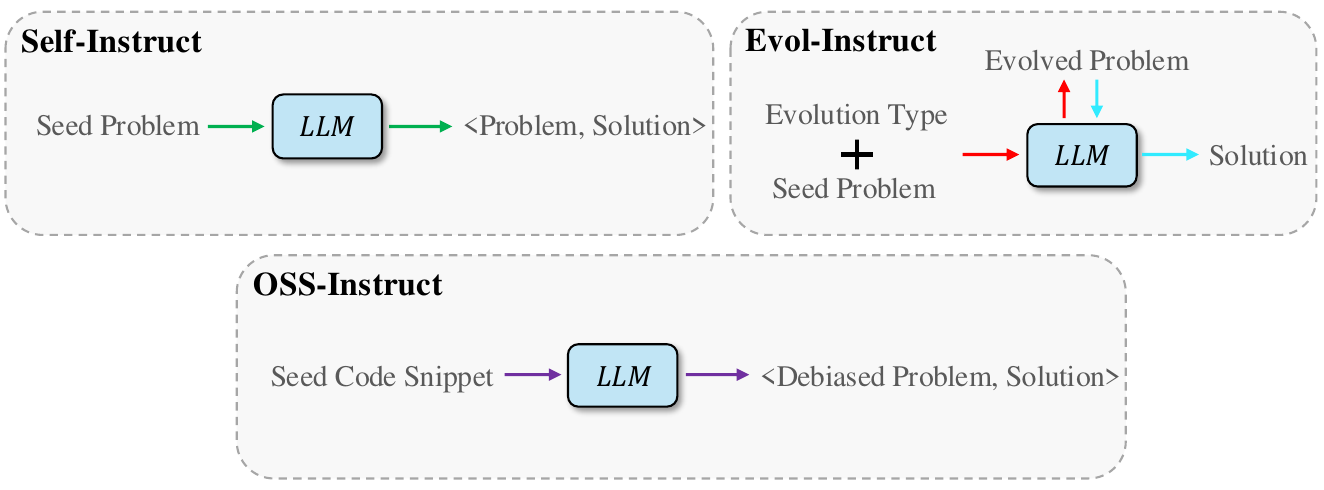}
\caption{\done{The comparison among three representative data synthesis methods used for generating instruction data with LLMs. The Code Alpaca \cite{codealpaca} employs the self-instruct method, whereas WizardCoder \cite{luo2023wizardcoder} and Magicoder \cite{wei2023magicoder} utilize the Evol-Instruct and OSS-Instruct methods, respectively.}}
\label{fig:data_synthesis}
\end{figure*}
Numerous studies have demonstrated that high-quality datasets are integral to enhancing the performance of \done{LLMs} in various downstream tasks \cite{brown2020language,meng2022generating,xie2023data,zhou2024lima,kopf2024openassistant,wettig2024qurating}. 
For instance, the LIMA model, a 65B parameter LLaMa language model fine-tuned with a standard supervised loss on a mere 1,000 meticulously curated prompts and responses, achieved performance on par with, or even superior to, GPT-4 in 43\% of evaluated cases. This figure rose to 58\% when compared to Bard and 65\% against \texttt{DaVinci003}, all without the use of reinforcement learning or human preference modeling \cite{zhou2024lima}. 
The QuRating initiative strategically selects pre-training data embodying four key textual qualities — writing style, facts \& trivia, required expertise, and educational value — that resonate with human intuition. Training a 1.3B parameter model on such data resulted in reduced perplexity and stronger in-context learning compared to baseline models \cite{wettig2024qurating}.

Despite these advancements, acquiring quality data remains a significant challenge due to issues such as data scarcity, privacy concerns, and prohibitive costs \cite{wang2023self,liu2024best}. 
Human-generated data is often labor-intensive and expensive to produce, and it may lack the necessary scope and detail to navigate complex, rare, or ambiguous scenarios.
As a resolution to these challenges, synthetic data has emerged as a viable alternative. By generating artificial datasets that replicate the intricacies of real-world information, models such as \texttt{GPT-3.5-turbo} \cite{gpt-3.5-turbo} and \texttt{GPT-4} \cite{achiam2023gpt} have enabled the creation of rich datasets without the need for human annotation \cite{wang2023self,hamalainen2023evaluating,liu2024best,moritzlaurer}. This approach is particularly beneficial in enhancing the instruction-following capabilities of LLMs, with a focus on generating synthetic instruction-based data.

A notable example of this approach is the Self-Instruct \cite{wang2023self} framework, which employs an off-the-shelf language model to generate a suite of instructions, inputs, and outputs. 
This data is then refined by removing invalid or redundant entries before being used to fine-tune the model. The empirical evidence supports the efficacy of this synthetic data generation methodology. 
Building upon this concept, the Alpaca \cite{alpaca} model, fine-tuned on 52k pieces of instruction-following data from a 7B parameter LLaMa \cite{touvron2023llama} model, exhibits performance comparable to the \texttt{text-davinci-003} model. 
WizardLM \cite{xu2023wizardlm} introduced the Evol-Instruct technique, which incrementally transforms simple instructions into more complex variants. The fine-tuned LLaMa model using this technique has shown promising results in comparison to established proprietary LLMs such as ChatGPT \cite{gpt-3.5-turbo} and GPT-4 \cite{achiam2023gpt}, to some extent.
Moreover, Microsoft has contributed to this field with their Phi series of models, predominantly trained on synthetic high-quality data, which includes Phi-1 (1.3B) \cite{gunasekar2023textbooks} for Python coding, Phi-1.5 (1.3B) \cite{li2023textbooks} for common sense reasoning and language understanding, Phi-2 (2.7B) \cite{phi-2} for advanced reasoning and language understanding, and Phi-3 (3.8B) \cite{abdin2024phi} for general purposes. These models have consistently outperformed larger counterparts across various benchmarks, demonstrating the efficacy of synthetic data in model training.

Drawing on the successes of data synthesis for general-purpose \done{LLMs}, researchers have expanded the application of synthetic data to the realm of code generation. The Code Alpaca model, as described in \cite{codealpaca}, has been fine-tuned on a 7B and 13B LLaMA model using a dataset of 20k instruction-following examples for code generation. This dataset was created by \texttt{text-davinci-003}\footnote{\href{https://platform.openai.com/docs/deprecations}{https://platform.openai.com}} and employed the Self-Instruct technique \cite{wang2023self}.
Building on this, the WizardCoder 15B \cite{luo2023wizardcoder} utilizes the Evol-Instruct technique to create an enhanced dataset of 78k evolved code instruction examples. This dataset originates from the initial 20k instruction-following dataset used by Code Alpaca \cite{codealpaca}, which was also generated by \texttt{text-davinci-003}. 
The WizardCoder model, fine-tuned on the StarCoder \cite{li2023starcoder} base model, achieved a 57.3\% \texttt{pass@1} on the HumanEval benchmarks. 
This performance not only surpasses all other open-source Code LLMs by a significant margin but also outperforms leading closed LLMs such as Anthropic’s Claude and Google’s Bard.
In a similar vein, Magicoder \cite{wei2023magicoder} introduces a novel data synthesis approach termed OSS-INSTRUCT which enlightens LLMs with open-source code snippets to generate high-quality instruction data for coding tasks. It aims to address the inherent biases often present in synthetic data produced by LLMs. 
Building upon CodeLlama \cite{roziere2023code}, the MagicoderS-CL-7B model — fine-tuned with 75k synthetic instruction data using the OSS-INSTRUCT technique and with \texttt{gpt-3.5-turbo-1106} as the data generator — has outperformed the prominent ChatGPT on the HumanEval Plus benchmark, achieving \texttt{pass@1} of 66.5\% versus 65.9\%.
In a noteworthy development, Microsoft has introduced the phi-1 model \cite{gunasekar2023textbooks}, a more compact LLM of only 1.3B parameters. Despite its smaller size, phi-1 has been trained on high-quality textbook data sourced from the web (comprising 6 billion tokens) and supplemented with synthetic textbooks and exercises generated with GPT-3.5 (1 billion tokens). It has achieved \texttt{pass@1} of 50.6\% on HumanEval and 55.5\% on MBPP, setting a new state-of-the-art for Python coding performance among existing small language models (SLMs).
The latest contribution to this field is from the BigCode team, which has presented StarCoder2-15B-instruct \cite{starcoder2instruct}, the first entirely self-aligned code LLM trained with a transparent and permissive pipeline. 
This model aligns closely with the OSS-INSTRUCT principles established by Magicoder, generating instructions based on seed functions filtered from the Stack v1 dataset \cite{kocetkov2022stack} and producing responses through self-validation. 
Unlike Magicoder, StarCoder2-15B-instruct employs its base model, StarCoder2-15B, as the data generator, thus avoiding reliance on large and proprietary LLMs like GPT-3.5-turbo \cite{gpt-3.5-turbo}.
\done{Figure \ref{fig:data_synthesis} illustrates the comparison between Self-Instruct, Evol-Instruct, and OSS-Instruct data synthesis methods.}

While synthetic data has demonstrated its potential across both small- and large-scale LMs for a variety of general and specialized tasks, including code generation, it also poses several challenges that must be addressed. These challenges include a lack of data diversity \cite{wettig2024qurating}, the need to ensure the factuality and fidelity of the information \cite{wood2021fake,van2023synthetic}, and the potential to amplify existing biases or introduce new ones \cite{barbierato2022methodology,gupta2021transitioning}.

\subsection{Pre-Training}\label{sec:pre-training}
\subsubsection{Model Architectures}
Since the inception of the Transformer architecture for machine translation \cite{vaswani2017attention}, it has become the de facto backbone for a multitude of \done{LLMs} that address a wide range of downstream tasks. 
The Transformer and its derivatives owe their prominence to their exceptional ability to parallelize computation and their powerful representational capacities \cite{zhao2023survey,yoo2024hyperclova}. 
Through innovative scaling techniques, such as Mixture-of-Experts (MoE) \cite{shazeer2017outrageously,cai2024shortcut} and Depth-Up-Scaling (DUS) \cite{kim2023solar}, the capacity of Transformer-based LLMs has expanded to encompass hundreds of billions or even trillions of parameters. 
These scaled-up models have exhibited a range of emergent abilities \cite{kaplan2020scaling,hoffmann2022training,wei2022emergent}, such as instruction following \cite{ouyang2022training}, in-context learning \cite{dong2022survey}, and step-by-step reasoning \cite{wei2022chain,huang2022towards} that were previously unforeseen.

In the domain of code generation using LLMs, the architecture of contemporary models generally falls into one of two categories: encoder-decoder models, such as CodeT5 \cite{wang2021codet5}, CodeT5+ \cite{wang2023codet5+}, and CodeRL \cite{le2022coderl}; or decoder-only models, such as Codex \cite{chen2021evaluating}, StarCoder \cite{li2023starcoder}, Code Llama \cite{roziere2023code}, and CodeGemma \cite{codegemma_2024}. 
These architectures are depicted in Figure \ref{fig:architecture}(b) and (c), respectively. 
For a comprehensive overview, Table \ref{tab:encoder_decoder_models} details the encoder-decoder architectures, while Table \ref{tab:decoder_only_models} focuses on the decoder-only models utilized in code generation.
\begin{table}[t]
\caption{\done{The overview of \done{LLMs} with encoder-decoder architectures for code generation.} 
}
\label{tab:encoder_decoder_models}
\centering
\scalebox{0.8}{
\rotatebox{0}{
    \begin{tabular}{lllcccc} 
    \toprule
    \textbf{Model} & \textbf{Institution} & \textbf{Size} & \textbf{Vocabulary} & \textbf{\makecell[c]{Context\\ Window}} & \textbf{Date} & \textbf{Open Source} \\
    \midrule
     PyMT5\cite{clement2020pymt5}  & Microsoft & 374M &50K &1024+1024  & 2020-10 & \\
     PLBART\cite{ahmad2021unified} & UCLA & 140M &	50K &1024+1024 &  2021-03 & \CheckmarkBold  \\
     CodeT5 \cite{wang2021codet5} & Salesforce & 60M, 220M, 770M & 32K &512+256 & 2021-09 & \CheckmarkBold  \\
    JuPyT5\cite{chandel2022training}  &Microsoft  & 350M & 50K & 1024+1024 &2022-01&   \\
     AlphaCode\cite{li2022competition}& DeepMind & \makecell[l]{284M, 1.1B, 2.8B,\\ 8.7B, 41.1B} &	8K & 1536+768 & 2022-02 & \\
    CodeRL\cite{le2022coderl} &Salesforce& 770M &	32K &512+256 &2022-06&\CheckmarkBold \\
     ERNIE-Code\cite{chai2022ernie} &  Baidu & 560M & 250K &1024+1024 & 2022-12 & \CheckmarkBold\\
    PPOCoder\cite{shojaee2023execution}  & Virginia Tech & 770M &32K &512+256 &2023-01&   \\
    CodeT5+\cite{wang2023codet5+}& Salesforce & \makecell[l]{220M, 770M, 2B,\\ 6B, 16B} & 50K &2048+2048 & 2023-05 & \CheckmarkBold \\
    CodeFusion\cite{singh2023codefusion}& Microsoft & 75M & 32k	& 128+128 &2023-10& \CheckmarkBold \\
    AST-T5\cite{gong2024ast}  &UC Berkeley & 226M & 32k & 512+200/300 &2024-01& \CheckmarkBold \\
    \bottomrule
    \end{tabular}
}
}
\end{table}
\begin{table}[t]
\caption{
\done{The overview of \done{LLMs} with decoder-only architectures for code generation.} 
}
\label{tab:decoder_only_models}
\centering
\scalebox{0.75}{
\rotatebox{0}{
    \begin{tabular}{lllcccc} 
    \toprule
    \textbf{Model} & \textbf{Institution} & \textbf{Size} & \textbf{Vocabulary} & \textbf{\makecell[c]{Context\\ Window}} & \textbf{Date} & \textbf{Open Source} \\
    \midrule
GPT-C \cite{svyatkovskiy2020intellicode} & Microsoft & 366M      &60K	&1024	 &  2020-05 	 &   \\
CodeGPT \cite{lu2021codexglue}   & Microsoft & 124M      	&	50K &1024	 &  2021-02 	 &  \CheckmarkBold \\
GPT-Neo\cite{gpt-neo} & EleutherAI & \makecell[l]{125M, 1.3B, 2.7B} 	& 50k	&	2048 &  2021-03 &  \CheckmarkBold \\
  GPT-J \cite{gpt-j} & EleutherAI & 6B  &	50k &	2048 &  2021-05 &  \CheckmarkBold  \\         
  Codex \cite{chen2021evaluating}  & OpenAI  & \makecell[l]{12M, 25M, 42M, \\85M, 300M, 679M,\\ 2.5B, 12B} & -	& 4096	 &  2021-07 &   \\
  CodeParrot \cite{tunstall2022natural}   & Hugging Face  & 110M, 1.5B & 33k &	1024 &  2021-11 &  \CheckmarkBold \\
 PolyCoder \cite{xu2022systematic}    & CMU & 160M, 400M, 2.7B &	50k	&2048	 &  2022-02 	 &  \CheckmarkBold \\
  CodeGen \cite{nijkamp2022codegen}   & Salesforce & \makecell[l]{350M, 2.7B, 6.1B, \\16.1B}  &	51k 	&	2048 &  2022-03 &  \CheckmarkBold \\
  GPT-NeoX \cite{black2022gpt}   & EleutherAI  &   20B	&	50k &	2048 & 2022-04  & \CheckmarkBold  \\
  PaLM-Coder \cite{chowdhery2023palm}   & Google  &  8B, 62B, 540B  & 256k  &	 2048	 &   2022-04 &   \\
  InCoder \cite{fried2022incoder}   & Meta & 1.3B, 6.7B     &	50k	& 2049	 &  2022-04 	 &  \CheckmarkBold \\
  PanGu-Coder \cite{christopoulou2022pangu}    & Huawei & 317M, 2.6B &	42k & 1024	 &  2022-07 	 &   \\
  PyCodeGPT \cite{zan2022cert}   & Microsoft & 110M       &	32k & 1024    &  2022-06      & \CheckmarkBold  \\
  CodeGeeX \cite{zheng2023codegeex}  & Tsinghua & 13B  	&	52k & 2048	 &  2022-09 	 &  \CheckmarkBold  \\
  BLOOM \cite{le2023bloom}   & BigScience  &    176B &	 251k &	- &   2022-11 & \CheckmarkBold  \\
  ChatGPT \cite{gpt-3.5-turbo}   & OpenAI  &  - & - &	16k & 2022-11 &  \CheckmarkBold \\
  SantaCoder \cite{allal2023santacoder}   & Hugging Face & 1.1B  &	  49k  	&2048	 &  2022-12 	 &  \CheckmarkBold \\
  LLaMA \cite{touvron2023llama}   &  Meta &   \makecell[l]{6.7B, 13.0B, 32.5B, \\65.2B} & 32K &	2048 & 2023-02  & \CheckmarkBold  \\
  GPT-4 \cite{achiam2023gpt}   & OpenAI  &    -   	& -	&	32K &  2023-03 &   \\
  CodeGen2 \cite{nijkamp2023codegen2}   & Salesforce  &  1B, 3.7B, 7B, 16B & 51k &	2048 &  2023-05 & \CheckmarkBold  \\
  replit-code \cite{replit-code}   & replit  &    3B   &	33k	& 2048 &  2023-05 &  \CheckmarkBold \\
  StarCoder \cite{li2023starcoder}   & Hugging Face & 15.5B  &	 49k  	&8192	 &  2023-05 	 & \CheckmarkBold  \\ 
  WizardCoder \cite{luo2023wizardcoder}   &  Microsoft &  15B, 34B & 49k  & 8192 &  2023-06 &  \CheckmarkBold \\
  phi-1 \cite{gunasekar2023textbooks}   & Microsoft & 1.3B    &	51k  	&2048	 &  2023-06 	 &  \CheckmarkBold \\
  CodeGeeX2 \cite{zheng2023codegeex}   & Tsinghua  &     6B  &	65k	&	8192 & 2023-07  &  \CheckmarkBold \\
  PanGu-Coder2 \cite{shen2023pangu}   &  Huawei  &  15B &  42k & 1024  & 2023-07  &   \\
  Llama 2 \cite{touvron2023llama2}   & Meta  &  7B, 13B, 70B   & 32K  &	4096 &  2023-07 &  \CheckmarkBold \\
  OctoCoder \cite{muennighoff2023octopack}   & Hugging Face  &  15.5B & 49k	& 8192 &  2023-08 &  \CheckmarkBold\\
  Code Llama \cite{roziere2023code}   & Meta  & 7B, 13B, 34B &	32k & 16384 &  2023-08 &  \CheckmarkBold \\
  CodeFuse \cite{liu2023mftcoder}  & Ant Group & 350M, 13B, 34B  &	101k &4096	 &  2023-09 	 &  \CheckmarkBold \\
  phi-1.5 \cite{li2023textbooks}   & Microsoft  & 1.3B & 51k	& 2048 & 2023-09 & \CheckmarkBold \\
  CodeShell \cite{xie2024codeshell}   & Peking University & 7B & 70k	& 8192 &  2023-10 & \CheckmarkBold  \\
 Magicoder \cite{wei2023magicoder}   & UIUC  & 7B & 32k & 16384 & 2023-12 & \CheckmarkBold  \\
  AlphaCode 2 \cite{alphacode2}  &  Google DeepMind & - &	- & - &	  2023-12 &   \\ 
  StableCode \cite{pinnaparaju2024stable}   &  StabilityAI  &   3B   &	 50k	& 16384 & 2024-01  &  \CheckmarkBold \\
 WaveCoder \cite{yu2023wavecoder}   & Microsoft  & 6.7B & 32k	&	16384 & 2023-12  & \CheckmarkBold  \\
  phi-2 \cite{phi-2}   & Microsoft  &     2.7B  	&	51k & 2048 &  2023-12 &  \CheckmarkBold \\
  DeepSeek-Coder \cite{guo2024deepseek}  & DeepSeek  &  1.3B, 6.7B, 33B &	32k	& 16384 &  2023-11 &  \CheckmarkBold \\
 StarCoder 2 \cite{lozhkov2024starcoder}   &  Hugging Face &  15B & 49k & 16384 & 2024-02  &  \CheckmarkBold \\
 Claude 3 \cite{claude3}  & Anthropic & - & - &	200K & 2024-03 &   \\
 CodeGemma \cite{codegemma_2024}  & Google  &2B, 7B  & 25.6k &  8192	& 2024-04 &  \CheckmarkBold \\
 Code-Qwen \cite{codeqwen}  & Qwen Group & 7B & 92K & 65536 &2024-04& \CheckmarkBold \\
 Llama3 \cite{llama3}   & Meta & 8B, 70B & 128K & 8192 & 2024-04 & \CheckmarkBold \\
 StarCoder2-Instruct \cite{starcoder2instruct} & Hugging Face &  15.5B  & 49K & 16384 & 2024-04  & \CheckmarkBold  \\	 
 Codestral \cite{codestral} & Mistral AI & 22B & 33k & 32k & 2024-05 & \CheckmarkBold \\
    \bottomrule
    \end{tabular}
}
}
\end{table}

\subsubsection{Pre-training Tasks}
In the initial phase, language models for code generation are typically trained from scratch using datasets consisting of manually annotated pairs of natural language descriptions and corresponding code snippets, within a supervised learning framework. 
However, manual annotation is not only laborious and time-consuming, but the efficacy of the resulting models is also constrained by both the volume and the quality of the available annotated data. 
This limitation is especially pronounced in the context of low-resource programming languages, such as Swahili and Yoruba, where annotated examples are scarce \cite{chen2022transferability,cassano2023knowledge}.
In light of these challenges, there has been a shift towards an alternative training strategy that involves pre-training models on extensive and unlabelled code corpora. 
This method is aimed at imbuing the models with a broad understanding of programming knowledge, encompassing elements like identifiers, code structure, and underlying semantics \cite{chen2021evaluating}. 
In this regard, two pre-training tasks have gained prominence for their effectiveness, namely Causal Language Modeling (CLM), also known as unidirectional language modeling or next-token prediction, and Denoising Autoencoding (DAE).
The CLM task can be applied to both decoder-only and encoder-decoder model architectures, while DAE tasks are specifically designed for encoder-decoder frameworks. 
It should also be noted that there is a variety of additional auxiliary pre-training tasks that can further enhance model performance. These include Masked Identifier Prediction, Identifier Tagging, Bimodal Dual Generation \cite{wang2021codet5}, Text-Code Matching, and Text-Code Contrastive Learning \cite{wang2023codet5+}. These tasks contribute to a more nuanced and comprehensive pre-training process, equipping the models with the capabilities necessary to handle a wide range of code generation scenarios.

\textbf{Causal Language Modeling.}  
In decoder-only LLMs, given a sequence of tokens $\mathbf{x}=\{x_1,\dots,x_n\}$, the CLM task refers to autoregressively predict the target tokens $x_i$ based on the preceding tokens $x_{<i}$ in a sequence. The causal language modeling objective for training decoder LLMs is to minimize the following likelihood: 
\begin{equation}
\begin{aligned}
    \mathcal{L}_{CLM}^{Decoder-only}(\mathbf{x})=-\log(\prod_{i=1}^n P_{\theta}(x_i\mid\mathbf{x}_{<i}))=\sum_{i=1}^n -\log P_{\theta}(x_i\mid\mathbf{x}_{<i})
\label{eq:clm_decoder_only}
\end{aligned}
\end{equation}
where $\mathbf{x}_{<i}$ represents the sequence of preceding tokens $\{x_1,\dots,x_{i-1}\}$ before $\mathbf{x}_{i}$ in the input, $\theta$ denotes the model parameters. The conditional probability $P_{\theta}(x_i|\mathbf{x}_{<i}))$ is modeled by adding a causal attention mask to the multi-head self-attention matrix of each Transformer block. To be specific, causal attention masking is implemented by setting the lower triangular part of the matrix to 0 and the remaining elements to $-\infty$, ensuring that each token $x_i$ attends only to its predecessors and itself. 
On the contrary, in encoder-decoder LLMs, a pivot token $x_k$ is randomly selected in a sequence of tokens and then regarding the context before it as the source sequence $\mathbf{x}_{in}=\{x_1,\dots,x_k\}$ of the encoder and the sequence after it as the target output $\mathbf{x}_{out}=\{x_{k+1},\dots,x_n\}$ of decoder. Formally, the causal language modeling objective for training encoder-decoder LLMs is to minimize loss function as follows:
\begin{equation}
\begin{aligned}
    \mathcal{L}_{CLM}^{Encoder-Decoder}(\mathbf{x})=-\log(\prod_{i=k+1}^n P_{\theta}(x_i\mid\mathbf{x}_{\leq k}, \mathbf{x}_{<i}))=\sum_{i=k+1}^n -\log P_{\theta}(x_i\mid\mathbf{x}_{\leq k}, \mathbf{x}_{<i})
\label{eq:clm_encoder_decoder}
\end{aligned}
\end{equation}
where $\mathbf{x}_{\leq k}$ is the source sequence input and $\mathbf{x}_{<i}$ denotes the target sequence autoregressively generated so far. 
During the inference phase, pre-trained LLMs that have been trained on large-scale code corpus can generate code in a zero-shot manner without the need for fine-tuning. This is achieved through the technique of prompt engineering, which guides the model to produce the desired output\footnote{For more information on prompt engineering, visit \href{https://www.promptingguide.ai}{https://www.promptingguide.ai}} \cite{radford2019language,brown2020language}. 
Additionally, recent studies have explored the use of few-shot learning, also referred to as in-context learning, to enhance model performance further \cite{li2023towards,patel2023evaluating}.


\textbf{Denoising Autoencoding.}  
In addition to causal language modeling (CLM), the denoising autoencoding (DAE) task has been extensively applied in pre-training encoder-decoder architectures for code generation, such as PLBART \cite{ahmad2021unified}, CodeT5 \cite{wang2021codet5}, and its enhanced successor, CodeT5+ \cite{wang2023codet5+}. 
Following T5 \cite{raffel2020exploring} and CodeT5 \cite{wang2021codet5}, the DAE refers to initially perturbing the source sequence by introducing randomly masked spans of varying lengths. 
This corrupted sequence serves as the input for the encoder. Subsequently, the decoder employs an autoregressive strategy to reconstruct the masked spans, integrating sentinel tokens to facilitate the generation process. 
This method has proven effective in improving the model's ability to generate semantically and syntactically accurate code by learning robust contextual representations \cite{wang2021codet5,wang2023codet5+}.
Formally, the denoising autoencoding objective for training encoder-decoder LLMs is to minimize the following likelihood:
\begin{equation}
    \mathcal{L}_{DAE}^{Encoder-Decoder}(\mathbf{x})= \sum_{i=1}^k -\log P_\theta(\mathbf{x}_{i}^{masked\_spans}\mid\mathbf{x}^{\backslash masked\_spans}, \mathbf{x}_{<i}^{masked\_spans})
\label{eq:dae}
\end{equation}
where $\theta$ denotes the model parameters, $\mathbf{x}^{\backslash masked\_spans}$ is the noisy input with masked spans, $\mathbf{x}^{masked\_spans}$ is the masked spans to predict from the decoder with $k$ denoting the number of tokens in $\mathbf{x}^{masked\_spans}$, and $\mathbf{x}_{<i}^{masked\_spans}$ is the span sequence autoregressively generated so far. 
Compared with CLM, the DAE task presents a more challenging scenario, as it necessitates a deeper understanding and capture of the intrinsic semantic relationships among token sequences by LLMs \cite{raffel2020exploring}.
  

\begin{figure*}[t]
\centering
\includegraphics[width=\linewidth]{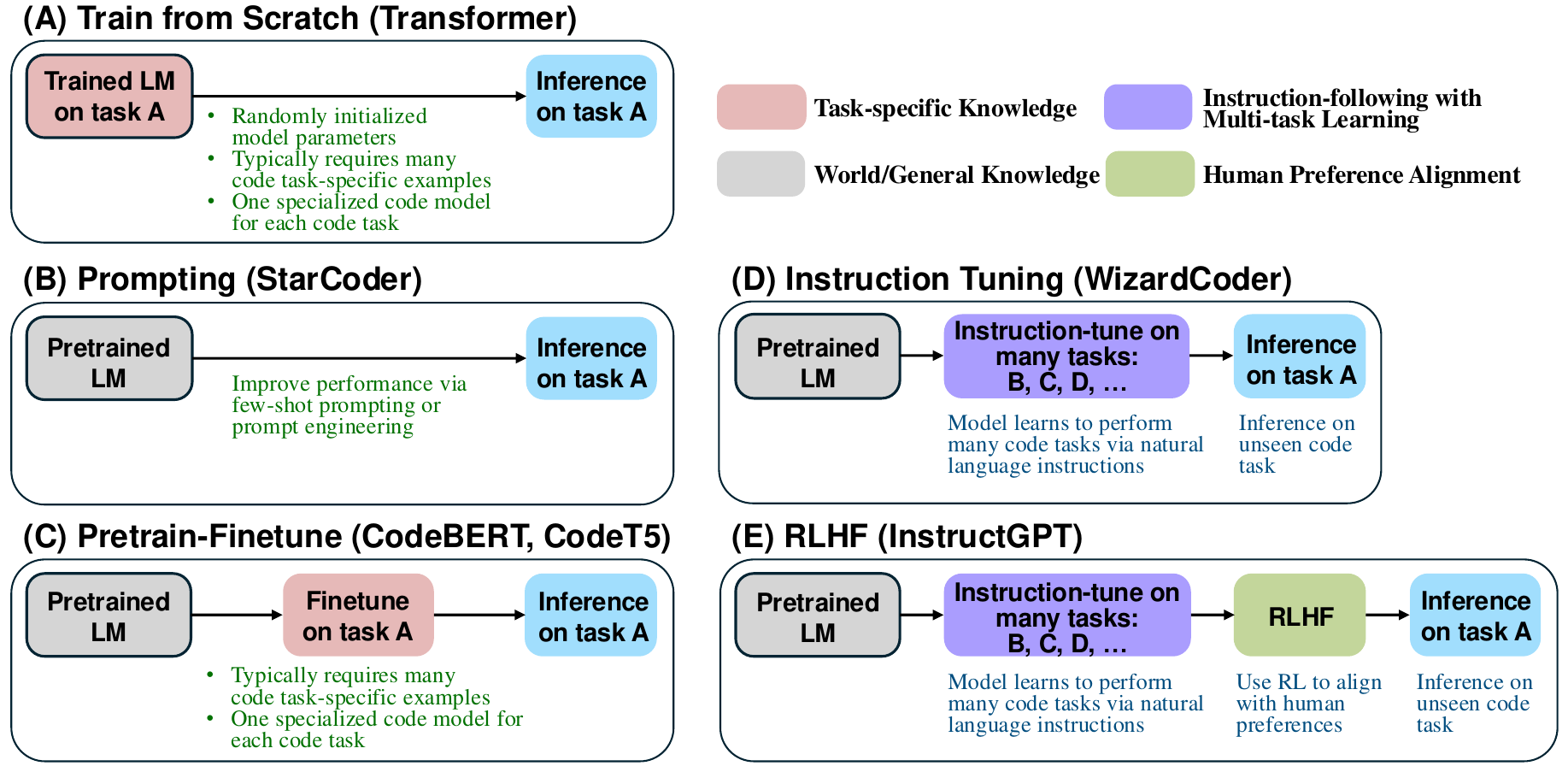}
\caption{\done{Comparison of instruction tuning with various fine-tuning strategies and prompting for code tasks, adapted from \cite{wei2021finetuned}. For (A), which involves training a Transformer from scratch, please refer to \cite{ahmad2020transformer} for its use in source code summarization task. In the case of (E), we utilize a representative RLHF \cite{ouyang2022training} as an example. Additional reinforcement learning methods, such as DPO \cite{rafailov2024direct}, are also applicable at this stage. 
}}
\label{fig:various_ft}
\end{figure*}

\subsection{Instruction Tuning}\label{sec:instruction_tuning}
\begin{figure*}[t]
\centering
\includegraphics[width=\linewidth]{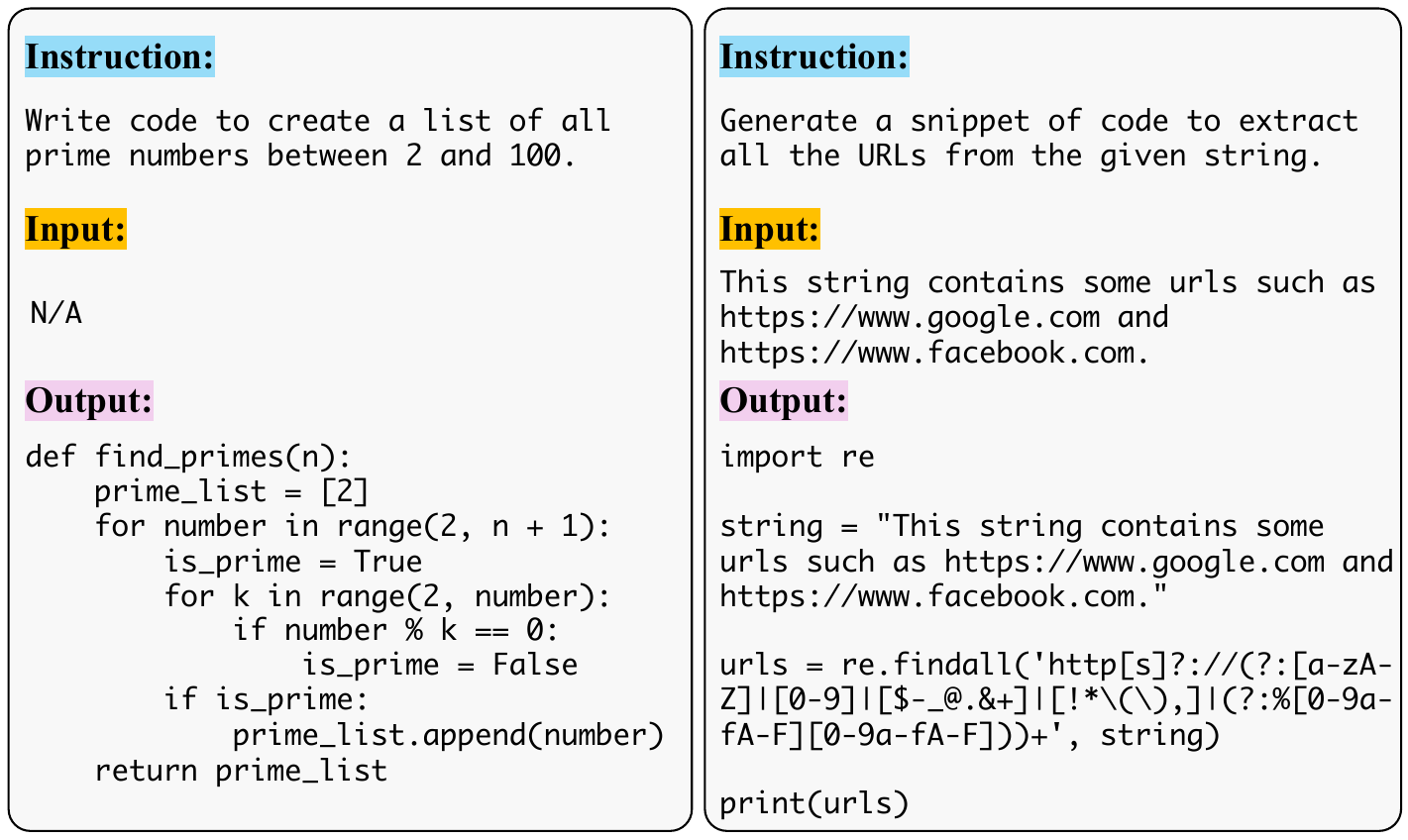}
\caption{Two exemplars of instruction data sampled from Code Alpaca \cite{codealpaca} used to instruction-tune pre-trained code LLM to enhance their alignment with natural language instructions.
The instruction corpus encompasses a variety of tasks, each accompanied by distinct instructions, such as prime numbers generation and URLs extraction.}
\label{fig:instruction}
\end{figure*}

After pre-training \done{LLMs} on large-scale datasets, the next phase typically involves augmenting the model's ability \done{to process and follow various instructions, known as instruction tuning}. 
\done{Instruction tuning generally refers to the supervised fine-tuning of pre-trained LLMs using datasets comprised of structured examples framed as various natural language instructions \cite{wei2021finetuned,ouyang2022training,iyer2022opt,zhang2023instruction}.
The comparison of instruction tuning with various fine-tuning strategies and prompting for code tasks is depicted in Figure \ref{fig:various_ft}.
} 
Two exemplars of instruction data sampled from Code Alpaca \cite{codealpaca} are demonstrated in Figure \ref{fig:instruction}.
It capitalizes on the heterogeneity of instruction types, positioning instruction tuning as a form of multi-task prompted training that significantly enhances the model's generalization to unseen tasks \cite{wei2021finetuned,sanh2022multitask,ouyang2022training,chung2024scaling}.

In the realm of code generation, natural language descriptions serve as the instructions guiding the model to generate corresponding code snippets. 
Consequently, a line of research on instruction tuning LLMs for code generation has garnered substantial interest across academia and industry. 
To perform instruction tuning, instruction data are typically compiled from source code with permissive licenses \cite{husain2019codesearchnet,kocetkov2022stack,lozhkov2024starcoder} (refer to Section \ref{sec:instruction_data}) or are constructed from synthetic code data \cite{luo2023wizardcoder,wei2023magicoder,starcoder2instruct} (refer to Section \ref{sec:data_synthesis}). 
These datasets are then utilized to fine-tune LLMs through a supervised learning paradigm.
However, the substantial computational resources required for full parameter fine-tuning (FFT) LLM pose a notable challenge, particularly in scenarios with constrained resources \cite{ding2022delta,lialin2023scaling}. 
To mitigate this issue, parameter-efficient fine-tuning (PEFT) has emerged as a compelling alternative strategy, gaining increasing attention for its potential to reduce resource consumption \cite{ding2022delta}.
In the following subsection, we categorize existing works based on their instruction-tuning strategies to provide a comprehensive and systematic review.

\subsubsection{Full Parameter Fine-tuning}
Full parameter fine-tuning (FFT) involves updating all parameters within a pre-trained model, as shown in Figure \ref{fig:finetune}(a). This approach is often preferred when ample computational resources and substantial training data are available, as it typically leads to better performance.
\cite{wang2021codet5} introduces an encoder-decoder pre-trained language model for code generation, named CodeT5+. They instruction-tune this model on a dataset comprising 20k instruction samples from Code Alpaca \cite{codealpaca}, resulting in an instruction-following model called InstructCodeT5+, which exhibited improved capabilities in code generation.
\cite{luo2023wizardcoder} leverages the Evol-Instruct data synthesis technique from WizardLM \cite{xu2023wizardlm} to evolve 20K code Alpaca \cite{codealpaca} instruction samples into a 78K code instruction dataset. This enriched dataset is then used to fine-tune the StarCoder base model, resulting in WizardCoder, which showcases notable advancements in code generation.
In a similar vein, inspired by the successes of WizardCoder \cite{luo2023wizardcoder} and RRHF \cite{yuan2023rrhf}, Pangu-Coder 2 \cite{shen2023pangu} applies the Evol-Instruct method to generate 68k high-quality instruction samples from the initial 20k Code Alpaca \cite{codealpaca} instruction samples. Additionally, they introduces a novel reinforcement learning via Rank Responses to align Test \& Teacher Feedback (RRTF), which further enhances the performance of Pangu-Coder 2 in code generation.
Diverging from synthetic instruction data generation methods, OctoPack \cite{muennighoff2023octopack} utilizes real-world data by curating CommitPack from the natural structure of Git commits, which inherently pair code changes with human-written instructions. 
This dataset, consisting of 4 terabytes of Git commits across 350 programming languages, is employed to fine-tune StarCoder \cite{li2023starcoder} and CodeGeeX2 \cite{zheng2023codegeex}, leading to the instruction-following code models of OctoCoder and OctoGeeX for code generation, respectively.
The most recent innovation comes from Magicoder \cite{wei2023magicoder}, who proposes OSS-INSTRUCT, a novel data synthesis method that leverages open-source code snippets to generate high-quality instruction data for code generation. This approach seeks to reduce the bias often present in synthetic data generated by LLM.
In line with OSS-INSTRUCT, the BigCode team introduces StarCoder2-15B-instruct \cite{starcoder2instruct}, which they claim to be the first entirely self-aligned \done{LLM} for code generation, trained with a fully permissive and transparent pipeline.
Moreover, \cite{codegemma_2024} harnesses open-source mathematics datasets, such as MATH \cite{hendrycks2021measuring} and GSM8k \cite{cobbe2021training}, along with synthetically generated code following the OSS-INSTRUCT \cite{wei2023magicoder} paradigm, to instruction-tune CodeGemma 7B, yielding exceptional results in mathematical reasoning and code generation tasks.

\subsubsection{Parameter-Efficient Fine-tuning}\label{sec:peft}
\begin{figure*}[t]
\centering
\includegraphics[width=\linewidth]{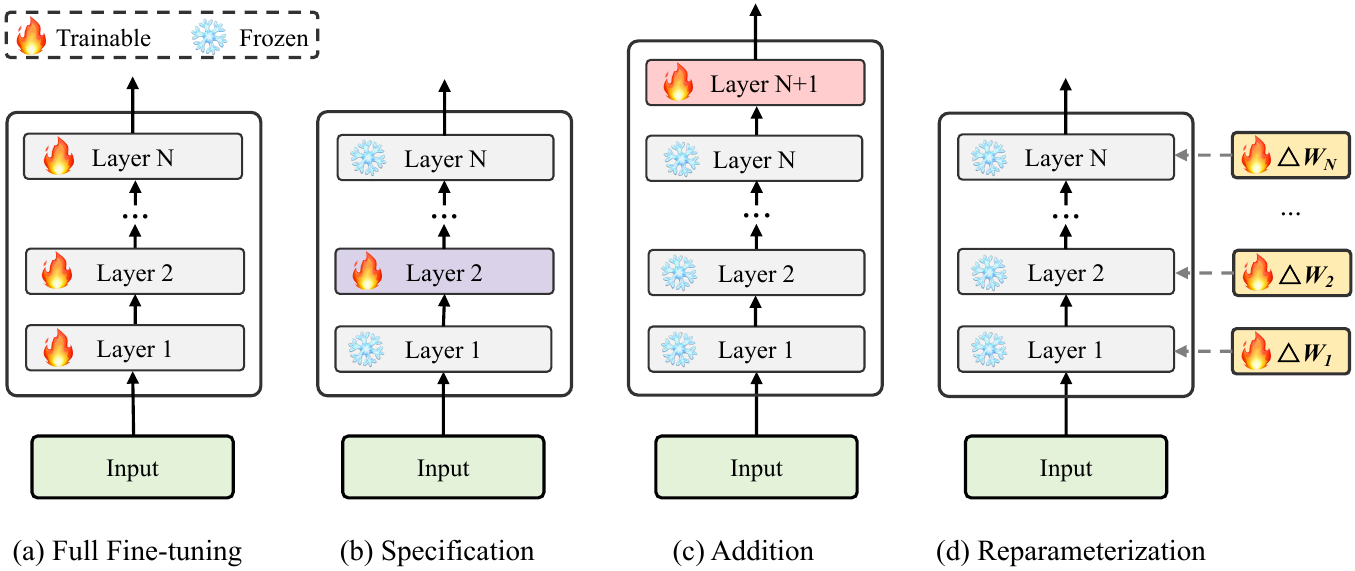}
\caption{An illustration of full parameter fine-tuning (FFT) and parameter-efficient fine-tuning (PEFT) methods.  
(a) refers to the Full Fine-tuning method, which updates all parameters of the base model during fine-tuning. 
(b) stands for the Specification-based PEFT method that conditionally fine-tunes a small subset of the model parameters while freezing the rest of the model, e.g. BitFit \cite{zaken2021bitfit}.
(c) represents the Addition-based PEFT method that fine-tunes the incremental parameters introduced into the base model or input, e.g. Adapter \cite{houlsby2019parameter}, Prefix-tuning
\cite{li2021prefix}, and Prompt-tuning \cite{lester2021power}.
(d) symbolizes the Reparameterization-based method which reparameterizes existing model parameters by low-rank transformation, e.g. LoRA \cite{hu2021lora}, QLoRA \cite{dettmers2024qlora}, and AdaLoRA \cite{zhang2023adaptive}.
}
\label{fig:finetune}
\end{figure*}

To mitigate the extensive computational and resource demands inherent in fine-tuning \done{LLMs}, the concept of parameter-efficient fine-tuning (PEFT) has emerged to focus on updating a minimal subset of parameters, which may either be a selection of the model's parameters or an array of additional parameters specifically introduced for the tuning process \cite{ding2022delta,lialin2023scaling}. The categorization of these methods is depicted in Figure \ref{fig:finetune}(b), (c), and (d). 
A plethora of innovative PEFT approaches have been developed, among which BitFit \cite{zaken2021bitfit}, Adapter \cite{houlsby2019parameter}, Prompt tuning \cite{lester2021power}, Prefix-tuning \cite{li2021prefix}, LoRA \cite{hu2021lora}, IA$^3$ \cite{liu2022few}, QLoRA \cite{dettmers2024qlora}, and AdaLoRA \cite{zhang2023adaptive} are particularly noteworthy.
A seminal study in this field, LoRA \cite{hu2021lora}, proposes a parameter update mechanism for a pre-trained weight matrix — such as those found in the key or value projection matrices of a Transformer block's multi-head self-attention layer — by factorizing the update into two low-rank matrices. Crucially, all original model parameters remain frozen, with only the pair of low-rank matrices being trainable. 
After fine-tuning, the product of these low-rank matrices can be seamlessly incorporated into the existing weight matrix through an element-wise addition. This process can be formally described as:
\begin{equation}
\begin{aligned}
    (\mathbf{W}_0 + \Delta\mathbf{W})x = \mathbf{W}_0x + \Delta\mathbf{W}x = \mathbf{W}_0^{frozen}x + \underbrace{\frac{\alpha}{r}\mathbf{B}^{trainable}_{up}\mathbf{A}^{trainable}_{down}}_{\Delta\mathbf{W}}x
\end{aligned}
\end{equation}
where $\mathbf{W}_0 \in \mathbb{R}^{d\times k}$ denotes a pre-trained weight matrix, $\mathbf{B}^{trainable}_{up} \in \mathbb{R}^{d\times r}$ and $\mathbf{A}^{trainable}_{down} \in \mathbb{R}^{r\times k}$ are two trainable low-rank matrixes and initialized by a zero matrix and a random Gaussian distribution $\mathcal{N}(0,\sigma^{2})$ respectively, to ensure $\Delta\mathbf{W}=0$ at the beginning of training. The rank $r \ll \min(d, k)$, the $\frac{\alpha}{r}$ is a scaling coefficient to balance the importance of the LoRA module, like a learning rate.

Despite the advancements in PEFT methods, their application in code generation remains limited. For instance, \cite{codeup} pioneered the use of parameter-efficient instruction-tuning on a Llama 2 \cite{touvron2023llama2} model with a single RTX 3090 GPU, leading to the development of a multilingual code generation model called CodeUp. 
More recently, ASTRAIOS \cite{zhuo2024astraios} conducted a thorough empirical examination of parameter-efficient instruction tuning for code comprehension and generation tasks. This study yielded several perceptive observations and conclusions, contributing valuable insights to the domain.

\subsection{Reinforcement Learning with Feedback}\label{sec:reinforcement_learning}
\done{LLMs} have exhibited remarkable instruction-following capabilities through instruction tuning. However, they often produce outputs that are unexpected, toxic, biased, or hallucinated outputs that do not align with users' intentions or preferences \cite{ouyang2022training,wang2023aligning,ji2023ai}. 
Consequently, aligning LLMs with human preference has emerged as a pivotal area of research. A notable work is InstructGPT \cite{ouyang2022training}, which further fine-tunes an instruction-tuned model utilizing reinforcement learning with human feedback (RLHF) on a dataset where labelers have ranked model outputs in order of quality, from best to worst. 
This method has been instrumental in the development of advanced conversational language models, such as ChatGPT \cite{gpt-3.5-turbo} and Bard \cite{manyika2023overview}.
Despite its success, acquiring high-quality human preference ranking data is a resource-intensive process \cite{lee2023rlaif}. 
To address this, Reinforcement Learning from AI Feedback (RLAIF) \cite{bai2022constitutional,lee2023rlaif} has been proposed to leverage powerful off-the-shelf LLMs (e.g., ChatGPT \cite{gpt-3.5-turbo} and GPT-4 \cite{achiam2023gpt}) to simulate human annotators by generating preference data.

Building on RLHF's success, researchers have explored reinforcement learning with feedback to enhance code generation in LLMs. 
Unlike RLHF, which relies on human feedback, this approach employs compilers or interpreters to automatically provide feedback on code samples through code execution on unit test cases, catalyzing the advancement of this research domain.
CodeRL \cite{le2022coderl} introduced an actor-critic reinforcement learning framework for code generation. In this setup, the language model serves as the actor-network, while a token-level functional correctness reward predictor acts as the critic. Generated code is assessed through unit test signals from a compiler, which can indicate compiler errors, runtime errors, unit test failures, or passes.
CompCoder \cite{wang2022compilable} enhances code compilability by employing compiler feedback, including language model fine-tuning, compilability reinforcement, and compilability discrimination strategies.
Subsequently, PPOCoder \cite{shojaee2023execution} integrates pre-trained code model CodeT5 \cite{wang2021codet5} with Proximal Policy Optimization (PPO) \cite{schulman2017proximal}. This integration not only utilizes execution (\textit{i.e.}, compilers or interpreters) feedback to assess syntactic and functional correctness but also incorporates a reward function that evaluates the syntactic and semantic congruence between abstract syntax tree (AST) sub-trees and data flow graph (DFG) edges in the generated code against the ground truth. 
Additionally, the framework applies a KL-divergence penalty to maintain fidelity between the actively learned policy and the referenced pre-trained model, enhancing the optimization process.
More recently, RLTF \cite{liu2023rltf} has proposed an online reinforcement learning framework that provides fine-grained feedback based on compiler error information and location, along with adaptive feedback that considers the ratio of passed test cases.

Despite these successes, reinforcement learning algorithms face inherent limitations such as inefficiency, instability, extensive resource requirements, and complex hyperparameter tuning, which can impede the performance and scalability of LLMs. 
To overcome these challenges, recent studies have introduced various variants of RL methods that do not rely on PPO, including DPO \cite{rafailov2024direct}, RRHF \cite{yuan2023rrhf}, and sDPO \cite{kim2024sdpo}. 
In essence, these methods aim to maximize the likelihood between the logarithm of conditional probabilities of preferred and rejected responses, which may be produced by LLMs with varying capabilities.
Inspired by RRHF \cite{yuan2023rrhf}, PanGu-Coder 2 \cite{shen2023pangu} leverages a novel framework, Reinforcement Learning via Rank Responses to align Test \& Teacher Feedback (RRTF), significantly enhancing code generation capabilities, as evidenced by \texttt{pass@1} of 62.20\% on the HumanEval benchmark.

Taking a step forward, the integration of more non-differentiable code features, such as coding style \cite{markovtsev2019style,chen2023duetcs} and readability \cite{buse2009learning}, into the reinforcement learning feedback for LLM-based code generation, presents an exciting avenue for future research.

\subsection{Prompting Engineering}\label{sec:prompting}
Large-scale language models (LLMs) such as GPT-3 and its successors have been trained on large-scale data corpora, endowing them with substantial world knowledge \cite{brown2020language,wei2021finetuned,ouyang2022training}. 
\done{
Despite this, crafting an effective prompting as a means of communicating with LLMs to harness their full potential remains a long-standing challenge \cite{liu2023pre}.
Recent advancements in prompting engineering have expanded the capabilities of LLMs, enabling more sophisticated task completion and enhancing both reliability and performance.} 
Notable techniques include Chain-of-Thought (CoT) \cite{wei2022chain}, Self-Consistency \cite{wang2022self}, Tree-of-Thought (ToT) \cite{yao2024tree}, Program of Thoughts (PoT) \cite{chen2022program}, Reasoning via Planning (RAP) \cite{hao2023reasoning}, ReAct \cite{yao2023react}, Self-Refine \cite{madaan2024self}, Reflexion \cite{shinn2024reflexion}, and LATS \cite{zhou2023language}.
\done{For instance, CoT significantly improves the LLMs' ability to perform complex reasoning by providing a few chain-of-thought demonstrations as exemplars in prompting.}

\begin{figure*}[t]
\centering
\includegraphics[width=0.95\linewidth]{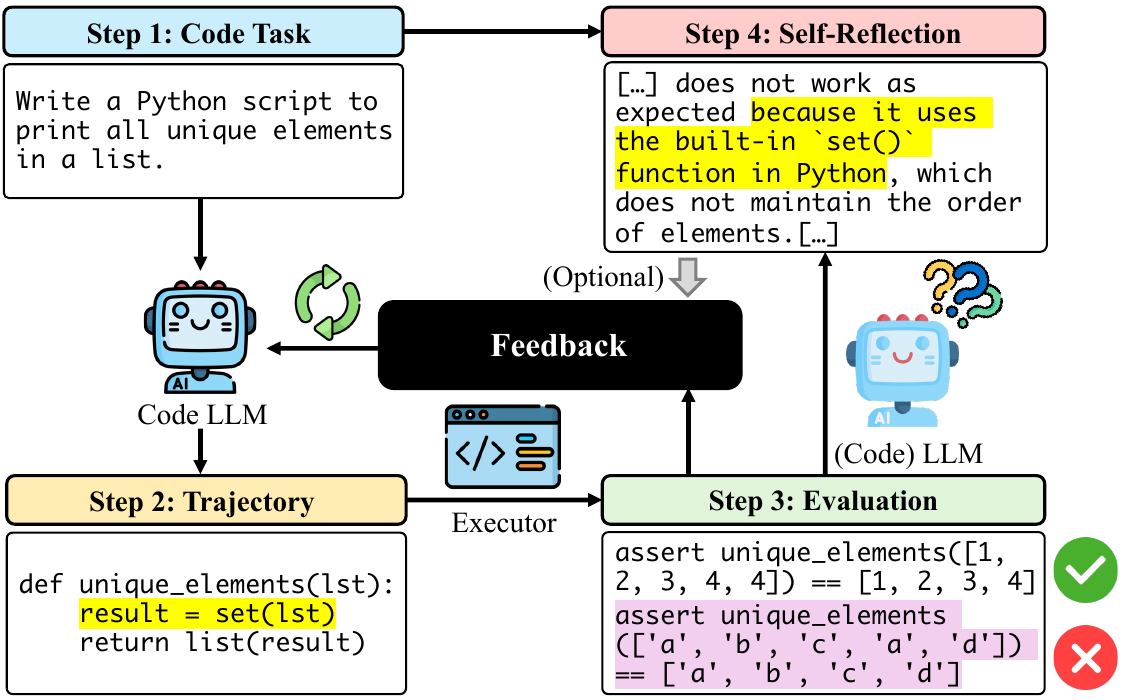}
\caption{An illustration of the self-improving code generation pipeline using prompts for \done{LLMs}. This process incorporates iterative self-refinement by integrating execution outcomes and includes an optional self-reflection mechanism to enhance generation quality.
}
\label{fig:reflection}
\end{figure*}

Prompting engineering is particularly advantageous as it bypasses the need for additional training and can significantly elevate performance. 
\done{Consequently, numerous studies have leveraged this technique for iterative and self-improving (refining) code generation within proprietary LLMs such as ChatGPT and GPT-4.} 
Figure \ref{fig:reflection} illustrates the general pipeline for self-improving code generation with LLMs.
For instance, Self-Debugging \cite{chen2023teaching} involves prompting an LLM to iteratively refine a predicted program by utilizing feedback composed of code explanations combined with execution results, which assists in identifying and rectifying errors. 
When unit tests are unavailable, this feedback can rely solely on code explanations. 
\done{
Similarly, LDB \cite{zhong2024ldb} prompts LLMs to refine generated code by incorporating debugging feedback, which consists of the evaluation of the correctness of variable values throughout runtime execution, as assessed by the LLMs.
}
In parallel, SelfEvolve \cite{jiang2023selfevolve} employs a two-stage process where LLMs first generate domain-specific knowledge for a problem, followed by a trial code. This code is then iteratively refined through interactive prompting and execution feedback.
An empirical investigation by \cite{olausson2023self} provides a comprehensive analysis of the self-repairing capabilities for code generation in models like Code Llama, GPT-3.5, and GPT-4, using problem sets from HumanEval and APPS. This study yields a series of insightful observations and findings, shedding light on the self-refinement effectiveness of these LLMs.
Moreover, Reflexion \cite{shinn2024reflexion} introduces a general approach for code generation wherein LLM-powered agents engage in verbal self-reflection on task feedback signals, storing these reflections in an episodic memory buffer to inform and improve decision-making in subsequent interactions.
LATS \cite{zhou2023language} adopts a novel strategy, utilizing LLMs as agents, value functions, and optimizers. It enhances decision-making by meticulously constructing trajectories through Monte Carlo Tree Search (MCTS) algorithms, integrating external feedback, and learning from experience. 
This approach has demonstrated remarkable results in code generation, achieving a \texttt{pass@1} of 94.4\% on the HumanEval benchmark with GPT-4.

Distinct from the aforementioned methods, CodeT \cite{chen2022codet} and LEVER \cite{ni2023lever} prompt LLMs to generate numerous code samples, which are then re-ranked based on execution outcomes to select the optimal solution. Notably, these approaches do not incorporate a self-refinement step to further improve code generation.

\subsection{Repository Level \& Long Context}\label{sec:repository_level}
In contemporary software engineering practices, modifications to a code repository are widespread and encompass a range of activities, including package migration, temporary code edits, and the resolution of GitHub issues. 
While \done{LLMs} showcase impressive prowess in function-level code generation, they often falter when grappling with the broader context inherent to a repository, such as import dependencies, parent classes, and files bearing similar names. 
These deficiencies result in suboptimal performance in repository-level code generation, as identified in recent studies \cite{shrivastava2023repository,shrivastava2023repofusion}.
The challenges faced by LLMs in this domain are primarily due to the following factors:
\begin{itemize}
    \item Code repositories typically contain intricate interdependencies scattered across various files, including shared utilities, configurations, and cross-API invocations, which arise from modular design principles \cite{zhang2023repocoder,bairi2023codeplan}.
    \item Repositories are characterized by their unique structures, naming conventions, and coding styles, which are essential for maintaining clarity and facilitating ongoing maintenance \cite{chen2023duetcs}.
    \item The vast context of an entire repository often exceeds the context length limitations of LLMs, thus hindering their ability to integrate comprehensive contextual information \cite{bairi2023codeplan}.
    \item LLMs may not have been adequately trained on extensive sets of repository data, such as proprietary software or projects that are still in development \cite{shrivastava2023repofusion}.
\end{itemize}

Given that the scope of a typical software repository encompasses hundreds of thousands of tokens, it is imperative to enhance the capacity of LLMs to handle extensive contexts when they are employed for repository-level code generation. Fortunately, recent advancements in positional encoding techniques, such as ALiBi \cite{press2021train} and RoPE \cite{su2024roformer}, have shown promise in improving the Transformer's ability to generalize from shorter training sequences to longer inference sequences \cite{zhao2023length}. This progress addresses the third challenge mentioned above to a certain degree, thereby enabling better contextualization of coding activities within full repositories.

To further refine LLMs for repository-level code completion, several innovative approaches have been introduced. 
RepoCoder \cite{zhang2023repocoder} leverages a similarity-based retrieval system within an iterative retrieval-generation paradigm to enrich the context and enhance code completion quality. 
In a similar vein, CoCoMIC \cite{ding2022cocomic} employs a cross-file context finder named CCFINDER to pinpoint and retrieve the most relevant cross-file contexts within a repository. 
RepoHyper \cite{phan2024repohyper} introduces a semantic graph structure, termed RSG, to encapsulate the expansive context of code repositories and uses an ``Expand and Refine'' retrieval method to obtain relevant code snippets. 
Moreover, a framework known as RLPG \cite{shrivastava2023repository} has been proposed to generate repository-level prompts that integrate the repository's structure with the relevant context across all files. 
However, the constant reliance on retrieval mechanisms has raised concerns regarding efficiency and robustness, as some retrieved contexts may prove unhelpful or harmful. In response, Repoformer \cite{wu2024repoformer} introduces a selective Retrieval-Augmented Generation (RAG) framework that judiciously bypasses retrieval when it is deemed redundant. This approach incorporates a self-supervised learning strategy that equips a code LLM with the ability to perform a self-assessment on the utility of retrieval for enhancing the quality of its output, thereby effectively utilizing potentially noisy retrieved contexts.

Additionally, RepoFusion \cite{shrivastava2023repofusion} has been developed to train models to combine multiple relevant contexts from a repository, aiming to produce more precise and context-aware code completions. 
In a novel approach, Microsoft's CodePlan \cite{bairi2023codeplan} frames repository-level coding tasks as a planning problem, generating a multi-step chain of edits (plan) where each step involves invoking an LLM on a specific code location, considering context from the entire repository, preceding code modifications, and task-specific instructions.

Advancing the state-of-the-art, \cite{zan2024codes} tackles the formidable challenge of NL2Repo, an endeavor that seeks to create a complete code repository from natural language requirements. To address this complex task, they introduce the CodeS framework, which strategically breaks down NL2Repo into a series of manageable sub-tasks using a multi-layer sketch approach. The CodeS framework comprises three distinct modules: 
1) RepoSketcher, for creating a directory structure of the repository based on given requirements; 
2) FileSketcher, for sketching out each file within that structure; and 
3) SketchFiller, for fleshing out the specifics of each function within the file sketches \cite{zan2024codes}.

Accordingly, a surge of benchmarks tailored for repository-level code generation has emerged, such as RepoEval \cite{zhang2023repocoder}, Stack-Repo \cite{shrivastava2023repofusion}, Repobench \cite{liu2023repobench}, EvoCodeBench \cite{li2024evocodebench}, SWE-bench \cite{jimenez2023swe}, CrossCodeEval \cite{ding2024crosscodeeval}, and SketchEval \cite{zan2024codes}. The detailed statistics and comparisons of these benchmarks are presented in Table \ref{tab:benchmark}.

Despite the progress made by these methods in repository-level code generation, significant challenges remain to be addressed. Programming developers are often required to invest considerable time in editing and debugging \cite{vaithilingam2022expectation,mozannar2022reading,shrivastava2023repofusion,barke2023grounded,bird2022taking}. However, the advent of LLM-powered coding agents, such as AutoCodeRover \cite{zhang2024autocoderover}, SWE-Agent \cite{swe-agent}, and OpenDevin \cite{OpenDevin}, has demonstrated their potential to tackle complex problems, paving the way for future exploration in this field (for more details, see Section \ref{sec:autonomous_agents}).

\subsection{Retrieval Augmented}\label{sec:retrieval_augmented}
\begin{figure*}[t]
\centering
\includegraphics[width=\linewidth]{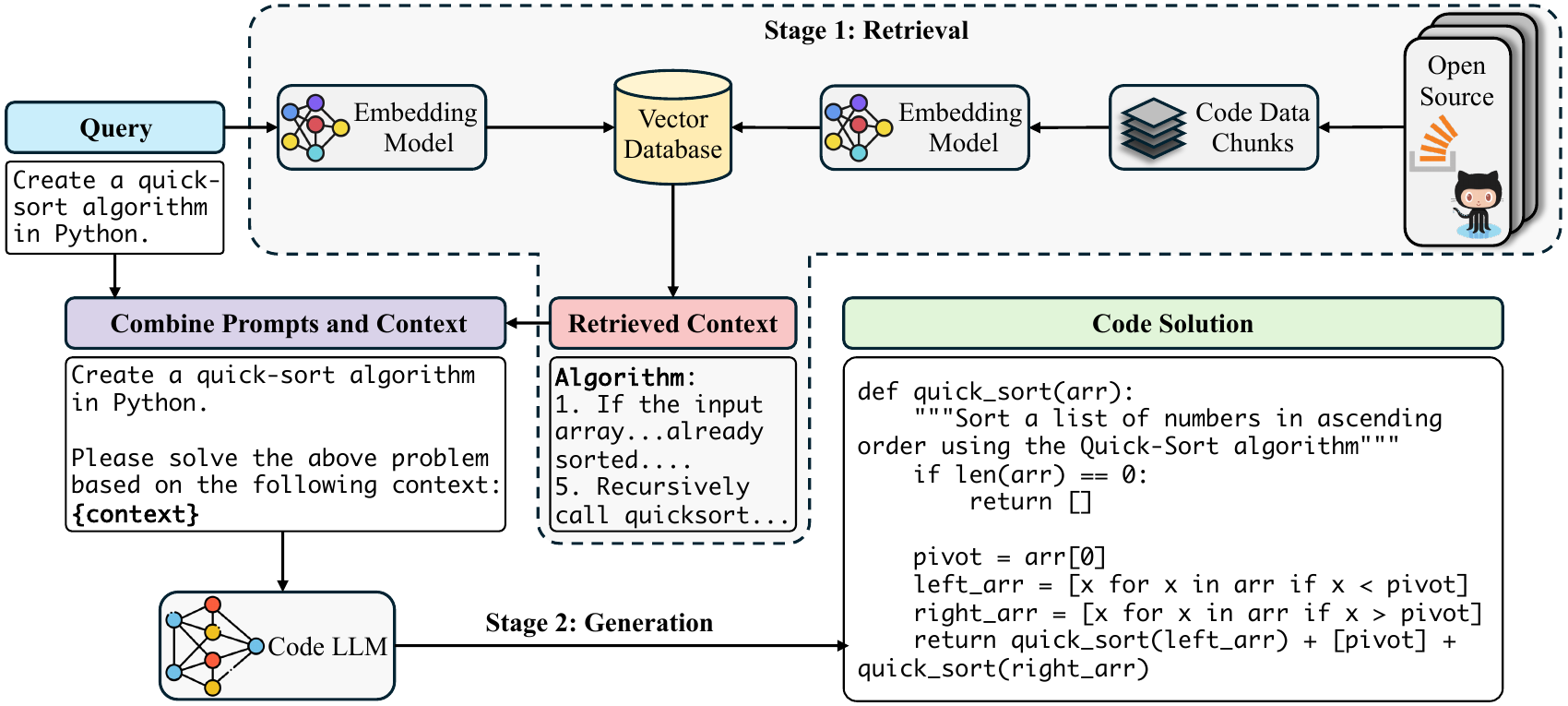}
\caption{A workflow illustration of the Retrieval-Augmented Code Generation (RACG). Upon receiving a query (instruction), the retriever selects the relevant contexts from a large-scale vector database. Subsequently, the retrieved contexts are merged with the query, and this combined input is fed into the generator (LLM) to produce the target code solution.
}
\label{fig:rag}
\end{figure*}
\done{LLMs} have exhibited impressive capabilities but are hindered by several critical issues such as hallucination \cite{liang2023holistic,zhang2023siren}, obsolescence of knowledge \cite{jang2022towards}, and non-transparent \cite{bommasani2021opportunities}, untraceable reasoning processes \cite{zhou2022least,wei2022chain,huang2023towards,gao2023retrieval}. 
While techniques like instruction-tuning (see Section \ref{sec:instruction_tuning}) and reinforcement learning with feedback (see Section \ref{sec:reinforcement_learning}) mitigate these issues, they also introduce new challenges, such as catastrophic forgetting and the requirement for substantial computational resources during training \cite{ovadia2023fine,gupta2024rag}.

Recently, Retrieval-Augmented Generation (RAG) has emerged as an innovative approach to overcoming these limitations by integrating knowledge from external databases. Formally defined, RAG denotes a model that, in response to queries, initially sources relevant information from an extensive corpus of documents, and then leverages this retrieved information in conjunction with the original query to enhance the response's quality and accuracy, especially for knowledge-intensive tasks. 
The RAG framework typically consists of a vector database, a retriever, a re-ranker, and a generator. 
It is commonly implemented using tools such as LangChain\footnote{LangChain facilitates the development of LLM-powered applications. \href{https://www.langchain.com}{https://www.langchain.com}} and LLamaIndex\footnote{LLamaIndex is a leading data framework for building LLM applications. \href{https://www.llamaindex.ai/}{https://www.llamaindex.ai}}. 
By performing continuous knowledge updates of the database and the incorporation of domain-specific data, RAG circumvents the need for re-training LLMs from scratch \cite{gao2023retrieval}. Consequently, RAG has substantially advanced LLM performance across a variety of tasks \cite{lewis2020retrieval,chen2024benchmarking}.

Due to the nature of code, code LLMs are also susceptible to the aforementioned issues that affect general-purpose LLMs. 
For instance, they may exhibit a hallucination phenomenon when instructions fall outside the scope of their training data or necessitate the latest programming packages. 
Given the dynamic nature of publicly available source-code libraries like PyTorch, which undergo frequent expansion and updates, deprecated calling methods can become a significant challenge. 
If Code LLMs are not updated in tandem with the latest functions and APIs, this can introduce potential errors and safety risks. 
Retrieval-Augmented Code Generation (RACG) stands as a promising solution to these concerns. A workflow illustration of the RACG is depicted in Figure \ref{fig:rag}.

Despite its potential, the adoption of RAG for code generation remains limited. 
Drawing inspiration from the common practice among programmers of referencing related code snippets, \cite{liu2020retrieval} introduced a novel retrieval-augmented mechanism with graph neural networks (GNNs), termed HGNN, which unites the advantages of similar examples retrieval with the generalization capabilities of generative models for code summarization, which is the reverse process of code generation.
\cite{parvez2021retrieval} pioneered a retrieval augmented framework named REDCODER for code generation by retrieving and integrating relevant code snippets from a source-code database, thereby providing supplementary context for the generation process. 
Subsequently, a retrieval-augmented code completion framework termed ReACC \cite{lu2022reacc} is proposed to leverage both lexical copying and semantic referencing of related code, achieving state-of-the-art performance on the CodeXGLUE benchmark \cite{lu2021codexglue}.
In the spirit of how programmers often consult textual resources such as code manuals and documentation to comprehend functionalities, DocPrompting \cite{zhou2022docprompting} explicitly utilizes code documentation by retrieving the relevant documentation pieces based on a natural language query and then generating the target code by blending the query with the retrieved information.

More recently, RepoCoder \cite{zhang2023repocoder}, an iterative retrieval-generation framework, is proposed for enhancing repository-level code completion by effectively utilizing code analogies across different files within a repository to inform and improve code suggestions.
Furthermore, breaking away from reliance on a singular source of retrieval, \cite{su2024arks} developed a multi-faceted ``knowledge soup'' that integrates web searches, documentation, execution feedback, and evolved code snippets. Then, it incorporates an active retrieval strategy that iteratively refines the query and enriches the knowledge soup, expanding the scope of information available for code generation.

Despite these advancements, several limitations in retrieval-augmented code generation warrant further exploration: 1) the quality of the retrieved information significantly impacts overall performance; 2) the effective integration of retrieved code information with the query needs optimization; 3) an over-reliance on retrieved information may lead to inadequate responses that fail to address the query's intent; 4) additional retrieved information necessitates larger context windows for the LLM, resulting in increased computational demands.

\subsection{Autonomous Coding Agents}\label{sec:autonomous_agents}
The advent of \done{LLMs} has marked the beginning of a new era of
potential pathways toward artificial general intelligence (AGI), capturing significant attention in both academia and industry \cite{xi2023rise,weng2023agent,wang2024survey,huang2024position}. 
A rapidly expanding array of applications for LLM-based autonomous agents, including AutoGPT \cite{autogpt}, AgentGPT \cite{agentgpt}, BabyAGI \cite{babyagi}, and AutoGen \cite{wu2023autogen}, underlines the promise of this technology.

\begin{figure*}[t]
\centering
\includegraphics[width=\linewidth]{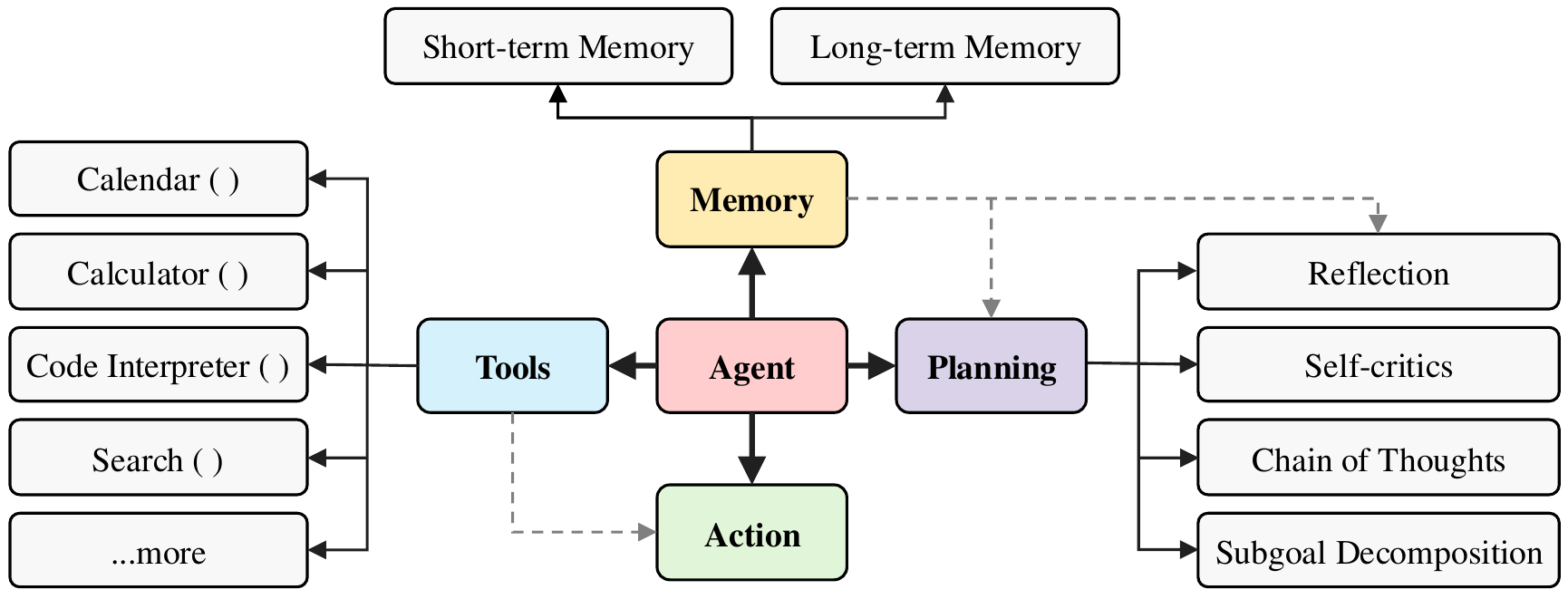}
\caption{The general architecture of an LLM-powered autonomous agent system, adapted from \cite{weng2023agent}. \textbf{Planning}: The agent decomposes large tasks into smaller, manageable sub-goals or engages in self-criticism and self-reflection on past actions to learn from mistakes and improve future performance. \textbf{Memory}: This component enables the agent to store and retrieve past information. \textbf{Tools}: The agent is trained to invoke external functions or APIs. \textbf{Action}: The agent executes actions, with or without the use of tools, to interact with the environment. The gray dashed lines represent the data flow within the system.
}
\label{fig:agent}
\end{figure*}

LLM-powered autonomous agents are systems endowed with sophisticated reasoning abilities, leveraging an LLM as a central computational engine or controller. This allows them to formulate and execute problem-solving plans through a series of tool-enabled functions or API calls. 
Moreover, these agents are designed to function within a shared environment where they can communicate and engage in cooperative, competitive, or negotiating interactions \cite{huang2023agentcoder,wu2023autogen,wang2024survey}. 
The typical architecture of such an agent encompasses an LLM-based Agent, a memory module, a planning component, and a tool utilization module, as depicted in Figure \ref{fig:agent}.

\begin{figure*}[t] \color{blue}
\centering
\includegraphics[width=0.95\linewidth]{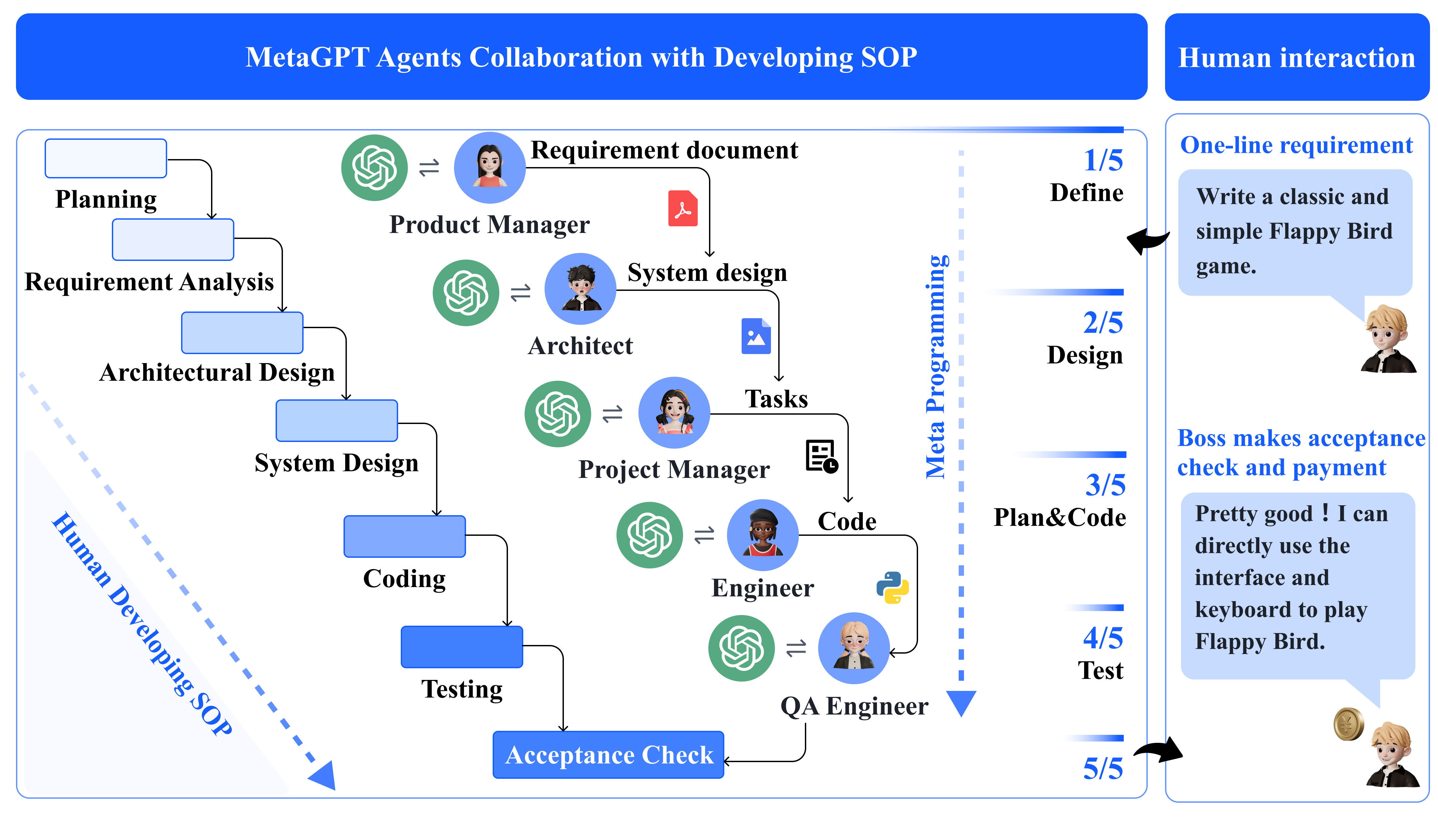}
\caption{\done{MetaGPT integrates human workflow efficiencies into LLM-based multi-agent collaboration to break down complex code-related tasks into specific, actionable procedures. These procedures are then assigned to various roles, such as Product Manager, Architect, and Engineer played by LLM. The image is sourced from the original paper \cite{hong2023metagpt}.
}}
\label{fig:matagpt}
\end{figure*}

In the realm of automated code generation, LLM-powered autonomous agents have demonstrated remarkable proficiency. 
For instance, AgentCoder \cite{huang2023agentcoder} achieved a groundbreaking \texttt{pass@1} of 96.3\% on the HumanEval benchmark, forwarding a step closer to the future of automated software development \cite{ishibashi2024self}. 
\done{The innovative meta-programming framework termed MetaGPT \cite{hong2023metagpt} integrates human workflow efficiencies into LLM-based multi-agent collaboration, as shown in Figure \ref{fig:matagpt}.}  
Furthermore, \cite{huang2023agentcoder} introduces AgentCoder, a multi-agent framework composed of three specialized agents, each with distinct roles and capabilities. These roles include a programmer agent responsible for code generation, a test designer agent tasked with generating unit test cases, and a test executor agent that executes the code and provides feedback. This division of labor within AgentCoder promotes more efficient and effective code generation.
CodeAct \cite{wang2024executable} distinguishes itself by utilizing executable Python code to consolidate LLM agent actions within a unified action space, in contrast to the generation of JSON or textual formats. 
Additionally, AutoCodeRover \cite{zhang2024autocoderover} is proposed to autonomously resolve GitHub issues for program enhancement.

To address the complexity of tasks within software engineering, two innovative autonomous AI software engineers Devin\footnote{\href{https://www.cognition.ai/introducing-devin}{https://www.cognition.ai/introducing-devin}}\cite{Devin} and OpenDevin\footnote{\href{https://github.com/OpenDevin/OpenDevin}{https://github.com/OpenDevin/OpenDevin}}\cite{OpenDevin}, have been released and rapidly garnered considerable interest within the software engineering (SE) and artificial general intelligence (AGI) community. 
Subsequently, an autonomous system, SWE-agent \cite{swe-agent}, leverages a language model to interact with a computer to address software engineering tasks, successfully resolving 12.5\% of issues on the SWE-bench benchmark \cite{jimenez2023swe}. 
L2MAC \cite{holt2023l2mac} has been introduced as the first practical, LLM-based, multi-agent, general-purpose stored-program automatic computer that utilizes a von Neumann architecture, designed specifically for the generation of long and consistent outputs.
At the time of writing this survey, OpenDevin has enhanced CodeAct with bash command-based tools, leading to the release of OpenDevin CodeAct 1.0 \cite{OpenDevin_CodeAct}, which sets a new state-of-the-art performance on the SWE-Bench Lite benchmark \cite{jimenez2023swe}.

Despite these remarkable advancements, the journey toward fully realized AI software engineers employing LLM-powered autonomous agents is far from complete \cite{xi2023rise,wang2024survey}. Critical aspects such as prompt design, context length, agent count, and toolsets call for further refinement and optimization, especially as problem complexities escalate \cite{ishibashi2024self}.

\subsection{Evaluation}\label{sec:evaluation}
\begin{table}[t!] 
\caption{
\done{The performance comparison of LLMs for code generation on the HumanEval \cite{chen2021evaluating} benchmark, measured by \texttt{Pass@1}. 
For models with various sizes, we report only the largest size version of each model with a magnitude of \texttt{B} parameters. $^\ddag$ denotes instruction-tuned models.
} 
}
\label{tab:performance_humaneval}
\centering
\scalebox{0.75}{
\rotatebox{0}{
    \begin{tabular}{clrcc}
    \toprule
    & \textbf{Model} & \textbf{Size} & \texttt{pass@1} (\%) & \textbf{Availability} \\ 
\midrule
    \multirow{17}{*}{\textbf{Closed Source}} 
         & GPT-4o-0513 \cite{gpt-4o} & - & 91.0 & \href{https://openai.com/index/hello-gpt-4o/}{[API Access]} \\
         & GPT-4-Turbo-0409 \cite{gpt-4-turbo} & - & 88.2 &  \href{https://openai.com/blog/new-models-and-developer-products-announced-at-devday}{[API Access]} \\
         & GPT-4-1106 \cite{achiam2023gpt}& - & 87.8  & \href{https://openai.com/index/gpt-4/}{[API Access]} \\
         & GPT-3.5-Turbo-0125 \cite{gpt-3.5-turbo}& - & 76.2  & \href{https://openai.com/index/new-embedding-models-and-api-updates}{[API Access]} \\
         & \cellcolor{yellow!40}Claude-3.5-Sonnet \cite{claude3} & \cellcolor{yellow!40}- & \cellcolor{yellow!40}\textbf{92.0} & \href{https://claude.ai/}{[API Access]} \\
         & Claude-3-Opus \cite{claude3} & - & 84.9 & \href{https://www.anthropic.com/news/claude-3-family}{[API Access]} \\
         & Claude-3-Sonnet \cite{claude3} & - & 73.0 & \href{https://www.anthropic.com/news/claude-3-family}{[API Access]} \\
         & Claude-3-Haiku \cite{claude3} & - & 75.9 & \href{https://www.anthropic.com/news/claude-3-family}{[API Access]} \\
         & Gemini-1.5-Pro \cite{reid2024gemini} & - & 84.1 & \href{https://deepmind.google/technologies/gemini/pro/}{[API Access]} \\
         & Gemini-1.5-Flash \cite{reid2024gemini} & - & 74.3 & \href{https://deepmind.google/technologies/gemini/flash/}{[API Access]} \\
         & Gemini-1.0-Ultra \cite{reid2024gemini} & - & 74.4 & \href{https://deepmind.google/technologies/gemini/ultra/}{[API Access]} \\
         & Gemini-1.0-Pro \cite{reid2024gemini} & - & 67.7 & \href{https://deepmind.google/technologies/gemini/pro/}{[API Access]} \\
         \cline{2-5}
         & $^\ddag$PanGu-Coder2 \cite{shen2023pangu} & 15B & 61.64     & - \\
         & PanGu-Coder \cite{christopoulou2022pangu} & 2.6B & 23.78        & - \\
         & Codex \cite{chen2021evaluating} & 12B & 28.81 & Deprecated \\
         & PaLM-Coder \cite{chowdhery2023palm} & 540B & 36     & - \\
         & AlphaCode \cite{li2022competition} & 1.1B & 17.1     & - \\
         \midrule
    \multirow{36}{*}{\textbf{Open Source}} 
         & $^\ddag$Codestral \cite{codestral} & 22B & 81.1 &  \href{https://huggingface.co/mistralai/Codestral-22B-v0.1}{[Checkpoint Download]} \\
         & \cellcolor{yellow!40}$^\ddag$DeepSeek-Coder-V2-Instruct \cite{zhu2024deepseek}  & \cellcolor{yellow!40}21B (236B) & \cellcolor{yellow!40}\textbf{90.2} & \href{https://huggingface.co/deepseek-ai/DeepSeek-Coder-V2-Instruct}{[Checkpoint Download]}\\
         & \cellcolor{yellow!40}$^\ddag$Qwen2.5-Coder-Instruct \cite{hui2024qwen2} & \cellcolor{yellow!40}7B & \cellcolor{yellow!40}88.4 & \href{https://huggingface.co/Qwen/Qwen2.5-Coder-7B-Instruct}{[Checkpoint Download]}\\
         & Qwen2.5-Coder \cite{hui2024qwen2} & 7B & 61.6  & \href{https://huggingface.co/Qwen/Qwen2.5-Coder-7B}{[Checkpoint Download]}\\
         & $^\ddag$StarCoder2-Instruct \cite{starcoder2instruct} &  15.5B  & 72.6 & \href{https://huggingface.co/bigcode/starcoder2-15b-instruct-v0.1}{[Checkpoint Download]} \\
         & $^\ddag$CodeGemma-Instruct \cite{codegemma_2024}  & 7B  & 56.1  & \href{https://huggingface.co/google/codegemma-7b-it}{[Checkpoint Download]}  \\
         & CodeGemma \cite{codegemma_2024}  & 7B  & 44.5  & \href{https://huggingface.co/google/codegemma-7b}{[Checkpoint Download]}  \\
         & StarCoder 2 \cite{lozhkov2024starcoder}  & 15B & 46.3 & \href{https://huggingface.co/bigcode/starcoder2-15b}{[Checkpoint Download]}\\
         & $^\ddag$WaveCoder-Ultra \cite{yu2023wavecoder} & 6.7B & 79.9  & \href{https://huggingface.co/microsoft/wavecoder-ultra-6.7b}{[Checkpoint Download]}  \\
         & $^\ddag$WaveCoder-Pro \cite{yu2023wavecoder} & 6.7B & 74.4  & \href{https://huggingface.co/microsoft/wavecoder-pro-6.7b}{[Checkpoint Download]}  \\
         & $^\ddag$WaveCoder-DS \cite{yu2023wavecoder} & 6.7B & 65.8  & \href{https://huggingface.co/microsoft/wavecoder-ds-6.7b}{[Checkpoint Download]}  \\
         & StableCode \cite{pinnaparaju2024stable} & 3B & 29.3  & \href{https://huggingface.co/stabilityai/stable-code-3b}{[Checkpoint Download]}  \\
         & CodeShell \cite{xie2024codeshell} & 7B & 34.32  & \href{https://huggingface.co/WisdomShell/CodeShell-7B}{[Checkpoint Download]}  \\
         & $\ddag$CodeQwen1.5-Chat \cite{codeqwen} & 7B & 83.5 & \href{https://huggingface.co/Qwen/CodeQwen1.5-7B-Chat}{[Checkpoint Download]}  \\
         & CodeQwen1.5 \cite{codeqwen} & 7B & 51.8 & \href{https://huggingface.co/Qwen/CodeQwen1.5-7B}{[Checkpoint Download]}  \\
         & $^\ddag$DeepSeek-Coder-Instruct \cite{guo2024deepseek} & 33B & 79.3  & \href{https://huggingface.co/deepseek-ai/deepseek-coder-33b-instruct}{[Checkpoint Download]}  \\
         & DeepSeek-Coder \cite{guo2024deepseek} & 33B & 56.1  & \href{https://huggingface.co/deepseek-ai/deepseek-coder-33b-base}{[Checkpoint Download]}  \\
         & replit-code \cite{replit-code} & 3B & 20.12  & \href{https://huggingface.co/replit/replit-code-v1-3b}{[Checkpoint Download]}  \\
         & $^\ddag$Magicoder$S$-CL \cite{wei2023magicoder} & 7B & 70.7 & \href{https://huggingface.co/ise-uiuc/Magicoder-S-CL-7B}{[Checkpoint Download]} \\
         & $^\ddag$Magicoder-CL \cite{wei2023magicoder} & 7B & 60.4 & \href{https://huggingface.co/ise-uiuc/Magicoder-CL-7B}{[Checkpoint Download]}\\
         & $^\ddag$WizardCoder \cite{luo2023wizardcoder} & 33B & 79.9        & \href{https://huggingface.co/WizardLM/WizardCoder-33B-V1.1}{[Checkpoint Download]}  \\
         & CodeFuse \cite{liu2023mftcoder} & 34B & 74.4     & \href{https://huggingface.co/TheBloke/CodeFuse-CodeLlama-34B-GGUF}{[Checkpoint Download]}  \\
         & Phi-1 \cite{gunasekar2023textbooks} & 1.3B & 50.6          & \href{https://huggingface.co/microsoft/phi-1}{[Checkpoint Download]}  \\
         & $^\ddag$Code Llama-Instruct \cite{roziere2023code} & 70B & 67.8 & \href{https://huggingface.co/codellama/CodeLlama-70b-Instruct-hf}{[Checkpoint Download]}  \\
         & Code Llama \cite{roziere2023code} & 70B & 53.0  & \href{https://huggingface.co/codellama/CodeLlama-70b-hf}{[Checkpoint Download]}  \\
         & $^\ddag$OctoCoder \cite{muennighoff2023octopack} & 15.5B & 46.2   & \href{https://huggingface.co/bigcode/octocoder}{[Checkpoint Download]}  \\
         & CodeGeeX2 \cite{zheng2023codegeex} & 6B & 35.9          & \href{https://huggingface.co/THUDM/codegeex2-6b}{[Checkpoint Download]}  \\
         & $^\ddag$InstructCodeT5+ \cite{wang2023codet5+} & 16B & 35.0          & \href{https://huggingface.co/Salesforce/instructcodet5p-16b}{[Checkpoint Download]}  \\
        & CodeGen-NL \cite{nijkamp2022codegen} & 16.1B & 14.24             & \href{https://huggingface.co/Salesforce/codegen-16B-nl}{[Checkpoint Download]}  \\
        & CodeGen-Multi \cite{nijkamp2022codegen} & 16.1B & 18.32        & \href{https://huggingface.co/Salesforce/codegen-16B-multi}{[Checkpoint Download]}  \\
        & CodeGen-Mono \cite{nijkamp2022codegen} & 16.1B & 29.28           & \href{https://huggingface.co/Salesforce/codegen-16B-mono}{[Checkpoint Download]}  \\
        & StarCoder \cite{li2023starcoder} & 15B & 33.60     & \href{https://huggingface.co/bigcode/starcoder}{[Checkpoint Download]}  \\
        & CodeT5+ \cite{wang2021codet5} & 16B & 30.9    & \href{https://huggingface.co/Salesforce/codet5p-16b}{[Checkpoint Download]}  \\
        & CodeGen2 \cite{nijkamp2023codegen2} & 16B & 20.46       & \href{https://huggingface.co/Salesforce/codegen2-16B_P}{[Checkpoint Download]}  \\
         & SantaCoder \cite{allal2023santacoder} & 1.1B & 14.0        & \href{https://huggingface.co/bigcode/santacoder}{[Checkpoint Download]}  \\
         & InCoder \cite{fried2022incoder} & 6.7B & 15.2     & \href{https://huggingface.co/facebook/incoder-6B}{[Checkpoint Download]}  \\
         & PolyCoder \cite{xu2022systematic} & 2.7B & 5.59   & \href{https://huggingface.co/NinedayWang/PolyCoder-2.7B}{[Checkpoint Download]}  \\
         & CodeParrot \cite{tunstall2022natural} & 1.5B & 3.99   & \href{https://huggingface.co/codeparrot/codeparrot}{[Checkpoint Download]} \\  
    \bottomrule
    \end{tabular}
}
}
\vspace{-10pt}
\end{table}
Despite the impressive capabilities of \done{LLMs}, they exhibit a range of behaviors that are both beneficial and potentially risky. 
These behaviors can enhance performance across various downstream tasks but may also introduce reliability and trustworthiness concerns in LLM deployment \cite{chen2021evaluating, xu2022systematic, chang2024survey}. 
Consequently, it is imperative to develop precise evaluation approaches to discern the qualitative and quantitive differences between models, thereby encouraging further advancements in LLM capabilities.

Evaluation strategies for LLMs in code generation mirror those for general-purpose LLMs and can be divided into three principal categories: metrics-based, human-centered, and LLM-based approaches. 
Detailed benchmarks for these evaluation strategies are presented in Section \ref{sec:benchmark} and summarized in Table \ref{tab:benchmark}. 
Subsequent subsections will provide a thorough analysis of each approach.

\subsubsection{Metrics}
The pursuit of effective and reliable automatic evaluation metrics for generated content is a long-standing challenge within the field of natural language processing (NLP) \cite{chen1998evaluation, papineni2002bleu, lin2004rouge}. 
At the early stage, most works directly leverage token-matching-based metrics, such as Exact Match, BLEU \cite{papineni2002bleu}, ROUGE \cite{lin2004rouge}, and METEOR \cite{banerjee2005meteor}, which are prevalent in text generation of NLP, to assess the quality of code generation.

While these metrics offer a rapid and cost-effective approach for assessing the quality of generated code, they often fall short of capturing the syntactical and functional correctness, as well as the semantic features of the code. 
To eliminate this limitation, CodeBLEU \cite{ren2020codebleu} was introduced, enhancing the traditional BLEU metric \cite{papineni2002bleu} by incorporating syntactic information through abstract syntax trees (AST) and semantic understanding via data-flow graph (DFG). 
Despite these improvements, the metric does not fully resolve issues pertaining to execution errors or discrepancies in the execution results of the generated code.
In light of these challenges, execution-based metrics have gained prominence for evaluating code generation, including \texttt{pass@k} \cite{chen2021evaluating}, \texttt{n@k} \cite{li2022competition}, test case average \cite{hendrycks2021measuring}, execution accuracy \cite{rajkumar2022evaluating}, and \texttt{pass@t} \cite{olausson2023self}. 
In particular, the \texttt{pass@k}, serving as a principal evaluation metric, assesses the probability that at least one out of $k$ code samples generated by a model will pass all unit tests. An unbiased estimator for \texttt{pass@k} introduced by \cite{chen2021evaluating} is defined as: 
\begin{equation}\label{eq:pass@k}
\begin{aligned}
    \texttt{pass@k} \coloneqq \mathbb{E}_\text{task} \left[1-\frac{\binom{n-c}{k}}{\binom{n}{k}}\right]
\end{aligned} 
\end{equation}
where $n$ is the total number of sampled candidate code solutions, $k$ is the number of randomly selected code solutions from these candidates for each programming problem, with $n\ge k$, and $c$ is the count of correct samples within the $k$ selected. 

Nevertheless, these execution-based methods are heavily dependent on the quality of unit tests and are limited to evaluating executable code \cite{zan2023large}. Consequently, when unit tests are unavailable, token-matching-based metrics are often employed as an alternative for evaluation. 
Furthermore, in scenarios lacking a ground truth label, unsupervised metrics such as perplexity (PPL) \cite{jelinek1977perplexity} can serve as evaluative tools. Perplexity quantifies an LLM's uncertainty in predicting new content, thus providing an indirect measure of the model's generalization capabilities and the quality of the generated code.

Taken together, while the aforementioned methods primarily focus on the functional correctness of code, they do not provide a holistic evaluation that encompasses other critical dimensions such as code vulnerability \cite{nappa2015attack}, maintainability \cite{ardito2020tool}, readability \cite{buse2009learning}, complexity and efficiency \cite{peitek2021program}, stylistic consistency \cite{markovtsev2019style}, and execution stability \cite{raemaekers2012measuring}. 
A comprehensive evaluation framework that integrates these aspects remains an open area for future research and development in the field of code generation assessment.

\subsubsection{Human Evaluation}
Given the intrinsic characteristics of code, the aforementioned automatic evaluation metrics are inherently limited in their capacity to fully assess code quality. 
For instance, metrics specifically designed to measure code style consistency are challenging to develop and often fail to capture this aspect adequately \cite{chen2023duetcs}. 
When it comes to repository-level code generation, the evaluation of overall code quality is substantially complicated due to the larger scale of the task, which involves cross-file designs and intricate internal as well as external dependencies, as discussed by \cite{bairi2023codeplan,shrivastava2023repofusion}.

To overcome these challenges, conducting human evaluations becomes necessary, as it yields relatively robust and reliable results.
Human assessments also offer greater adaptability across various tasks, enabling the simplification of complex and multi-step evaluations. 
Moreover, human evaluations are essential for demonstrating the effectiveness of certain token-matching-based metrics, such as CodeBLEU \cite{ren2020codebleu}. 
These studies typically conduct experiments to evaluate the correlation coefficient between proposed metrics and quality scores assigned by actual users, demonstrating their superiority over existing metrics.

Moreover, in an effort to better align \done{LLMs} with human preferences and intentions, InstructGPT \cite{ouyang2022training} employs human-written prompts and demonstrations, and model output ranking in the fine-tuning of LLMs using reinforcement learning from human feedback (RLHF). 
Although similar alignment learning techniques have been applied to code generation, the feedback in this domain typically comes from a compiler or interpreter, which offers execution feedback, rather than from human evaluators. 
Notable examples include CodeRL \cite{le2022coderl}, PPOCoder \cite{shojaee2023execution}, RLTF \cite{liu2023rltf}, and PanGu-Coder2 \cite{shen2023pangu}. 
Further information on this topic is available in Section \ref{sec:reinforcement_learning}.

Nonetheless, human evaluations are not without drawbacks, as they can be prone to certain issues that may compromise their accuracy and consistency. 
For instance, 
1) personalized tastes and varying levels of expertise among human evaluators can introduce biases and inconsistencies into the evaluation process;
2) conducting comprehensive and reliable human evaluations often necessitates a substantial number of evaluators, leading to significant expenses and time-consuming;
3) the reproducibility of human evaluations is often limited, which presents challenges in extending previous evaluation outcomes or monitoring the progress of LLMs, as highlighted by \cite{zhao2023survey}.

\subsubsection{LLM-as-a-Judge}
\begin{figure*}[t]
\centering
\includegraphics[width=0.95\linewidth]{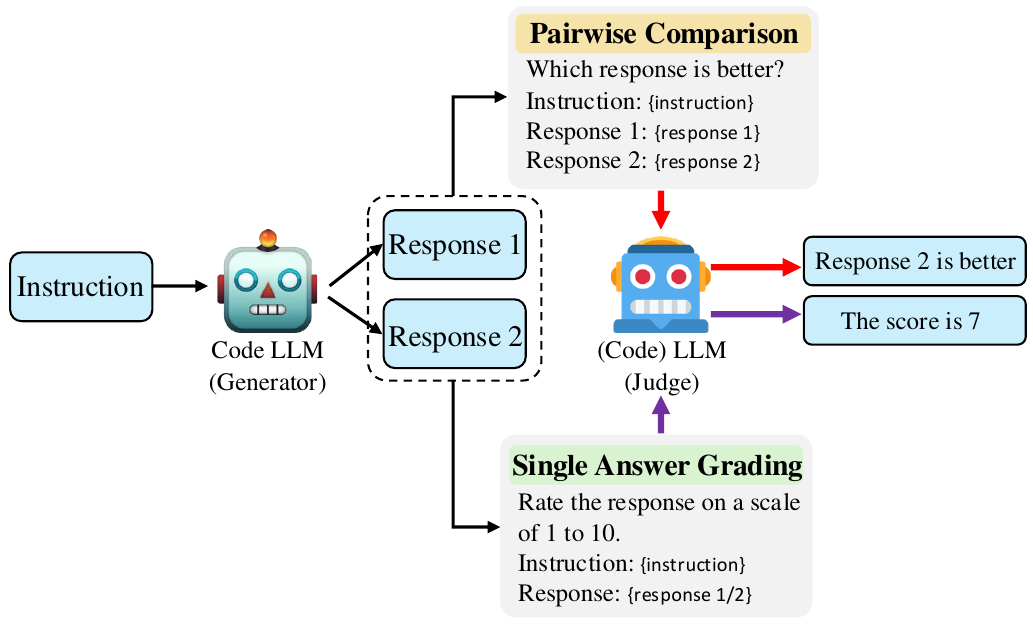}
\caption{\done{The pipeline of (Code) LLM-as-a-judge for evaluating generated code by Code LLMs. There are primarily two types of approaches: pairwise comparison and single answer grading.} 
}
\label{fig:llm-as-judge}
\end{figure*}
The powerful instruction-following capabilities of \done{LLMs} have stimulated researchers to innovatively investigate the potential of LLM-based evaluations. 
The LLM-as-a-Judge \cite{zheng2024judging} refers to the application of advanced proprietary LLMs (e.g., GPT4, Gemini, and Claud 3) as proxies for human evaluators. This involves designing prompts with specific requirements to guide LLMs in conducting evaluations, as demonstrated by AlpacaEval \cite{alpaca_eval} and MT-bench \cite{zheng2024judging}. 
This method reduces reliance on human participation, thereby facilitating more efficient and scalable evaluations. 
Moreover, LLMs can offer insightful explanations for the assigned rating scores, thereby augmenting the interpretability of evaluations \cite{zhao2023survey}.

Nevertheless, the use of LLM-based evaluation for code generation remains relatively underexplored compared with general-purpose LLM. 
\done{The pipeline of (Code) LLM-as-a-judge for evaluating generated code by Code LLMs is depicted in Figure \ref{fig:llm-as-judge}.}
A recent work \cite{zhuo2024ice} introduces the ICE-Score evaluation metric, which instructs LLM for code assessments. 
This approach attains superior correlations with functional correctness and human preferences, thereby eliminating the requirement for test oracles or references.
As the capabilities of LLM continue to improve, we anticipate seeing more research in this direction.

Despite their scalability and explainability, the effectiveness of LLM-based evaluation is constrained by the inherent limitations of the chosen LLM. 
Several studies have shown that most LLMs, including GPT-4, suffer from several issues, including position, verbosity, and self-enhancement biases, as well as restricted reasoning ability \cite{zheng2024judging}.
Specifically, position bias refers to the tendency of \done{LLMs} to disproportionately favor responses that are presented in certain positions, which can skew the perceived quality of answers based on their order of presentation. 
Meanwhile, verbosity bias describes the inclination of LLMs to prefer lengthier responses, even when these are not necessarily of higher quality compared to more concise ones. 
Self-enhancement bias, on the other hand, is observed when LLMs consistently overvalue the quality of the text they generate \cite{zheng2024judging,zhao2023survey}.
Moreover, due to their inherent limitations in tackling complex reasoning challenges, LLMs may not be entirely reliable as evaluators for tasks that require intensive reasoning, such as those involving mathematical problem-solving.
However, these shortcomings can be partially addressed through the application of deliberate prompt engineering and fine-tuning techniques, as suggested by \cite{zheng2024judging}.

\done{\subsubsection{Empirical Comparison}

\begin{figure*}[t]
\centering
\includegraphics[width=\linewidth]{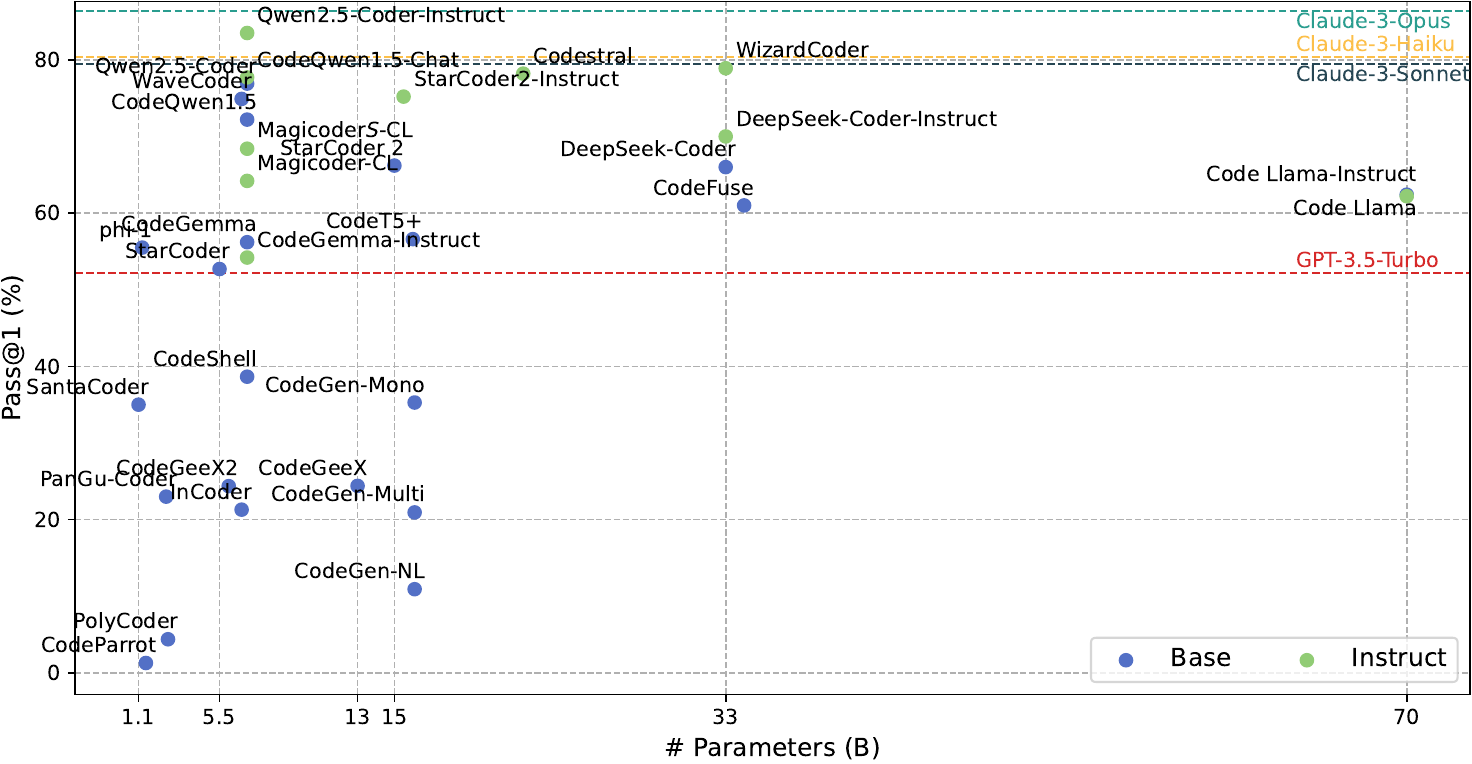}
\caption{\done{The performance comparison of LLMs for code generation on the MBPP \cite{austin2021program} benchmark, measured by \texttt{Pass@1}. For models with various sizes, we report only the largest size version of each model with a magnitude of \texttt{B} parameters.} 
}
\label{fig:mbpp_performance}
\end{figure*}

\begin{figure*}[t]
\centering
\includegraphics[width=\linewidth]{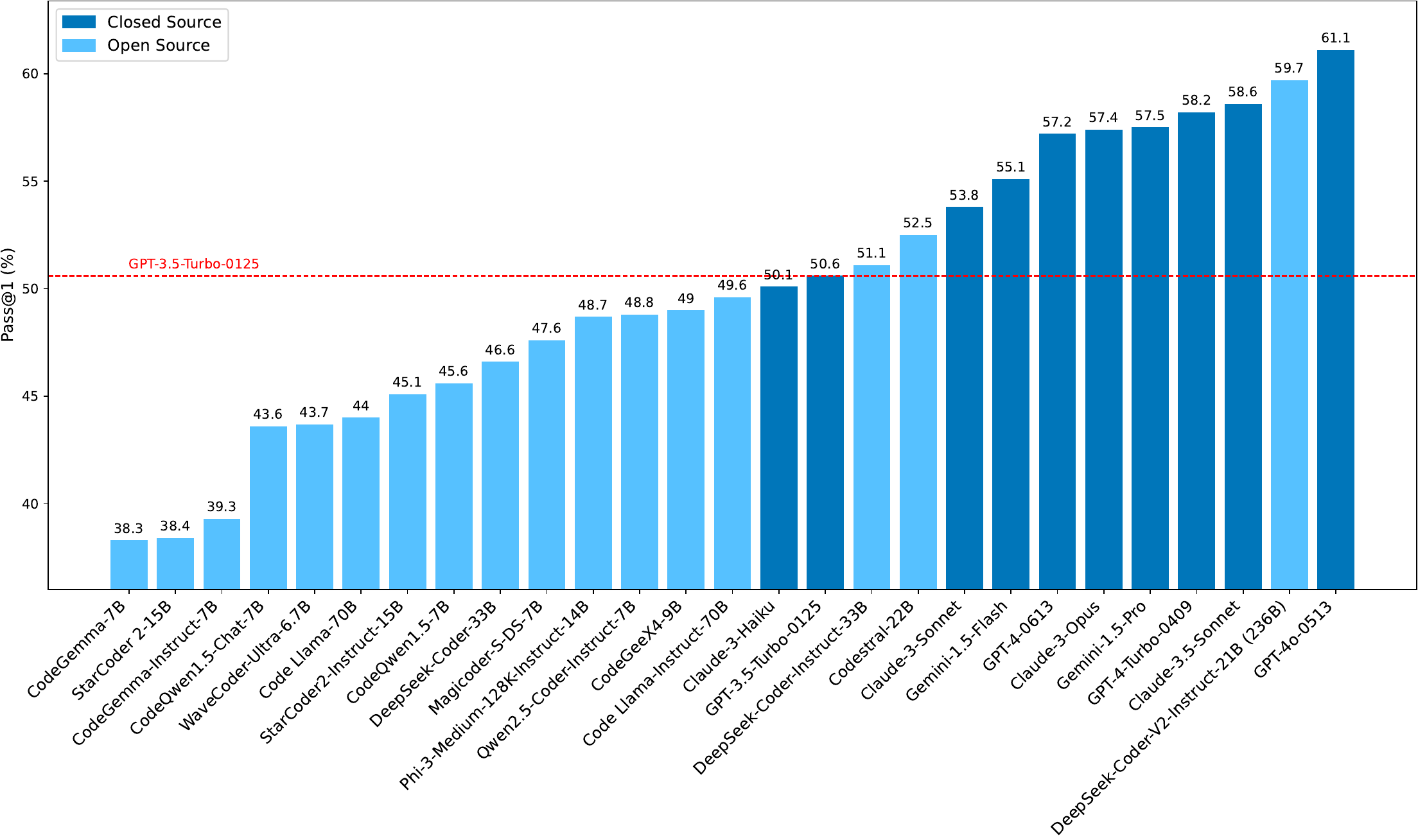}
\caption{\done{The performance comparison of LLMs for code generation on the BigCodeBench \cite{zhuo2024bigcodebench} benchmark, measured by \texttt{Pass@1}. For models with various sizes, we report only the largest size version of each model with a magnitude of \texttt{B} parameters.} 
}
\label{fig:bigcodebench_performance}
\end{figure*}
In this section, we present a performance comparison of LLMs for code generation using the well-regarded HumanEval, MBPP, and the more practical and challenging BigCodeBench benchmarks. 
This empirical comparison aims to highlight the progressive enhancements in LLM capabilities for code generation. 
These benchmarks assess an LLM's ability to generate source code across various levels of difficulty and types of programming tasks. 
Specifically, HumanEval focuses on complex code generation, MBPP targets basic programming tasks, and BigCodeBench emphasizes practical and challenging programming tasks.

Due to the limitations in computational resources we faced, we have cited experimental results from original papers or widely recognized open-source leaderboards within the research community, such as the HumanEval Leaderboard \footnote{\href{https://paperswithcode.com/sota/code-generation-on-humaneval}{https://paperswithcode.com/sota/code-generation-on-humaneval}}, EvalPlus Leaderboard \footnote{\href{https://evalplus.github.io/leaderboard.html}{https://evalplus.github.io/leaderboard.html}}, Big Code Models Leaderboard  \footnote{\href{https://huggingface.co/spaces/bigcode/bigcode-models-leaderboard}{https://huggingface.co/spaces/bigcode/bigcode-models-leaderboard}}, and BigCodeBench Leaderboard \footnote{\href{https://bigcode-bench.github.io/}{https://bigcode-bench.github.io/}}. 
We report performance on HumanEval using the \texttt{pass@1} metric, as shown in Table \ref{tab:performance_humaneval}, while MBPP and BigCodeBench results are presented with \texttt{pass@1} in Figures \ref{fig:mbpp_performance} and \ref{fig:bigcodebench_performance}, respectively.

We offer the following insights:
\begin{itemize}
    \item The performance gap between open-source and closed-source models across the three benchmarks is gradually narrowing. For instance, on the HumanEval benchmark, DeepSeek-Coder-V2-Instruct with 21B activation parameters and Qwen2.5-Coder-Instruct 7B achieve 90\% and 88.4\% pass@1, respectively. 
    These results are comparable to the much larger closed-source LLMs, such as Claude-3.5-Sonnet, which achieves 92.0\% pass@1. 
    On the MBPP benchmark, Qwen2.5-Coder-Instruct 7B with 83.5\% pass@1 significantly outperforms GPT-3.5-Turbo with 52.2\% pass@1 and closely rivals the closed-source Claude-3-Opus with 86.4\% pass@1. 
    On the BigCodeBench, DeepSeek-Coder-V2-Instruct achieves 59.7\%, surpassing all compared closed-source and open-source LLMs except for slightly falling behind GPT-4o-0513, which achieves 61.1\%.
    \item Generally, as the number of model parameters increases, the performance of code LLMs improves. However, Qwen2.5-Coder-Instruct 7B achieves 88.4\% pass@1, outperforming larger models like StarCoder2-Instruct 15.5B with 72.6\% pass@1, DeepSeek-Coder-Instruct 33B with 79.3\% pass@1, and Code Llama-Instruct 70B with 67.8\% pass@1 on the HumanEval benchmark. Similar trends are observed across the other two benchmarks, suggesting that code LLMs with 7B parameters may be sufficiently capable for code generation task.
    \item Instruction-tuned models consistently outperform their base (pretrained) counterparts across the HumanEval and MBPP benchmarks. For instance, Qwen2.5-Coder-Instruct surpasses Qwen2.5-Coder by an average of 26.04\%, StarCoder2-Instruct improves upon StarCoder 2 by an average of 35.20\%, and CodeGemma-Instruct enhances CodeGemma by an average of 11.26\%. Additionally, DeepSeek-Coder-Instruct outperforms DeepSeek-Coder by an average of 23.71\%, while Code Llama-Instruct shows a 13.80\% improvement over Code Llama. Detailed results can be found in Table \ref{tab:instruct_improve}. These findings underscore the effectiveness of instruction tuning, although the quality of the instruction tuning dataset plays a critical role in determining model performance \cite{luo2023wizardcoder,zhou2024lima}.
    \begin{table}[t] 
\caption{\done{The performance improvement of instruction-tuned models over their pretrained counterparts on the HumanEval, MBPP, and BigCodeBench benchmarks. 
The last two rows demonstrate the average improvement on the first two benchmarks and the three benchmarks, respectively.
}
}
\label{tab:instruct_improve}
\centering
\scalebox{0.8}{
\rotatebox{0}{
    \begin{tabular}{lccccc}
    \toprule
    & \textbf{\makecell[c]{Qwen2.5-Coder-\\Instruct 7B}} & \textbf{\makecell[c]{StarCoder2-\\Instruct 15.5B}} & \textbf{\makecell[c]{CodeGemma-\\Instruct 7B}} & \textbf{\makecell[c]{DeepSeek-Coder-\\Instruct 33B}} & \textbf{\makecell[c]{Code Llama-\\Instruct 70B}} \\ 
\midrule
    \textbf{HumanEval} & 43.51\% & 56.80\% & 26.07\% & 41.35\% & 27.92\% \\
    \textbf{MBPP} & 8.58\% & 13.60\% & \cellcolor{gray!15}-3.56\% & 6.06\% & \cellcolor{gray!15}-0.32\% \\
    \textbf{BigCodeBench} & - & 17.45\% & 2.61\% & 9.66\% & 12.73\% \\
    \midrule
    \textbf{\# Avg. Imp. H. M.} & 26.04\% & 35.20\% & 11.26\%  & 23.71\% & 13.80\% \\
    \textbf{\# Avg. Imp. H. M. B.} & - & 29.28\% & 8.37\% & 19.02\% & 13.44\% \\
    \bottomrule
    \end{tabular}
}
}
\end{table}
    \item Performance on the HumanEval benchmark is nearly saturated. However, MBPP, which involves basic programming tasks, and BigCodeBench, which involves more practical and challenging programming tasks, demand more capable code LLMs. Additionally, while these benchmarks primarily evaluate the functional correctness of code, they do not provide a comprehensive assessment across other critical dimensions. Developing a more holistic evaluation framework that integrates various aspects remains an open area for future research and development in LLMs for code generation evaluation.
\end{itemize}

\textbf{Discussion}:
We discuss certain code LLMs in Table \ref{tab:performance_humaneval} for clarity:
(1) General LLMs accessed via API are not specifically trained on large code corpora but achieve state-of-the-art performance in code generation, such as Claude-3.5-Sonnet with 92.0\% pass@1 on HumanEval benchmark.
(2) AlphaCode targets code generation for more complex and unseen problems that require a deep understanding of algorithms and intricate natural language, such as those encountered in competitive programming.
The authors of AlphaCode found that large-scale model sampling to navigate the search space, such as 1M samples per problem for CodeContests, followed by filtering based on program behavior to produce a smaller set of submissions, is crucial for achieving good and reliable performance on problems that necessitate advanced reasoning.
(3) Phi-1 1.3B is a specialized LLM for code, trained on ``textbook quality'' data from the web (6B tokens) and synthetically generated textbooks and exercises using GPT-3.5 (1B tokens).
(4) Code Llama 70B is initialized with Llama 2 model weights and continually pre-trained on 1T tokens from a code-heavy dataset and long-context fine-tuned with approximately 20B tokens. However, Code Llama-Instruct 70B is fine-tuned from Code Llama-Python 70B without long-context fine-tuning, using an additional 260M tokens to better follow human instructions. 
Surprisingly, these models underperform compared to smaller parameter Code LLMs like Qwen2.5-Coder-Instruct 7B, DeepSeek-Coder-V2-Instruct 21B, and Codestral 22B across all three benchmarks. The underlying reasons for this discrepancy remain unclear and warrant further exploration.
(5) Unlike other open-source Code LLMs, DeepSeek-Coder-V2-Instruct is further pre-trained on DeepSeek-V2 \cite{liu2024deepseek}, which employs a Mixture-of-Experts (MoE) architecture with only 21B activation parameters out of 236B parameters, using an additional 6 trillion tokens composed of 60\% source code, 10\% math corpus, and 30\% natural language corpus. For a comprehensive understanding of MoE in LLMs, please refer to \cite{cai2024survey}.

}\label{sec:empirical_comparison}
\done{\subsection{Code LLMs Alignment}\label{sec:responsible_codeai}
The pre-training of LLMs for next-token prediction, aimed at maximizing conditional generation likelihood across vast textual corpora, equips these models with extensive world knowledge and emergent capabilities \cite{brown2020language}. 
This training approach enables the generation of coherent and fluent text in response to diverse instructions. 
Nonetheless, LLMs can sometimes misinterpret human instructions, produce biased content, or generate factually incorrect information (commonly referred to as hallucinations), which may limit their practical utility \cite{wang2023aligning,zhao2023survey,ji2023ai}.

Aligning LLMs with human intentions and values, known as LLM alignment, has consequently become a critical research focus \cite{ji2023ai,wang2023aligning}. 
Key objectives frequently discussed in the context of LLM alignment include robustness, interpretability, controllability, ethicality, trustworthiness, security, privacy, fairness, and safety. 
In recent years, significant efforts have been made by researchers to achieve this alignment, employing techniques such as Reinforcement Learning with Human Feedback (RLHF) \cite{ouyang2022training}.

However, the alignment of Code LLMs has not been extensively explored. 
Compared to text generation, aligning code generation with human intentions and values is even more crucial. For instance, users without programming expertise might prompt Code LLM to generate source code and subsequently execute it on their computers, potentially causing catastrophic damage. Some potential risks include:
\begin{itemize}
    \item \textbf{Malware Infection}: The code could contain viruses, worms, or trojans that compromise our system's security.
    \item \textbf{Data Loss}: It might delete or corrupt important files and data.
    \item \textbf{Unauthorized Access}: It can create backdoors, allowing attackers to access our system remotely.
    \item \textbf{Performance Issues}: The code might consume excessive resources, slowing down our system.
    \item \textbf{Privacy Breaches}: Sensitive information, such as passwords or personal data, might be stolen.
    \item \textbf{System Damage}: It may alter system settings or damage hardware components.
    \item \textbf{Network Spread}: It could propagate across networks, affecting other devices.
    \item \textbf{Financial Loss}: If the code is ransomware, it might encrypt data and demand payment for decryption.
    \item \textbf{Legal Consequences}: Running certain types of malicious code can lead to legal repercussions.
\end{itemize} 
As illustrated, aligning Code LLMs to produce source code consistent with human preferences and values is of paramount importance in software development. A recent study \cite{yang2024robustness} provides the first systematic literature review identifying seven critical non-functional properties of LLMs for code, beyond accuracy, including robustness, security, privacy, explainability, efficiency, and usability. This study is highly pertinent to the alignment of Code LLMs. We recommend readers refer to this survey for more detailed insights.


In this survey, we identify five core principles that serve as the key objectives for aligning Code LLMs: Green, Responsibility, Efficiency, Safety, and Trustworthiness (collectively referred to as \textbf{GREST}). These principles are examined from a broader perspective. Each category encompasses various concepts and properties, which are summarized in Table \ref{tab:codellm_alignment}. 
\begin{table}[t] 
\centering
\caption{\done{
Five core principles serve as the key objectives for Code LLMs alignment: Green, Responsibility, Efficiency, Safety, and Trustworthiness (collectively referred to as \textbf{GREST}).
}}
\label{tab:codellm_alignment}
\scalebox{0.65}{
\begin{tabular}{ll}
\toprule
\textbf{Principles} & \textbf{Involved Concepts and Properties} \\ 
\midrule
\textbf{Green} & 
\makecell[l]{
     \textbf{Energy Efficiency}: Minimizing computational energy use and reduce environmental impact and financial costs.\\
     \textbf{Sustainable Materials}: Leveraging eco-friendly infrastructure and servers for code generation, lowering long-term expenses. \\
     \textbf{Carbon Footprint}: Reducing emissions associated with model training and inference to enhance efficiency and save costs. \\
     \textbf{Resource Optimization}: Efficiently utilizing computational resources to minimize waste and reduce expenses in code generation. \\
     \textbf{Recycling Management}: Responsibly dispose of hardware used in model development to reduce waste management costs. \\
     \textbf{Renewable Energy}: Utilizing renewable energy sources for powering training and inference processes to decrease energy costs.  \\
     \textbf{Lifecycle Assessment}: Evaluating the environmental and financial impacts of models from creation to deployment and disposal.
} \\
\midrule
\textbf{Responsibility} & 
\makecell[l]{
     \textbf{Ethical Considerations}: Adhering to ethical guidelines to ensure responsible use and deployment of generated code. \\
     \textbf{Accountability}: Establishing clear lines of responsibility for code generation outcomes and potential impacts. \\
     \textbf{User Education}: Providing resources and guidance to help users understand and responsibly use generated code. \\
     \textbf{Impact Assessment}: Evaluating the social and technical implications of code generation to minimize negative effects. \\
     \textbf{Regulatory Compliance}: Ensuring that generated code adheres to relevant laws (e.g., copyright) and industry regulations.
} \\
\midrule
\textbf{Efficiency} & 
\makecell[l]{
     \textbf{Model Optimization}: Streamlining models to reduce computational load and improve speed. \\
     \textbf{Prompt Engineering}: Designing effective prompts to generate accurate code efficiently. \\
     \textbf{Resource Management}: Allocating computational resources wisely to balance speed and cost. \\
     \textbf{Inference Optimization}: Enhancing the inference process to quickly generate code with minimal latency. \\
     \textbf{Parallel Processing}: Utilizing parallelism to speed up code generation tasks. \\
     \textbf{Caching Mechanisms}: Implementing caching to reuse previous results and reduce redundant computations. \\
     \textbf{Evaluation Metrics}: Using precise metrics to assess and improve the efficiency of code outputs.
} \\
\midrule
\textbf{Safety} & 
\makecell[l]{
     \textbf{Input Validation}: Ensuring inputs (prompts) are safe and sanitized to prevent malicious exploitation. \\
     \textbf{Security Audits}: Regularly reviewing generated code for vulnerabilities and potential exploits. \\
     \textbf{Monitoring and Logging}: Keeping track of generation outputs to quickly identify and address safety issues. \\
     \textbf{User Access Control}: Limiting access to generation capabilities to trusted users to minimize risk. \\
     \textbf{Continuous Updates}: Regularly updating models with the latest safety protocols and security patches. \\
     \textbf{Ethical Guidelines}: Implementing ethical standards to guide safe and responsible code generation.
} \\
\midrule
\textbf{Trustworthiness} & 
\makecell[l]{
     \textbf{Reliability}: Ensuring that generated code consistently meets functional requirements and performs as expected. \\
     \textbf{Transparency}: Providing clear explanations of how code is generated to build user confidence. \\
     \textbf{Verification and Testing}: Using rigorous testing frameworks to ensure the generated code accuracy and reliability. \\
     \textbf{Bias Mitigation}: Actively working to identify and reduce biases in code generation to ensure fairness and impartiality. \\
     \textbf{User Feedback Integration}: Continuously incorporating user feedback to refine and improve code generation processes. \\
     \textbf{Documentation}: Providing comprehensive documentation for generated code to enhance understanding and trust.
    } \\
\bottomrule
\end{tabular}
}
\end{table}
In the following, we define each principle and briefly introduce a few notable works to enhance understanding.

\textbf{Green}: 
The Green principle underscores the importance of environmental sustainability in the development and deployment of LLMs for code generation. This involves optimizing energy consumption and reducing both the carbon footprint and financial costs associated with training and inference processes.
Currently, training, inference, and deployment of Code LLMs are notably resource-intensive. For example, training GPT-3, with its 175 billion parameters, required the equivalent of 355 years of single-processor computing time and consumed 284,000 kWh of energy, resulting in an estimated 552.1 tons of CO$_2$ emissions \cite{samsi2023words}. 
Furthermore, a ChatGPT-like application, with an estimated usage of 11 million requests per hour, can produce emissions of 12.8k metric tons of CO$_2$ per year, which is 25 times the carbon emissions associated with training GPT-3 \cite{chien2023reducing}.
To mitigate these costs, several techniques are often employed, such as the development of specialized hardware (e.g., Tensor Processing Units (TPUs) and Neural Processing Units (NPUs)), model compression methods (e.g., quantization and knowledge distillation), parameter-efficient fine-tuning (PEFT), and the use of renewable energy sources.
For instance, Shi et al. \cite{shi2024greening} applied knowledge distillation to reduce the size of CodeBERT \cite{feng2020codebert} and GraphCodeBERT \cite{guo2020graphcodebert}, resulting in optimized models of just 3MB. These models are 160 times smaller than the original large models and significantly reduce energy consumption by up to 184 times and carbon footprint by up to 157 times. Similarly, Wei et al. \cite{wei2023towards} utilized quantization techniques for Code LLMs such as CodeGen \cite{nijkamp2022codegen} and Incoder \cite{fried2022incoder} by employing lower-bit integers (e.g., \texttt{int8}). This approach reduced storage requirements by 67.3\% to 70.8\%, carbon footprint by 28.8\% to 55.0\%, and pricing costs by 28.9\% to 55.0\%.

\textbf{Responsibility}: 
The Responsibility principle in the context of Code LLMs underscores the importance of ethical considerations, fairness, and accountability throughout their lifecycle. This involves addressing biases in training data, ensuring fairness and transparency in model decision-making, maintaining accountability for outputs, adhering to applicable laws (e.g., copyright), implementing safeguards against misuse, and providing clear communication about the model's capabilities and limitations.
Specifically, 
\begin{itemize}
    \item \textit{Bias Mitigation}. Biases in code generation can lead to flawed software and reinforce stereotypes, potentially causing significant societal impacts. For example, an Code LLM that inherits biases from its training data may produce source code/software that inadvertently discriminates against certain user groups. This can result in applications that fail to meet the diverse needs of users, promoting exclusionary practices and reinforcing existing stereotypes \cite{mouselinos2022simple,liu2023uncovering}.
    \item \textit{Fairness and Transparency}. A lack of fairness and transparency in Code LLM decision-making can result in biased or suboptimal code solutions. If the model's decision-making process is opaque, developers might unknowingly introduce code that favors specific frameworks or libraries, thereby limiting innovation and diversity in software development. This opacity can create unfair advantages and hinder collaborative efforts within tech communities \cite{bogina2022educating}.
    \item \textit{Legal Compliance}. Compliance with relevant laws, such as licensing and copyright, is crucial when using Code LLMs for code generation to avoid legal complications. If an Code LLM generates code snippets that inadvertently infringe on existing copyrights, it can lead to legal disputes and financial liabilities for developers and organizations \cite{xu2024first}. Such risks may discourage the use of advanced AI tools, thus stifling innovation and affecting growth and collaboration within the tech community.
    \item \textit{Accountability}. Without accountability for code generated by Code LLMs, addressing bugs or security vulnerabilities becomes challenging. If a model generates faulty code leading to a security breach, the absence of clear accountability can result in significant financial and reputational damage for companies. This uncertainty can delay critical issue resolution and impede trust in AI-assisted development \cite{liesenfeld2023opening}.
    \item \textit{Misuse Prevention}. Failing to implement mechanisms to prevent the misuse of Code LLMs can enable the creation of harmful software. For example, models could be exploited to generate malware or unauthorized scripts, posing cybersecurity risks. Without proper safeguards, these models can facilitate malicious activities, threatening both individual and organizational security \cite{mousavi2024investigation}.
    \item \textit{Clear Communication}. Without clear communication about a model's capabilities and limitations, developers may misuse the model or overestimate its abilities. Relying on the model to generate complex, mission-critical code without human oversight can lead to significant software failures. Misunderstanding its limitations can result in faulty implementations and lost productivity \cite{ross2023programmer}.
\end{itemize}
To adhere to this principle, potential mitigation methods include bias detection and mitigation, quantification and evaluation, and adherence to ethical guidelines. 
Liu et al. \cite{liu2023uncovering} propose a new paradigm for constructing code prompts, successfully uncovering social biases in code generation models, and developing a dataset along with three metrics to evaluate overall social bias. 
Recently, Xu et al. \cite{xu2024first} introduced LiCoEval, an evaluation benchmark for assessing the license compliance capabilities of LLMs. Additionally, incorporating diverse perspectives in development teams and engaging with stakeholders from various communities can further align Code LLM outputs with ethical standards and societal values.

\textbf{Efficiency}: 
The Efficiency principle emphasizes optimizing the performance and speed of Code LLMs for code generation while minimizing the computational resources required for training and inference. 
For instance, training the GPT-3 model, which consists of 175 billion parameters, demands substantial resources. It requires approximately 1,024 NVIDIA V100 GPUs, costing around $4.6$ million and taking approximately 34 days to complete the training process. 
To address these challenges, various techniques are employed, including model compression methods such as pruning, quantization, and knowledge distillation. Additionally, optimized algorithms like AdamW, parallel strategies such as tensor, pipeline, and data parallelism, and parameter-efficient fine-tuning (PEFT) (see Section \ref{sec:peft}) are often utilized. 
For a comprehensive and detailed discussion on methods to enhance the efficiency of Code LLMs for code generation, please refer to Section 4.5.2, ``Efficiency Enhancement'', in \cite{yang2024robustness}.

\textbf{Safety}: 
The Safety principle of Code LLMs is of utmost importance due to their potential to introduce vulnerabilities, errors, or privacy breaches into software systems. 
Ensuring safety involves comprehensive testing and validation processes to detect and mitigate these risks.
For instance, attackers might compromise the training process of LLMs by injecting malicious examples into the training data, a method known as data poisoning attacks \cite{schuster2021you}. 
Even when attackers lack access to the training process, they may employ techniques like the black-box inversion approach introduced by Hajipour et al. \cite{hajipour2024codelmsec}. This method uses few-shot prompting to identify prompts that coax black-box code generation models into producing vulnerable code. 
Furthermore, Yang et al. \cite{yang2024unveiling} and Al-Kaswan et al. \cite{al2024traces} reveals that Code LLMs, such as CodeParrot \cite{codeparrot}, can memorize training data, potentially outputting personally identifiable information like emails, names, and IP addresses, thereby posing significant privacy risks. Additionally, Yuan et al. \cite{yuan2023gpt} demonstrate that engaging with ChatGPT and GPT-4 in non-natural languages can circumvent safety alignment measures, leading to unsafe outcomes, such as ``The steps involved in stealing money from a bank.''.
To bolster the safety of LLMs in code generation, it is crucial to detect and eliminate privacy-related information from training datasets. 
For example, approaches outlined in \cite{fried2022incoder} and \cite{allal2023santacoder} utilize carefully crafted regular expressions to identify and remove private information from training data. 
To counteract black-box inversion, implementing prompt filtering mechanisms is recommended to identify and block prompts that might result in insecure code generation. 
Moreover, adversarial training can enhance the model's resilience to malicious prompts. Employing reinforcement learning methods can further align Code LLMs with human preferences, thereby reducing the likelihood of producing harmful outputs.

\textbf{Trustworthiness}: 
The Trustworthiness principle focuses on developing Code LLMs that users can depend on for accurate and reliable code generation, which is crucial for their acceptance and widespread adoption. 
Achieving this requires ensuring model transparency, providing explanations for decisions, and maintaining consistent performance across various scenarios.
For instance, Ji et al. \cite{ji2023benchmarking} propose a causal graph-based representation of prompts and generated code to identify the causal relationships between them. This approach offers insights into the effectiveness of Code LLMs and assists end-users in understanding the generation. 
Similarly, Palacio et al. \cite{palacio2023evaluating} introduce ASTxplainer, a tool that extracts and aggregates normalized model logits within Abstract Syntax Tree (AST) structures. This alignment of token predictions with AST nodes provides visualizations that enhance end-user understanding of Code LLM predictions.
Therefore, by prioritizing trustworthiness, we can bolster user confidence and facilitate the integration of Code LLMs into diverse coding environments.
By adhering to the aforementioned principles as key objectives for aligning Code LLMs, researchers and developers can create LLMs for code generation that are not only capable but also ethical, sustainable, and user-centric.

}\label{sec:grest_llm4code}

\begin{table}[p!]
\caption{The overview of code assistant applications powered by \done{LLMs}. The column labeled `\textbf{PLs}' and `\textbf{IDEs}' indicate programming languages and integrated development environments, respectively \cite{zan2023large}.}
\label{tab:products}
\centering
\scalebox{0.71}{
\rotatebox{270}{
    \begin{tabular}{llllll} 
        \toprule
        \textbf{Institution} & \textbf{Products} & \textbf{Model} & \textbf{Supported Features} & \textbf{Supported PLs} & \textbf{Supported IDEs} \\ 
        \toprule
        GitHub \& OpenAI & GitHub Copilot \cite{chen2021evaluating} & Codex & 
        \begin{tabular}[c]{@{}l@{}} Code Completions, Code Generation, \\
        Coding Questions Answering,\\ Code Refactoring, Code Issues Fix, \\Unit Test Cases Generation, \\Code Documentation Generation\end{tabular} &
        \begin{tabular}[c]{@{}l@{}}Java, Python, JavaScript, TypeScript,\\Perl,~R,~PowerShell, Rust,~SQL,~CSS,~\\Ruby,  Julia,~C\#,~PHP,~Swift,~C++,Go,\\HTML, JSON, SCSS,~.NET, Less,\\T-SQL, Markdown\end{tabular} & 
        \begin{tabular}[c]{@{}l@{}}Visual Studio, VS Code,~Neovim,\\JetBrains IDE\end{tabular}  \\ 
        \midrule
        Zhipu AI & CodeGeeX \cite{zheng2023codegeex} & CodeGeeX & 
        \begin{tabular}[c]{@{}l@{}} Code Generation, Code Translation, \\Code Completion, Code Interpretation, \\Code Bugs Fix, Comment Generation, \\AI Chatbot\end{tabular} &
        \begin{tabular}[c]{@{}l@{}}PHP, Go,~C,~C\#,~C++,~Rust,~Perl, CSS,\\Java, Python, JavaScript, TypeScript, \\Objective C++, Objective C, Pascal,\\HTML, SQL,~Kotlin,~R, Shell, Cuda,\\Fortran, Tex, Lean,~Scala\end{tabular} & 
        \begin{tabular}[c]{@{}l@{}}Clion, RubyMine, AppCode,~Aqua,\\IntelliJ IDEA, VS Code, PyCharm,\\Android Studio, WebStorm,~Rider,\\GoLand, DataGrip, DataSpell\end{tabular} \\ 
        \midrule
        Amazon & CodeWhisperer \cite{CodeWhisperer} & $-$ & 
        \begin{tabular}[c]{@{}l@{}} Code Completion, Code Explanation, \\ Code Translation, \\
        Code Security Identification, \\ Code Suggestion \end{tabular} &
        \begin{tabular}[c]{@{}l@{}}Java, Python, TypeScript, JavaScript,\\C\#\end{tabular} & 
        \begin{tabular}[c]{@{}l@{}}JetBrains IDE, VS Code, AWS Cloud9,\\AWS Lambda\end{tabular} \\
        \midrule
        Codeium & Codeium \cite{Codeium} & $-$ & 
        \begin{tabular}[c]{@{}l@{}} Code Completion, Bug Detection,\\Code Suggestions, AI Chatbot,\\Test Type Generation,\\Test Plan Creation,\\
        Codebase Search\end{tabular} &
        \begin{tabular}[c]{@{}l@{}} More than 70 languages in total,\\
        including but not limited to:\\
        C, C\#, C++, Dart, CSS, Go, Elixir,\\HTML, Haskell, Julia, Java, JavaScript,\\Lisp, Kotlin, Lua, Objective-C,\\Perl, Pascal, PHP, Protobuf,\\R, Python, Ruby, Scala, Rust,\\Swift, SQL, TS, Vue\end{tabular} & 
        \begin{tabular}[c]{@{}l@{}} JetBrains, VSCode, Visual Studio,\\Colab, Jupyter, Deepnote,\\Notebooks, Databricks, Chrome,\\Vim, Neovim, Eclipse, Emacs,\\
        VSCode Web IDEs, Sublime Text\end{tabular} \\
        \midrule
        Huawei & CodeArts Snap \cite{shen2023pangu} & PanGu-Coder &
        \begin{tabular}[c]{@{}l@{}}Code Generation, Code Explanation\\ Research and Development Knowledge\\ Question and Answer\\ Code Comment, Code Debug \\Unit Test Case Generation \end{tabular} &
        \begin{tabular}[c]{@{}l@{}}Java, Python \end{tabular} & 
        \begin{tabular}[c]{@{}l@{}}PyCharm, VS Code, IntelliJ \end{tabular} \\
        \midrule
        Tabnine & TabNine \cite{TabNine} & $-$ & 
        \begin{tabular}[c]{@{}l@{}}Code Generation, Code Completion,\\Code Explanation, Bug Fix,\\
        Code Recommendation, Code Refactoring,\\
        Code Test Generation,\\Docstring Generation\end{tabular} &
        \begin{tabular}[c]{@{}l@{}}Python, Javascript, Java, TypeScript,\\HTML, Haskell, Matlab, Kotlin, Sass,\\Go, PHP, Ruby, C, C\#, C++,~Swift,~\\Rust,~CSS,~Perl,~Angular, Dart, React,\\Objective C, NodeJS, Scala,~\end{tabular} & 
        \begin{tabular}[c]{@{}l@{}}Sublime, PyCharm, Neovim,~Rider,\\VS Code, IntelliJ IDE, Visual Studio,\\PhpStorm, Vim, RubyMine,~DataGrip,\\Android Studio, WebStorm,~Emacs,\\Clion, Jupyter Notebook,~JupyterLab,\\Eclipse,~GoLand, AppCode\end{tabular}  \\ 
        \midrule
        Replit & Replit\cite{Replit} & replit-code & 
        \begin{tabular}[c]{@{}l@{}} Code Completion, Code Editing,\\Code Generation, Code Explanation,\\
        Code Suggestion, Code Test Generation\end{tabular} &
        \begin{tabular}[c]{@{}l@{}} C\#, Bash, C, CSS, C++, Java, Go,\\HTML, JavaScript, Perl, PHP,\\
        Ruby, Python, R, SQL, Rust\end{tabular} & 
        \begin{tabular}[c]{@{}l@{}} $-$\end{tabular} \\
        \bottomrule
    \end{tabular}
}
}
\end{table}
\begin{figure*}[t]
\centering
\includegraphics[width=0.95\linewidth]{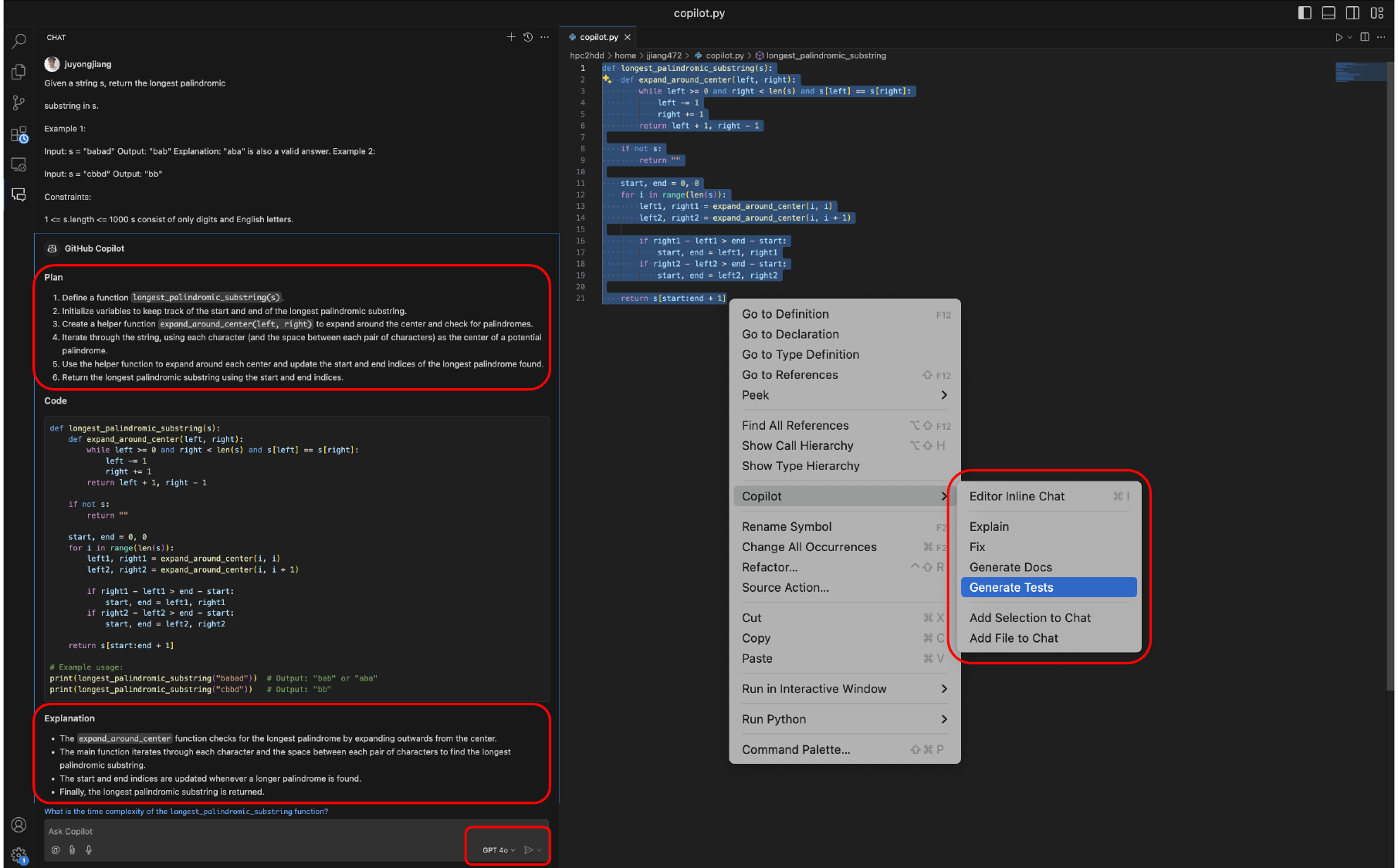}
\includegraphics[width=0.95\linewidth]{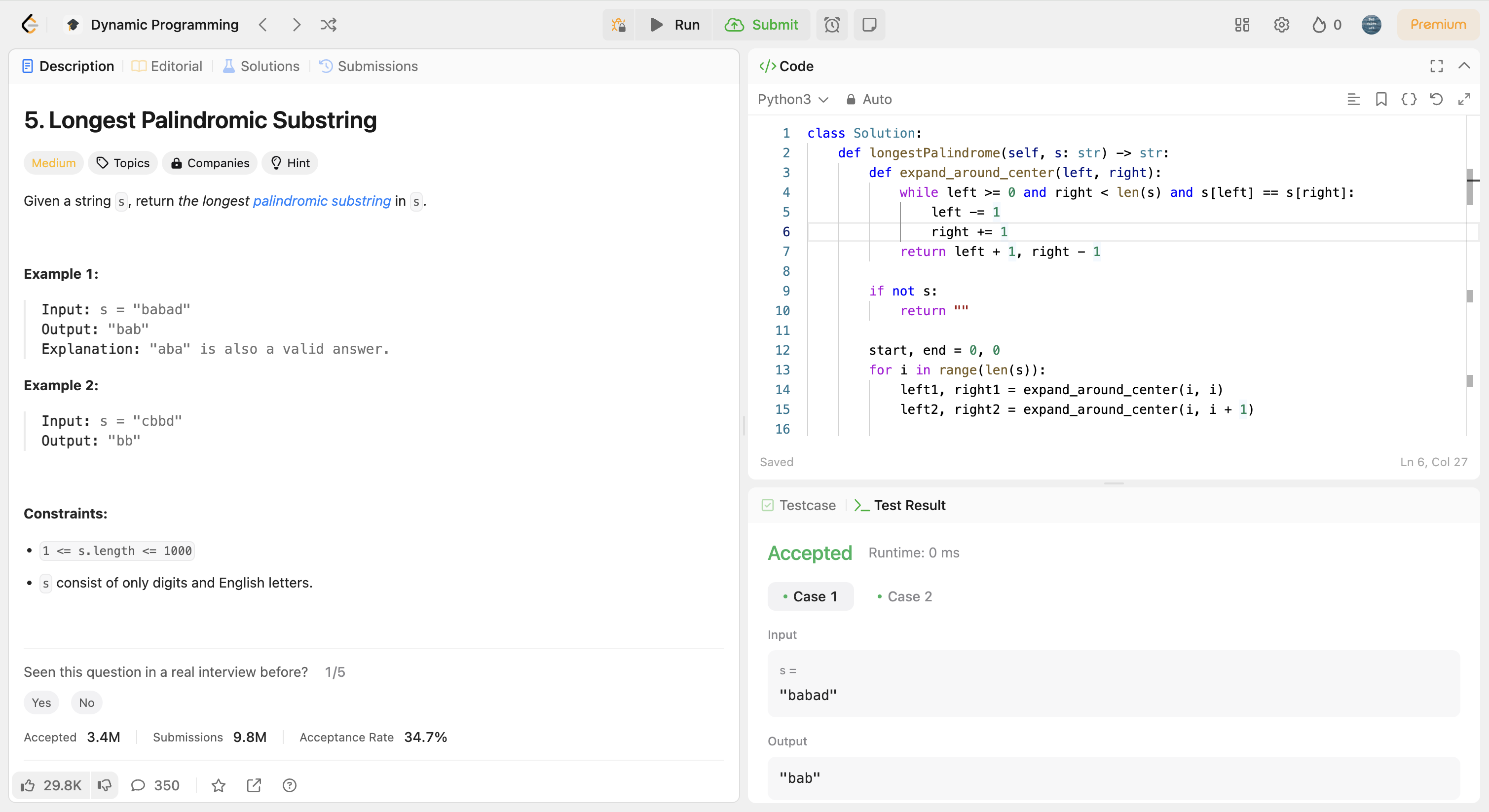}
\caption{\done{An exemplar of GitHub Copilot to demonstrate how to use development tools powered by LLMs, including powerful GPT 4o, o1-preview (Preview), and o1-mini (Preview). 
To illustrate its capabilities, we input the description of the ``5. Longest Palindromic Substring'' problem from LeetCode into Copilot's chat box. 
The code generated by Copilot is then submitted to the online judge platform, where it is successfully accepted.
}}
\label{fig:copilot}
\end{figure*}
\subsection{Applications}\label{sec:application}
Code LLMs have been integrated with development tools and platforms, such as
integrated development environments (IDEs) and version control systems, improving programming efficiency substantially. In this section, we will briefly introduce several widely used applications as coding assistants. The statistics of these applications are provided in Table \ref{tab:products}.

\textbf{GitHub Copilot.}
GitHub Copilot, powered by OpenAI's Codex, is an AI pair programmer that helps you write better code faster. Copilot suggests whole lines or blocks of code as you type, based on the context provided by your existing code and comments. It's trained on a dataset that includes a significant portion of the public code available on GitHub, which enables it to understand a wide range of programming languages and coding styles. Copilot not only improves productivity but also serves as a learning tool by providing programmers with examples of how certain functions can be implemented or how specific problems can be solved.

\textbf{CodeGeeX.}
CodeGeeX stands out as a multifaceted programming assistant, proficient in code completion, comment generation, code translation, and developer interactions. 
Its underlying code generation LLM has been refined with extensive training on vast amounts of code data, exhibiting superior performance on benchmarks like HumanEval, HumanEval-X, and DS1000. 
Renowned for supporting multilingual code generation, CodeGeeX plays a pivotal role in enhancing the efficiency of code development.

\textbf{CodeWhisperer.}
Amazon's CodeWhisperer is a versatile, machine learning-driven code generator that offers on-the-fly code recommendations. Tailored to your coding patterns and comments, CodeWhisperer provides personalized suggestions that range from succinct comments to complex functions, all aimed at streamlining your coding workflow.

\textbf{Codeium.}
Codeium is an AI-accelerated coding toolkit that offers a suite of functions, including code completion, explanation, translation, search, and user chatting. Compatible with over 70 programming languages, Codeium delivers fast and cutting-edge solutions to coding challenges, simplifying the development process for its users.

\textbf{CodeArts Snap.}
Huawei's CodeArts Snap is capable of generating comprehensive function-level code from both Chinese and English descriptions. This tool not only reduces the monotony of manual coding but also efficiently generates test code, in addition to providing automatic code analysis and repair services.

\textbf{Tabnine.}
Tabnine is an AI coding assistant that empowers development teams to leverage AI for streamlining the software development lifecycle while maintaining strict standards for privacy, security, and compliance. With a focus on enhancing coding efficiency, code quality, and developer satisfaction, Tabnine offers AI-driven automation that is tailored to the needs of your team. Supporting over one million developers worldwide, Tabnine is applicable across various industries.

\textbf{Replit.}
Replit is a multifunctional platform that caters to a diverse array of software development needs. As a complimentary online IDE, it facilitates code collaboration, and cloud services, and fosters a thriving developer community. Replit also enables users to compile and execute code in more than 50 programming languages directly within a web browser, eliminating the need for local software installations.

\done{
To illustrate the use of development tools powered by LLMs, we employ GitHub Copilot within Visual Studio Code (VS Code) as our example. Note that 
\begin{itemize}
    \item [\textcircled{1}] For details on using the GitHub Copilot extension in VS Code, please refer to the useful document at \href{https://code.visualstudio.com/docs/copilot/overview}{https://code.visualstudio.com/docs/copilot/overview}. 
    \item [\textcircled{2}] If you would like to get free access to Copilot as a student, teacher, or open-source maintainer, please refer to this tutorial at \href{https://docs.github.com/en/copilot/managing-copilot/managing-copilot-as-an-individual-subscriber/managing-your-copilot-subscription/getting-free-access-to-copilot-as-a-student-teacher-or-maintainer}{https://docs.github.com/en/copilot/managing-copilot/managing-copilot-as-an-individual-subscriber/managing-your-copilot-subscription/getting-free-access-to-copilot-as-a-student-teacher-or-maintainer} and GitHub education application portal at \\ \href{https://education.github.com/discount_requests/application}{https://education.github.com/discount\_requests/application}.
\end{itemize}
As depicted in the upper section of Figure \ref{fig:copilot}, users can interact with Copilot through the chat box in the lower left corner, where they can inquire about various coding-related tasks. 
This feature is now supported by the advanced capabilities of GPT-4o, o1-preview (Preview), and o1-mini (Preview).
From the generated content, Copilot demonstrates the ability to plan solutions to coding problems. It can write code and subsequently explain the generated code to enhance user comprehension. 
Within the right-side workspace, users can engage in inline chat conversations to generate or refactor source code, conduct code explanations, fix coding errors, resolve issues encountered during terminal command executions, produce documentation comments, and generate unit tests.
To illustrate its capabilities, we input the description of the ``5. Longest Palindromic Substring'' problem from LeetCode into Copilot's chat box. 
The code generated by Copilot is then submitted to the online judge platform, where it is successfully accepted, as shown at the lower section of Figure \ref{fig:copilot}.
}
\section{Challenges \& Opportunities}\label{sec:challenges}
According to our investigations, the LLMs have revolutionized the paradigm of code generation and achieved remarkable performance. 
Despite this promising progress, there are still numerous challenges that need to be addressed. These challenges are mainly caused by the gap between academia and practical development. For example, in academia, the HumanEval benchmark has been established as a de facto standard for evaluating the coding proficiency of LLMs. However, many works have illustrated the evaluation of HumanEval can't reflect the scenario of practical development \cite{jimenez2023swe,du2024evaluating,liu2024your,ding2024crosscodeeval}.
In contrast, these serious challenges offer substantial opportunities for further research and applications. 
In this section, we pinpoint critical challenges and identify promising opportunities, aiming to bridge the research-practicality divide. 

\textbf{Enhancing complex code generation at repository and software scale.}
In practical development scenarios, it often involves a large number of complex programming problems of varying difficulty levels \cite{zhang2022automated,li2022competition}. 
While LLMs have shown proficiency in generating function-level code snippets, these models often struggle with more complex, unseen programming problems, repository- and software-level problems that are commonplace in real-world software development.
To this end, it requires strong problem-solving skills in LLM beyond simply functional-level code generation. 
For example, 
AlphaCode \cite{li2022competition} achieved an average ranking in the top 54.3\% in programming competitions where an understanding of algorithms and complex natural language is required to solve competitive programming problems.
\cite{jimenez2023swe} argues that existing LLMs can't resolve real-world GitHub issues well since the best-performing model, Claude 2, is able to solve a mere 1.96\% of the issues.
The reason for poor performance is mainly attributed to the weak reasoning capabilities \cite{huang2022towards}, complex internal- and external- dependencies \cite{bairi2023codeplan}, and context length limitation of LLMs \cite{bairi2023codeplan}.
Therefore, the pursuit of models that can handle more complex, repository- and software-level code generation opens up new avenues for automation in software development and makes programming more productive and accessible.


\textbf{Innovating model architectures tuned to code structures.}
Due to their scalability and effectiveness, Transformer-based LLM architectures have become dominant in solving code generation task. 
Nevertheless, they might not be optimally designed to capture the inherent structure and syntax of programming languages (PLs) \cite{guo2020graphcodebert,guo2022unixcoder,ma2022code,kou2023model}. Code has a highly structured nature, with a syntax that is more rigid than natural language. This presents a unique challenge for LLMs, which are often derived from models that were originally designed for natural language processing (NLP).
The development of novel model architectures that inherently understand and integrate the structural properties of code represents a significant opportunity to improve code generation and comprehension. Innovations such as tree-based neural networks \cite{mou2014tbcnn}, which mirror the abstract syntax tree (AST) representation of code, can offer a more natural way for models to learn and generate programming languages. Additionally, leveraging techniques from the compiler theory, such as intermediate representations (IR) \cite{li2022unleashing}, could enable models to operate on a more abstract and generalizable level, making them effective across multiple programming languages \cite{paul2024ircoder}. By exploring architectures beyond the traditional sequential models, researchers can unlock new potentials in code generation.

\textbf{Curating high-quality code data for pre-training and fine-tuning of LLMs.}
The efficacy of LLMs largely depends on the quality and diversity of code datasets used during pre-training and fine-tuning phases \cite{zhou2024lima,kopf2024openassistant,wettig2024qurating}. Currently, there is a scarcity of large, high-quality datasets that encompass a wide range of programming tasks, styles, and languages. This limitation constrains the ability of LLMs to generalize across unseen programming tasks, different coding environments, and real-world software development scenarios.
The development of more sophisticated data acquisition techniques, such as automated code repositories mining \cite{linstead2007mining}, advanced filtering algorithms, and code data synthesis \cite{liu2024best} (see Section \ref{sec:data_synthesis}), can lead to the creation of richer datasets. Collaborations with industry partners (e.g., GitHub) could also facilitate access to proprietary codebases, thereby enhancing the practical relevance of the training material. Furthermore, the adoption of open-source models for dataset sharing can accelerate the collective effort to improve the breadth and depth of code data available for LLM research.

\textbf{Developing comprehensive benchmarks and metrics for coding proficiency evaluation in LLMs.}
Current benchmarks like HumanEval may not capture the full spectrum of coding skills required for practical software development \cite{ni2023l2ceval}. Additionally, metrics often focus on syntactic correctness or functional accuracy, neglecting aspects such as code efficiency \cite{peitek2021program}, style \cite{chen2023duetcs}, readability \cite{buse2009learning}, or maintainability \cite{ardito2020tool}.
The design of comprehensive benchmarks that simulate real-world software development challenges could provide a more accurate assessment of LLMs' coding capabilities. These benchmarks should include diverse programming tasks of varying difficulty levels, such as debugging \cite{zhong2024ldb}, refactoring \cite{shirafuji2023refactoring}, and optimization \cite{ishida2024langprop}, and should be complemented by metrics that evaluate qualitative aspects of code. The establishment of community-driven benchmarking platforms could facilitate continuous evaluation and comparison of LLMs for code generation across the industry and academia.

\textbf{Support for low-resource, low-level, and domain-specific programming languages.}
LLMs are predominantly trained in popular high-level programming languages, leaving low-resource, low-level, and domain-specific languages underrepresented. This lack of focus restricts the applicability of LLMs in certain specialized fields and systems programming \cite{thakur2023benchmarking}.
Intensifying research on transfer learning and meta-learning approaches may enable LLMs to leverage knowledge from high-resource languages to enhance their performance on less common ones \cite{chen2022transferability,cassano2023knowledge}. 
Additionally, partnerships with domain experts can guide the creation of targeted datasets and fine-tuning strategies to better serve niche markets. The development of LLMs with a capacity for multilingual code generation also presents a significant opportunity for broadening the scope of applications.

\textbf{Continuous learning for LLMs to keep pace with evolving coding knowledge.}
The software development landscape is continuously evolving, with new languages, frameworks, and best practices emerging regularly. LLMs risk becoming outdated if they cannot adapt to these changes and incorporate the latest programming knowledge \cite{jang2022towards,wang2023knowledge}.
While retrieval augmented code generation mitigates these issues, the performance is limited by the quality of the retrieval context 
While retrieval-augmented code generation offers a partial solution to these issues, its effectiveness is inherently constrained by the quality of retrieved context.
\cite{lu2022reacc,zhou2022docprompting,zhang2023repocoder}. 
Therefore, establishing mechanisms for continuous learning and updating of LLMs can help maintain their relevance over time. This could involve real-time monitoring of code repositories to identify trends and innovations, as well as the creation of incremental learning systems that can assimilate new information without forgetting previously acquired knowledge. Engaging the LLMs in active learning scenarios where they interact with human developers may also foster ongoing knowledge acquisition.

\textbf{Ensuring code safety and aligning LLM outputs with human coding preferences.}
Ensuring the safety and security of code generated by LLMs is a paramount concern, as is their ability to align with human preferences and ethical standards. Current models may inadvertently introduce vulnerabilities or generate code that does not adhere to desired norms \cite{chen2021evaluating,yang2024robustness}.
Research into the integration of formal verification tools within the LLM pipeline can enhance the safety of the produced code. Additionally, developing frameworks for alignment learning that capture and reflect human ethical preferences can ensure that the code generation process aligns with societal values \cite{ouyang2022training,qi2023fine}. Transparent and explainable AI methodologies can also contribute to building trust in the LLM-generated code by making the decision-making process more accessible to developers.
\section{Conclusion}\label{sec:conclusion}
In this survey, we provide a systematic literature review, serving as a valuable reference for researchers investigating the cutting-edge progress in LLMs for code generation. A thorough introduction and analysis for data curation, the latest advances, performance evaluation, \done{ethical implications, environmental impact,} and real-world applications are illustrated.  
In addition, we present a historical overview of the evolution of LLMs for code generation in recent years and offer an empirical comparison using the widely recognized \done{HumanEval, MBPP, and the more practical and challenging BigCodeBench benchmarks} to highlight the progressive enhancements in LLM capabilities for code generation. 
Critical challenges and promising opportunities regarding the gap between academia and practical development are also identified for future investigation. 
Furthermore, we have established a dedicated resource website to continuously document and disseminate the most recent advances in the field.
We hope this survey can contribute to a comprehensive and systematic overview of LLM for code generation and promote its thriving evolution. 
We optimistically believe that LLM will ultimately change all aspects of coding and automatically write safe, helpful, accurate, trustworthy, and controllable code, like professional programmers, and even solve coding problems that currently cannot be solved by humans.


\bibliographystyle{ACM-Reference-Format}
\bibliography{ref}

\end{document}